%% file: samplepaper.tex
\documentclass{article}

\usepackage{arxiv}

\usepackage[utf8]{inputenc}
\usepackage[T1]{fontenc}

\usepackage{amsmath, amssymb, amsfonts}
\usepackage{bm}

\usepackage{graphicx}
\usepackage{subcaption}
\usepackage{caption}
\usepackage{wrapfig}
\usepackage{adjustbox}

\usepackage{booktabs}
\usepackage{tabularx}
\usepackage{array}

\usepackage{algorithm}
\usepackage{algpseudocode}

\usepackage{url}
\usepackage{nicefrac}
\usepackage{microtype}
\usepackage{xcolor}
\usepackage{comment}
\usepackage{cancel}
\usepackage{placeins}

\usepackage[numbers]{natbib}

\usepackage{hyperref}
\hypersetup{
  colorlinks=true,
  linkcolor=blue,
  citecolor=blue,
  urlcolor=blue,
  filecolor=blue,
}

\newcommand{\ms}[2]{\ensuremath{#1_{\pm #2}}}
\newcommand{\bms}[2]{\ensuremath{\mathbf{#1}_{\pm #2}}}

\newcommand{\maketitlesupplementary}{%
  \begin{center}
    {\LARGE Supplementary Material}\\[0.8em]
    {\large ZAYAN: Disentangled Contrastive Transformer for Tabular Remote Sensing Data\footnotemark}
  \end{center}
  \footnotetext{This paper has been accepted for presentation at the 28th International Conference on Pattern Recognition (ICPR 2026) in Lyon, France.}
}

\begin{document}

\title{
ZAYAN: Disentangled Contrastive Transformer for Tabular Remote Sensing Data
\thanks{
This paper has been accepted for presentation at the 28th International Conference on Pattern Recognition (ICPR 2026) in Lyon, France. 
Code: \url{https://github.com/zadid6pretam/ZAYAN}. 
PyPI: \texttt{pip install zayan}.
}
}

\author{
  \textsuperscript{*}Al Zadid Sultan Bin Habib\textsuperscript{1}
  \quad
  Tanpia Tasnim\textsuperscript{2}
  \quad
  Md.\ Ekramul Islam\textsuperscript{3}
  \quad
  \textsuperscript{+}Muntasir Tabasum\textsuperscript{1}
  \\[0.5em]
  \textsuperscript{1}Lane Dept. of Computer Science and Electrical Engineering\textsuperscript{*}
  and Dept. of Geology \& Geography\textsuperscript{+},\\
  West Virginia University, Morgantown, WV 26506, USA\\
  \texttt{\{\textsuperscript{*}ah00069,\textsuperscript{+}mt00079\}@mix.wvu.edu}
  \\[0.5em]
  \textsuperscript{2}Department of Computer Science and Engineering,\\
  Green University of Bangladesh, Narayanganj-1461, Bangladesh\\
  \texttt{tanpia@cse.green.edu.bd}
  \\[0.5em]
  \textsuperscript{3}Department of Computer Science and Engineering,\\
  Stamford University Bangladesh, Dhaka-1217, Bangladesh\\
  \texttt{eislam706@gmail.com}
}
\date{}
\maketitle
\bibpunct{[}{]}{,}{n}{}{,}
\begin{abstract}
Learning informative representations from tabular data in remote sensing and environmental science is challenging due to heterogeneity, scarce labels, and redundancy among features. We present ZAYAN (\underline{Z}ero-\underline{A}nchor d\underline{Y}namic fe\underline{A}ture e\underline{N}coding), a self-supervised, feature-centric contrastive framework for tabular data. ZAYAN performs contrastive learning at the feature rather than sample level, removing the need for explicit anchor selection and any reliance on class labels, while encouraging a redundancy-minimized, disentangled embedding space. The framework has two modules: ZAYAN-CL, which pretrains feature embeddings via a zero-anchor contrastive objective with dynamic perturbations/masking, and ZAYAN-T, a Transformer that conditions on these embeddings for downstream classification. Across eight datasets (six remote-sensing tabular benchmarks and two remote-sensing-driven flood-prediction tables from satellite and GIS products), ZAYAN achieves superior accuracy, robustness, and generalization over tabular deep learning baselines, with consistent gains under label scarcity and distribution shift. These results indicate that feature-level contrastive learning and dynamic feature encoding provide an effective recipe for learning from tabular sensing data.

\end{abstract}
\section{Introduction}
\label{intro}
Remote sensing and environmental analyses increasingly rely on large-scale tabular datasets derived from satellite imagery, sensor networks, and environmental surveys~\cite{wang2022self,jain2022multimodal}. Such data bring high dimensionality, feature redundancy, measurement noise, and limited labels, which can undermine conventional supervised learning~\cite{wang2025survey,yoon2020vime}. As a result, Self-Supervised Learning (SSL), especially Contrastive Learning (CL), has become attractive for building label-efficient representations~\cite{chen2020simple,bahriscarf}. Most CL frameworks, however, operate at the instance level with anchor-based objectives and view augmentations, paying less attention to feature-level structure that is central to tabular representation quality~\cite{stojnic2021self}. Remote-sensing SSL methods such as SeCo~\cite{Introduc35:online}, CACo~\cite{mall2023change}, MATTER~\cite{akiva2022self}, SACo+~\cite{stival2025semantically}, SwiMDiff~\cite{tian2024swimdiff}, and Cross-Scale MAE~\cite{tang2023cross} leverage temporal, multimodal, or multi-scale signals to pretrain image encoders, yet they remain instance/patch-centric and do not dynamically encode feature interactions for tabular inputs. Building on recent benchmark studies that systematically compare tabular deep learning and classical models for urban land cover classification~\cite{tdl}, we extend this line of work to a broader suite of remote-sensing and related datasets and ask: can we design a feature-level SSL framework that explicitly tackles redundancy and heterogeneity in environmental tabular data?

We propose ZAYAN (\underline{Z}ero-\underline{A}nchor d\underline{Y}namic fe\underline{A}ture e\underline{N}coding), a self-supervised Transformer framework that integrates feature-level CL. ZAYAN employs a zero-anchor objective that contrasts features rather than samples, removing the need for explicit anchors and labels. The framework has two components: (1) ZAYAN-CL, a pretraining module that learns informative, disentangled, and redundancy-reduced feature embeddings using a feature-level InfoNCE loss with dynamic perturbations/masking and a redundancy penalty; and (2) ZAYAN-T, a downstream Transformer that consumes these embeddings for supervised classification while preserving the angular/structural relations established during pretraining. Intuitively, ZAYAN encourages each feature to be close to its perturbed views while pushing apart redundant or uninformative directions, yielding a representation where environmental signals of interest are amplified and noise is suppressed. Our contributions are:
\begin{itemize}
  \item[\textbf{C1}] \textbf{ZAYAN-CL (feature-level contrastive pretraining).}
  A feature-level, zero-anchor contrastive pretraining scheme that yields robust,
  disentangled, and redundancy-minimized embeddings without anchors or labels
  for tabular remote sensing data.

  \item[\textbf{C2}] \textbf{ZAYAN-T (Transformer backbone).}
  A Transformer backbone that leverages the pretrained feature embeddings and
  preserves their structural and angular geometry for improved supervised
  classification.

  \item[\textbf{C3}] \textbf{Comprehensive remote-sensing benchmarks.}
  An extensive evaluation on eight tabular sensing datasets spanning remote-sensing
  benchmarks and relevant flood-event classification, derived from satellite, GIS,
  environmental, and socio-economic indicators, building on and extending prior
  tabular benchmark studies in this domain~\cite{tdl}.

  \item[\textbf{C4}] \textbf{Robustness and deployment-oriented gains.}
  Consistent gains over traditional machine-learning and tabular deep-learning
  baselines in predictive performance, robustness, and generalization, including
  under label scarcity, distribution shift, and deployment-oriented diagnostics
  (calibration, OOD behaviour, and triage-style evaluation).
\end{itemize}

\section{Related Work}
\label{related}
We review SSL for remote sensing, Transformer and other deep models for tabular data, and applied machine learning for image-derived remote sensing and flood prediction.

\textbf{A. SSL in remote sensing.}
Remote-sensing SSL has progressed from patch/view-level contrast to spatio-temporal and multi-scale pretraining. Early contrastive methods align augmented views or seasonal variants of the same scene patch (e.g., CMC \cite{stojnic2021self}, SeCo \cite{Introduc35:online}) and incorporate semantic or multispectral cues (SACo+ \cite{stival2025semantically}). MATTER refines material/texture representations for downstream classification \cite{akiva2022self}. Masked reconstruction and cross-scale design further improve pretraining (Cross-Scale MAE \cite{tang2023cross}, SatMAE \cite{cong2022satmae}), while diffusion-based matching captures global context (SwiMDiff \cite{tian2024swimdiff}). Geography-aware SSL injects spatial coordinates to respect geospatial relations \cite{ayush2021geography}. To address scarcity and shift, few-shot and change-aware schemes adapt pretrained encoders or mine challenging positives (RS-FewShotSSL \cite{RS-FewShotSSL}, CACo \cite{mall2023change}). These advances largely remain instance/patch-centric and do not explicitly model feature-level interactions in tabular inputs.

\textbf{B. Transformer models for tabular data.}
Transformers have been adapted to structured data via attention over features and tokens. TabNet uses sequential attentive masks to select salient features \cite{b1}. TabTransformer embeds categorical fields as tokens and attends over feature dependencies before fusing with numerical variables \cite{b3}. FT-Transformer treats all features as tokens with regularized attention and virtual tokens \cite{b5}. SAINT augments Transformer blocks with inter-sample contrastive/triplet objectives \cite{saint}, and AutoInt targets higher-order feature interactions via multi-head attention \cite{b40}. Data-prior pretraining with TabPFN and TabPFN v2 trains compact Transformers on massive synthetic tables for few-shot inference \cite{b4,tabpfnv2}. Despite strong results, these methods typically operate at the sample level, treat features uniformly as tokens, and do not directly disentangle redundant feature dimensions.

\textbf{C. Other tabular deep learning and classic models.}
Beyond Transformers, a growing family of specialized tabular architectures remains highly competitive. 
Recent nearest-neighbour–aware models such as TabR~\cite{tabr} and the parameter-efficient ensemble TabM~\cite{tabm} push state-of-the-art performance by combining strong tabular backbones with kNN-style refinement and lightweight ensembling. TabICL~\cite{tabicl} proposes a tabular foundation model for in-context learning on large collections of tables, while TANDEM~\cite{tandem} leverages hybrid autoencoders and model-based augmentation to improve robustness in low-label regimes. ProtoGate~\cite{protogate} introduces prototype-based networks with global-to-local feature selection tailored to biomedical data, and NLP-inspired sequence models such as TabulaRNN~\cite{mamb} further explore recurrent and RNN-style designs for tabular deep learning. In parallel, earlier architectures like TabSeq, which imposes a sequential feature ordering via an attention-guided autoencoder~\cite{b6}, DeepFM, which unifies factorization machines with feed-forward networks~\cite{deepfm}, TANGOS, which regularizes toward sparse informative subsets via graph-based propagation~\cite{b42}, NODE, which embeds oblivious decision trees into a differentiable ensemble~\cite{b2}, and Deep \& Cross Networks (DCN), which explicitly model feature crosses~\cite{dcn}, remain strong neural baselines. On the classic side, gradient-boosted decision trees and related families (AdaBoost~\cite{freund1997decision}, GBM~\cite{friedman2001greedy}, LGBM~\cite{lgbm}, XGBoost~\cite{b45}, CatBoost~\cite{catboost}), together with logistic regression, random forests, SVMs, decision trees, $k$-nearest neighbours, naive Bayes, 1-D CNNs, and shallow MLPs, continue to be competitive reference points on many tabular benchmarks.

\textbf{D. Applied ML for remote sensing and environmental data.}
Land-cover mapping pipelines integrate crowdsourced data, spectral indices, and hierarchical processing: OpenStreetMap–Landsat fusion with random forests/naive Bayes \cite{Johnson2016}, spatially weighted segment-level fusion \cite{Johnson2015}, and standardized products from NLCD, C-CAP, CDL, and LANDFIRE \cite{Wickham2014, Jin2019}. Comparative studies evaluate deep, ensemble, and SVM methods for object-based urban classification \cite{Jozdani2019}, while demographic drivers of land transitions are modeled statistically \cite{Ducey2018}. Time-series phenology from Sentinel-2 supports standardized vegetation dynamics \cite{Sanchez2020phenology}, and GPU-accelerated BFAST enables near-real-time break detection \cite{vonMehren2018bfast}. Flood prediction leverages terrain/hydrology predictors with tree ensembles \cite{Wang2015rflood}, comprehensive ML surveys \cite{Mosavi2018}, multi-basin generalization frameworks \cite{Sidrane2019multiBasin}, and bi-temporal attention U-Nets for flood detection \cite{Yadav2022siamese}.

\textbf{Summary and gap.}
Despite substantial progress, remote-sensing SSL is predominantly instance/patch-level, and tabular Transformers often treat every feature as an identical token, leaving redundancy and heterogeneity under-modeled in remote sensing and environmental tabular data. Building on the benchmark study of Tabasum et al.\ that systematically compared tabular deep learning and classical models for urban land cover classification~\cite{tdl}, we extend this line of work to a broader suite of tabular remote sensing datasets and a richer family of modern tabular architectures. Methods that rely on large labeled corpora or heavy synthetic pretraining can be brittle under label scarcity and domain shift. ZAYAN addresses these gaps with zero-anchor, feature-level contrastive pretraining coupled with a Transformer that preserves disentangled, redundancy-minimized embeddings tailored to tabular data of remote sensing or environmental sources.
\section{Methodology}
\label{method}
\begin{figure*}[t]
  \centering
  \includegraphics[width=\textwidth]{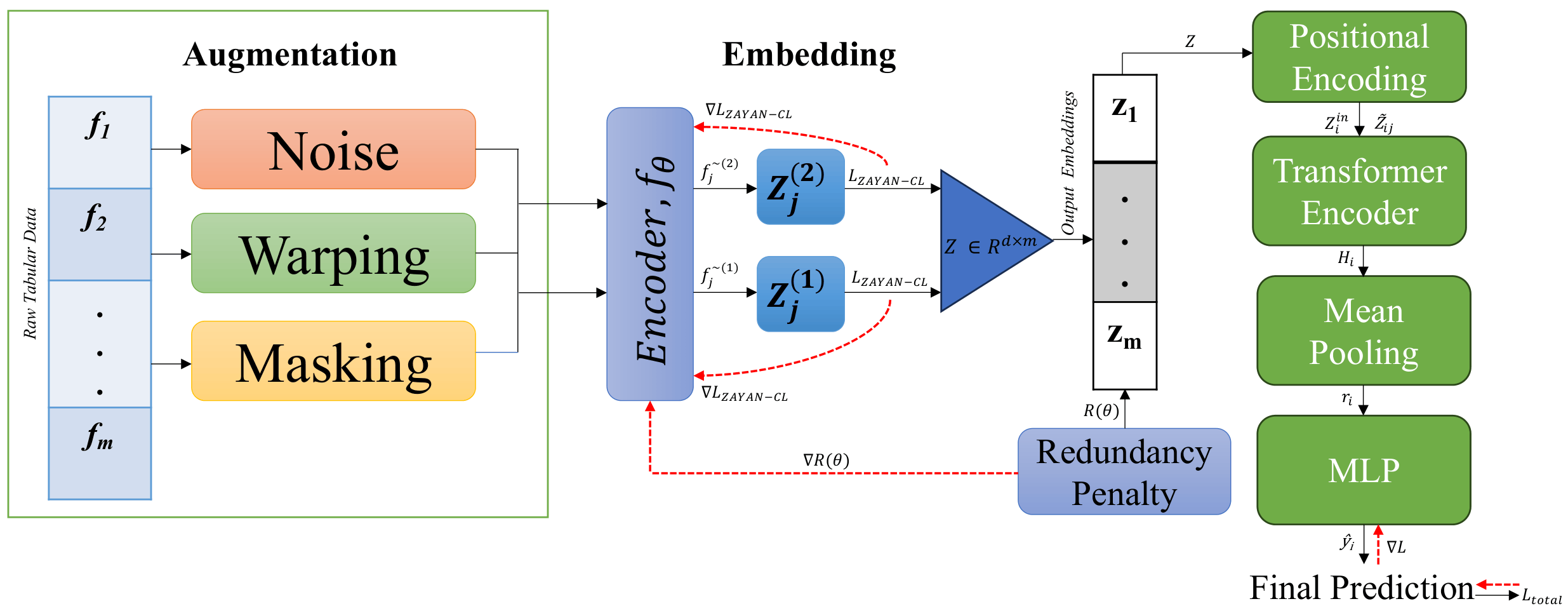}
\caption{
End-to-end architecture of the proposed ZAYAN framework. Raw tabular features undergo stochastic augmentations (noise, warping, masking) and are encoded by a shared encoder $f_\theta$ trained via feature-level contrastive loss $\mathcal{L}_{\text{ZAYAN-CL}}$ and redundancy penalty $\mathcal{R}(\theta)$. The embeddings $Z$ are then processed through a Transformer with positional encoding, aggregated via mean pooling, and passed through an MLP for the final prediction $\hat{y}_i$. Gradients from supervised and contrastive losses jointly update $f_\theta$ (indicated by red dashed arrows).
}

  \label{fig:zayan_arch}
\end{figure*}

ZAYAN is a self-supervised feature learning-enabled Transformer framework designed for remote sensing tabular data. Zero-anchor dynamic feature encoding refers to contrastive feature representation learning without relying on anchor samples or labels, dynamically encoding each feature independently into a structured embedding space. ZAYAN consists of two key components: (1) ZAYAN-CL, a feature-level contrastive learning module for pretraining, and (2) ZAYAN-T, a ZAYAN-aware Transformer that leverages the learned feature representations for downstream classification. Unlike traditional sample-level contrastive paradigms, ZAYAN disentangles feature embeddings by aligning augmented views of each feature independently without relying on anchors or class labels, thus enabling scalable and semantically robust representation learning. Figure~\ref{fig:zayan_arch} illustrates the architecture for ZAYAN combining both ZAYAN-CL and ZAYAN-T.\newline
\textbf{A. ZAYAN-CL:}
Given a tabular dataset $X \in \mathbb{R}^{N \times m}$ with $N$ samples and $m$ features. Each feature column $\mathbf{f}_j \in \mathbb{R}^N$ represents the $j$-th feature vector which is individually augmented to produce two stochastic views. ZAYAN-CL aims to find a set of feature embeddings $\{\mathbf{z}_j \in \mathbb{R}^d\}_{j=1}^m$, where $d$ is the embedding dimension. These embeddings are optimized to maximize agreement between augmented views of the same feature, minimize redundancy across features, and reside on a unit hypersphere to preserve angular diversity.  We define an encoder $f_\theta: \mathbb{R}^N \rightarrow \mathbb{R}^d$ mapping features to $d$-dimensional embeddings $\mathbf{z}_j \in \mathbb{R}^d$. The encoder's goal is to produce representations that minimize redundancy while maximizing informativeness. We solve Eq.~\ref{eq1}.
\begin{equation}
\label{eq1}
\min_{\theta} \; \mathcal{L}_\text{ZAYAN-CL}(\theta) + \lambda \cdot \mathcal{R}(\theta)
\end{equation}
Where $\mathcal{L}_\text{ZAYAN-CL}(\theta)$ is a feature-level contrastive InfoNCE loss, $\mathcal{R}(\theta)$ is a redundancy penalty, and $\lambda \geq 0$ balances these two objectives. To learn robust embeddings, we generate two augmented views $\tilde{\mathbf{f}}_j^{(1)}, \tilde{\mathbf{f}}_j^{(2)}$ for each feature vector $\mathbf{f}_j$. We apply Gaussian noise such that $\tilde{\mathbf{f}}_j^{(1)} = \mathbf{f}_j + \epsilon$, where $\epsilon \sim \mathcal{N}(0, \sigma^2)$. We also apply quantile warping with jitter, which maps original feature values to quantile ranks and perturbs them, along with random masking that replaces elements in $\mathbf{f}_j$ with zeros. Given embeddings $\mathbf{z}_j^{(1)} = f_\theta(\tilde{\mathbf{f}}_j^{(1)})$ and $\mathbf{z}_j^{(2)} = f_\theta(\tilde{\mathbf{f}}_j^{(2)})$, ZAYAN-CL employs a feature-level InfoNCE-based contrastive loss defined in Eq.~\ref{eq2}:
\begin{equation}
\label{eq2}
\mathcal{L}_\text{ZAYAN-CL}(\theta) = -\sum_{j=1}^{m} \log \frac{\exp(\frac{\mathbf{z}_j^{(1)\top}\mathbf{z}_j^{(2)}}{\tau\|\mathbf{z}_j^{(1)}\|\|\mathbf{z}_j^{(2)}\|})}{\sum_{k \neq j} \exp(\frac{\mathbf{z}_j^{(1)\top}\mathbf{z}_k^{(2)}}{\tau\|\mathbf{z}_j^{(1)}\|\|\mathbf{z}_k^{(2)}\|})}
\end{equation}
where $\tau$ is a temperature parameter controlling the distribution sharpness of similarities. This loss explicitly encourages embeddings of augmented views of the same feature (positive pairs) to be similar, while embeddings from different features (negative pairs) are pushed apart. Specifically, minimizing $\mathcal{L}_\text{ZAYAN-CL}$ directly corresponds to maximizing a lower bound on the mutual information between the two augmented views of each feature, as shown through the InfoNCE formulation in Eq.~\ref{eq3}.
\begin{equation}
\label{eq3}
I(Z^{(1)}; Z^{(2)}) \geq \log(m) - \mathcal{L}_\text{ZAYAN-CL}(\theta)
\end{equation}
Minimizing $\mathcal{L}_\text{ZAYAN-CL}(\theta)$ maximizes a lower bound on $I(Z^{(1)}; Z^{(2)})$, encouraging robust and invariant feature representations. Unlike sample-level contrastive learning, ZAYAN-CL applies InfoNCE at the feature level, ensuring embeddings retain maximal information across augmented views and remain resilient to stochastic perturbations. To explicitly discourage redundancy and encourage diversity among feature embeddings, we introduce a redundancy penalty based on pairwise cosine similarity between embeddings in Eq.~\ref{eq4}.
\begin{equation}
\label{eq4}
\mathcal{R}(\theta) = \sum_{i \neq j} \left(\frac{\mathbf{z}_i^\top \mathbf{z}_j}{\|\mathbf{z}_i\|\|\mathbf{z}_j\|}\right)^2
\end{equation}
Eq.~\ref{eq4} directly penalizes the squared cosine similarity between each pair of distinct embeddings, explicitly encouraging them to become orthogonal or maximally dissimilar in angular space. However, the direct summation involves a computational complexity of $\mathcal{O}(m^2)$, which can become costly for larger feature dimensions. To address computational efficiency, we reformulate the redundancy penalty using matrix notation as denoted in Eq.~\ref{eq5}.
\begin{equation}
\label{eq5}
\mathcal{R}(\theta) = \|\mathbf{Z}^\top \mathbf{Z} - \mathbf{I}\|_F^2
\end{equation}
where $\mathbf{Z} = [\mathbf{z}_1, \dots, \mathbf{z}_m] \in \mathbb{R}^{d \times m}$ is the matrix of feature embeddings and $\mathbf{I}$ is the identity matrix of size $m \times m$. This matrix form succinctly enforces orthogonality among feature embeddings, pushing the cross-correlation terms in the matrix $\mathbf{Z}^\top \mathbf{Z}$ towards zero while maintaining unit magnitude embeddings along the diagonal. The matrix-based formulation clarifies the geometric constraint and is preferred for its computational efficiency, especially in high-dimensional settings. After pretraining, we encode individual scalar features for a sample $\mathbf{x}_i = [x_{i1}, \dots, x_{im}]$ in Eq.~\ref{eq6}:
\begin{equation}
\label{eq6}
\mathbf{h}_i = \text{concat}(f_\theta(x_{i1}), \dots, f_\theta(x_{im})) \in \mathbb{R}^{m d}
\end{equation}
Geometrically, ZAYAN-CL maps each feature $\mathbf{f}_j$ to an embedding $\mathbf{z}_j$ on the unit hypersphere in $\mathbb{R}^d$, using cosine similarity $\text{sim}(\mathbf{z}_i, \mathbf{z}_j) = \cos(\theta_{ij})$ to capture angular relations. Minimizing the contrastive loss reduces angles between positive pairs and increases separation among negatives, promoting normalized, diverse, and discriminative embeddings. Algorithm~\ref{algo:zayan-cl} refers to the pseudocode for ZAYAN-CL. \\
\begin{algorithm}[t]
\caption{ZAYAN-CL: Feature-Level Contrastive Learning}
\label{algo:zayan-cl}
\begin{algorithmic}[1]
\small
\Require $X\in\mathbb{R}^{N\times m}$, encoder $f_\theta$, epochs $E$, temperature $\tau$, learning rate $\eta$, redundancy weight $\lambda$
\For{$e=1$ to $E$}
  \For{$j=1$ to $m$}
    \State $(\tilde{\mathbf f}_j^{(1)},\tilde{\mathbf f}_j^{(2)})\gets \mathrm{Augment}(\mathbf f_j)$
    \State $\mathbf z_j^{(k)}\gets \mathrm{norm}(f_\theta(\tilde{\mathbf f}_j^{(k)})),\;\; k\in\{1,2\}$
  \EndFor
  \State $S_{jk}\gets \mathbf z_j^{(1)\top}\mathbf z_k^{(2)}/\tau$
  \State $\mathcal L_{\mathrm{CL}}\gets-\sum_{j=1}^m \log \frac{\exp(S_{jj})}{\sum_{k\ne j}\exp(S_{jk})}$
  \State $Z\gets[\mathbf z_1^{(1)},\dots,\mathbf z_m^{(1)}],\;\; \mathcal R\gets\|Z^\top Z-\mathbf I\|_F^2$
  \State $\theta\gets\theta-\eta\nabla_\theta(\mathcal L_{\mathrm{CL}}+\lambda\mathcal R)$
\EndFor
\Ensure $f_\theta$
\end{algorithmic}
\end{algorithm}
\textbf{B. ZAYAN-T:}
After contrastive pretraining with ZAYAN-CL, we obtain a set of disentangled and normalized feature embeddings $\mathbf{Z}_i = [\mathbf{z}_{i1}, \dots, \mathbf{z}_{im}] \in \mathbb{R}^{m \times d}$ for each sample $\mathbf{x}_i$. These embeddings are input to a downstream Transformer-based classifier that respects the angular structure and independence induced by ZAYAN-CL. We treat $\{\mathbf{z}_{ij}\}_{j=1}^m$ as a sequence of tokens. To inject feature position and contrastive structure, we add a ZAYAN-aware positional encoding $\mathbf{p}_j$ to each embedding as Eq.~\ref{eq7} where $\mathbf{Z}_i^\text{in} = [\tilde{\mathbf{z}}_{i1}, \dots, \tilde{\mathbf{z}}_{im}]$. We pass this sequence through a Transformer encoder in Eq.~\ref{eq8}.
\begin{gather}
\tilde{\mathbf{z}}_{ij} = \mathbf{z}_{ij} + \mathbf{p}_j
\label{eq7}\\[-2pt]
\mathbf{H}_i = \text{TransformerEncoder}(\mathbf{Z}_i^\text{in}) \in \mathbb{R}^{m \times d}
\label{eq8}\\[-2pt]
\mathbf{r}_i = \frac{1}{m} \sum_{j=1}^m \mathbf{h}_{ij}
\label{eq_pooling}
\end{gather}
The final representation $\mathbf{r}_i$ is computed by averaging the Transformer outputs across features (Eq.~\ref{eq_pooling}). This pooled vector is then fed into the MLP to obtain the predicted output $\hat{y}_i$ as defined in Eq.~\ref{eq9}. The total training loss combines cross-entropy classification loss with a ZAYAN-preserving penalty as Eq.~\ref{eq10} where $\mathcal{L}_{\text{CE}}$ is the standard cross-entropy loss (Eq.~\ref{eq11}) and $\mathcal{L}_{\text{preserve}}$ ensures the downstream transformer preserves the structure of ZAYAN-CL embeddings (Eq.~\ref{eq12}) where $\mathbf{h}_{ij}$ is the transformer output token corresponding to feature $j$. ZAYAN-T allows task-aware fine-tuning while preserving the angular and independence constraints learned during pretraining by ZAYAN-CL. The transformer backbone further enables feature interactions and contextual reasoning, making ZAYAN applicable to supervised tasks such as remote sensing classification precisely.
\begin{align}
\hat{y}_i &= \text{MLP}(\text{Aggregate}(\mathbf{H}_i)) \label{eq9}\\[-2pt]
\mathcal{L}_{\text{total}} &= \mathcal{L}_{\text{CE}} + \gamma \cdot \mathcal{L}_{\text{preserve}} \label{eq10}\\[-2pt]
\mathcal{L}_{\text{CE}} &= - \sum_{i=1}^N y_i \log(\hat{y}_i) \label{eq11}\\[-2pt]
\mathcal{L}_{\text{preserve}} &= \sum_{i=1}^N \sum_{j=1}^m \| \mathbf{z}_{ij} - \mathbf{h}_{ij} \|^2 \label{eq12}
\end{align}
\begin{algorithm}[t]
\caption{ZAYAN: ZAYAN-CL Pretraining and ZAYAN-T Fine-Tuning}
\label{alg:zayan}
\begin{algorithmic}[1]
\small
\Require Data $\{(x_i,y_i)\}_{i=1}^N$, encoder $f_\theta$, Transformer $g_\phi$, positional embeddings $\{p_j\}_{j=1}^m$, epochs $E_{\mathrm{T}}$, weight $\gamma$

\State Run Alg.~\ref{algo:zayan-cl} on $X\in\mathbb{R}^{N\times m}$ to obtain frozen feature embeddings $Z=[z_1,\dots,z_m]^\top$
\For{$e=1$ to $E_{\mathrm{T}}$}
  \For{minibatch $\mathcal{B}$}
    \ForAll{$i\in\mathcal{B}$}
      \State $\mathbf{H}_i \gets g_\phi([z_1+p_1,\dots,z_m+p_m])$, \quad
             $\mathbf{r}_i \gets \frac{1}{m}\sum_{j=1}^m \mathbf{h}_{ij}$, \quad
             $\hat y_i \gets \mathrm{MLP}_\phi(\mathbf{r}_i)$
    \EndFor
    \State $\mathcal{L} \gets \mathrm{CE}(\{(\hat y_i,y_i)\}_{i\in\mathcal{B}}) + \gamma \sum_{i\in\mathcal{B}}\sum_{j=1}^m \|\mathbf{h}_{ij}-z_j\|^2$
    \State Update $\phi$ (optionally $\theta$) using $\nabla \mathcal{L}$
  \EndFor
\EndFor
\Ensure Trained model $(f_\theta,g_\phi,Z)$
\end{algorithmic}
\end{algorithm}
\section{Experimental Results}
\label{exp}
\textbf{A. Datasets:} We benchmark our framework on eight datasets spanning diverse domains: Urban Land Cover (675 samples, 147 features; 9‑class classification) \citep{urban} and Forest Type Mapping (523 samples, 27 features; 4‑class classification) \citep{forest}; a 3000‑sample subsample of Crop Mapping (325834 original rows, 174 features; 7‑class classification) \citep{crop}; Wilt (4839 samples, 5 features; binary classification) \citep{wilt}; Flood Risk in India (10000 samples, 21 features; binary classification) \citep{indianflood}; Pluvial Flood (144401 samples, 9 features; 5‑way classification with 50\% injected noise) \citep{pluvial}; Satellite Image Classification aka RSI-CB256 (5631 images, 2048‑dim ResNet \cite{resnet} embeddings PCA→512 \cite{pca}; 4‑way classification) \citep{Satellit17:online}; and Census of Individual Trees (65324 samples, 10 features; binary classification with added synthetic balancing) \citep{tree}. These benchmarks collectively cover a range of sample sizes, feature dimensionalities, noise levels, and modality origins, making them perfect for validating ZAYAN’s ability to learn robust, redundancy-minimized embeddings across heterogeneous environmental and remote-sensing tabular data. See supplementary material for dataset sources. \newline
\textbf{B. Baseline Models:} We compare ZAYAN against a comprehensive suite of baselines. 
First, we include traditional learners such as logistic regression, random forest, SVM, decision tree, KNN, and naive Bayes, as well as boosted tree ensembles AdaBoost~\citep{freund1997decision}, GBM~\citep{friedman2001greedy}, XGBoost~\citep{b45}, LightGBM~\citep{lgbm}, and CatBoost~\citep{catboost}. We further evaluate deep architectures including MLP, 1D CNN, DeepFM~\cite{deepfm}, TabNet~\cite{b1}, FT-Transformer~\cite{b5}, TabTransformer~\citep{b3}, SAINT~\citep{saint}, DCN~\citep{dcn}, AutoInt~\citep{b40}, the sequence-dependent TabSeq~\citep{b6}, and the regularization-based TANGOS~\citep{b42}. Our benchmark also covers probabilistic and tree-based methods such as TabPFN~\citep{b4}, TabPFN v2~\citep{tabpfnv2}, NODE~\citep{b2}, as well as recent state-of-the-art tabular models TabR~\citep{tabr}, TabM~\citep{tabm}, TabICL~\citep{tabicl}, TANDEM~\citep{tandem}, the RNN-based TabulaRNN~\citep{mamb}, and the prototype-guided ProtoGate~\citep{protogate}. Together, these baselines span classical, ensemble, deep, probabilistic, and modern foundation-style tabular methods, providing a strong and diverse benchmark to rigorously validate ZAYAN’s advantages. \newline
\textbf{C. Evaluation Metrics:} We evaluate all models via 5-fold cross-validation, reporting mean accuracy $\pm$ standard deviation on each dataset. To compare methods across datasets, we compute each model’s average rank~\cite{fried} over the eight benchmarks and select the top 10 models by rank. For these, we calculate the Critical Difference (CD) score and display a CD diagram~\cite{cd}, and perform the Friedman test with Nemenyi post-hoc analysis~\cite{nemen}, summarizing pairwise significance levels in a heatmap. In addition, for the same top-10 models, we plot Dolan-Moré performance profiles \citep{dolan} to assess how often each method is close to the per-dataset best, and visualize the distribution of per-dataset accuracies via classification accuracy boxplots. \newline
\textbf{D. Implementation and Training Details:} 
ZAYAN and all baseline models that required hyperparameter tuning (tree ensembles, deep tabular architectures, and sequence models) were tuned using Optuna~\citep{optuna} with 150 trials per dataset, following the standard tabular deep learning practice of Gorishniy et al.~\citep{tabm,tabr,embed,b5}, and evaluated via 5-fold cross-validation, reporting mean$\pm$standard deviation accuracy. TabPFN and TabPFN v2 were used with their recommended default configurations from the public implementations. For example, on the Urban Land Cover dataset, Optuna selected learning rates $cl\_lr=2.06\times10^{-3}$ and $t\_lr=4.23\times10^{-4}$, moderate weight decay ($\approx 10^{-4}$–$10^{-3}$), embedding dimension $128$, hidden dimension $512$, $8$ attention heads, $4$ transformer layers, batch size $32$, and dropout $\approx 0.10$ (pretraining) / $0.40$ (fine-tuning). Full search ranges and per-dataset best configurations are summarized in supplementary material. All experiments were implemented in PyTorch with CUDA and AMP, using standard CPU/GPU hardware.\newline
\textbf{E. Computational Complexity:} 
Let $N$ denote the number of samples, $m$ the number of features, and $d$ the embedding dimension. 
ZAYAN-CL runs in $\mathcal{O}(Nm + m^2 d)$ time and $\mathcal{O}(Nm + m^2)$ memory, where the quadratic $m^2$ term arises from the similarity matrix and redundancy penalties. ZAYAN-T adds $\mathcal{O}(Nm^2 d)$ time for self-attention and $\mathcal{O}(Nmd)$ for the structure-preserving loss, both using $\mathcal{O}(md)$ space. In practice, single-run training times range from 22.8\,s (RSI-CB256) to 2354.8\,s (Pluvial Flood) across our benchmarks, and sparse or approximate attention mechanisms could further reduce the $m^2$ overhead for very wide feature sets.\newline
\begin{table}[t]
\centering
\setlength{\tabcolsep}{2.0pt}
\renewcommand{\arraystretch}{0.95}
\footnotesize
\caption{
Mean$_{\pm\text{std}}$ 5-fold CV accuracy (\%) for all models across 8 datasets under the standard average-tie rule. Here, $^{\$}$: 50\% noise added to training data; $^{*}$: 3000-sample subsample from the original dataset; $^{+}$: synthetic samples added to balance classes and increase difficulty; $^{\#}$: ResNet image features PCA-downsampled to 512-D. TabPFN is limited to datasets with $\leq 1000$ samples and $\leq 100$ features; TabPFN v2 supports $\leq 10{,}000$ samples and $\leq 500$ features.
}
\label{tab:cv_results}
\resizebox{\textwidth}{!}{%
\begin{tabular}{lccccccccc}
\toprule
\textbf{Model} & \textbf{Urban} & \textbf{Wilt} & \textbf{Crop$^{*}$} & \textbf{Forest} & \textbf{Tree$^{+}$} & \textbf{RSI$^{\#}$} & \textbf{Pluvial$^{\$}$} & \textbf{Indian} & \textbf{Avg. Rank} \\
\midrule
Log. Reg. & \ms{70.61}{1.06} & \ms{71.40}{0.80} & \ms{98.93}{0.91} & \ms{75.30}{1.04} & \ms{63.02}{0.46} & \ms{99.70}{0.13} & \ms{82.07}{1.12} & \ms{47.70}{1.22} & \ms{16.38}{7.43} \\
Random Forest & \ms{79.29}{1.22} & \ms{78.00}{1.14} & \ms{99.50}{0.28} & \ms{77.50}{1.52} & \ms{57.61}{0.11} & \ms{99.34}{0.27} & \ms{86.45}{0.85} & \ms{50.30}{0.93} & \ms{14.00}{6.58} \\
SVM & \ms{68.44}{1.93} & \ms{63.40}{1.27} & \ms{99.18}{0.82} & \ms{71.40}{0.67} & \ms{58.72}{0.33} & \ms{98.51}{0.38} & \ms{74.90}{1.30} & \ms{49.40}{1.15} & \ms{22.31}{6.64} \\
Decision Tree & \ms{74.75}{1.40} & \ms{80.40}{0.57} & \ms{98.71}{0.88} & \ms{72.00}{1.43} & \ms{57.79}{0.05} & \ms{97.35}{0.43} & \ms{70.41}{1.42} & \ms{51.10}{1.17} & \ms{18.56}{8.51} \\
KNN & \ms{70.22}{0.73} & \ms{66.40}{1.41} & \bms{99.66}{0.34} & \ms{77.20}{1.00} & \ms{66.77}{0.49} & \ms{98.85}{0.40} & \ms{76.70}{1.18} & \ms{50.40}{0.96} & \ms{13.94}{6.75} \\
Naive Bayes & \ms{77.51}{0.73} & \ms{67.60}{0.76} & \ms{91.58}{0.79} & \ms{71.40}{0.60} & \ms{64.71}{0.40} & \ms{88.17}{1.09} & \ms{62.81}{1.20} & \ms{48.80}{1.05} & \ms{22.81}{8.25} \\
MLP & \ms{69.63}{0.59} & \ms{86.00}{0.60} & \ms{96.67}{0.57} & \ms{76.00}{0.97} & \ms{61.50}{0.33} & \ms{99.54}{0.15} & \ms{89.14}{0.98} & \ms{50.47}{0.89} & \ms{14.13}{4.97} \\
CatBoost \citep{catboost} & \ms{80.08}{1.25} & \ms{83.40}{2.32} & \ms{99.46}{0.15} & \ms{77.90}{1.63} & \ms{58.80}{0.11} & \ms{99.50}{0.07} & \ms{89.87}{0.67} & \ms{49.13}{0.78} & \ms{12.13}{7.21} \\
AdaBoost \citep{freund1997decision} & \ms{63.12}{1.40} & \ms{65.20}{1.95} & \ms{92.46}{1.08} & \ms{79.40}{1.95} & \ms{65.04}{0.20} & \ms{91.81}{1.84} & \ms{89.87}{1.34} & \ms{50.33}{1.04} & \ms{18.25}{9.00} \\
XGBoost \citep{b45} & \ms{68.05}{2.03} & \ms{84.40}{2.25} & \ms{99.57}{0.33} & \ms{76.00}{2.13} & \ms{58.41}{0.18} & \ms{99.43}{0.16} & \ms{89.87}{0.71} & \ms{51.53}{0.83} & \ms{12.06}{8.44} \\
LGBM \citep{lgbm} & \ms{69.26}{0.53} & \ms{62.80}{0.96} & \ms{97.21}{1.74} & \ms{74.15}{1.83} & \ms{58.56}{0.07} & \ms{99.54}{0.10} & \ms{84.93}{0.66} & \ms{48.70}{0.91} & \ms{21.00}{7.08} \\
GBM \citep{friedman2001greedy} & \ms{72.78}{1.95} & \ms{81.00}{0.65} & \ms{96.25}{1.04} & \ms{77.90}{1.21} & \ms{66.96}{0.19} & \ms{99.36}{0.12} & \ms{89.85}{0.79} & \ms{50.87}{0.88} & \ms{11.94}{5.55} \\
1D CNN & \ms{73.37}{1.75} & \ms{63.20}{1.53} & \ms{98.24}{0.92} & \ms{75.40}{0.68} & \ms{60.78}{1.13} & \ms{99.66}{0.13} & \ms{87.11}{0.94} & \ms{50.53}{0.77} & \ms{15.25}{6.80} \\
TabNet \citep{b1} & \ms{64.10}{1.16} & \ms{63.60}{1.31} & \ms{96.48}{0.82} & \ms{78.20}{1.57} & \ms{66.66}{0.12} & \ms{98.45}{0.12} & \ms{86.89}{1.50} & \ms{50.73}{1.13} & \ms{17.00}{7.18} \\
FT-T \citep{b5} & \ms{73.18}{0.77} & \ms{70.00}{0.68} & \ms{97.20}{0.71} & \ms{72.30}{1.64} & \ms{66.84}{0.15} & \ms{98.20}{0.15} & \ms{88.44}{0.85} & \ms{50.27}{0.79} & \ms{16.38}{5.93} \\
TabT \citep{b3} & \ms{74.75}{0.78} & \ms{64.40}{1.24} & \ms{97.10}{1.70} & \ms{75.10}{1.34} & \ms{62.96}{0.30} & \ms{97.86}{0.24} & \ms{79.78}{0.98} & \ms{50.00}{0.92} & \ms{19.31}{4.01} \\
TabSeq \citep{b6} & \ms{73.96}{0.96} & \ms{63.40}{0.55} & \ms{98.68}{0.61} & \ms{74.50}{1.66} & \ms{64.48}{0.12} & \ms{98.58}{0.10} & \ms{40.67}{1.64} & \ms{49.47}{1.18} & \ms{19.56}{7.06} \\
SAINT \citep{saint} & \ms{63.50}{1.29} & \ms{66.80}{1.86} & \ms{95.38}{1.98} & \ms{79.10}{1.24} & \ms{63.50}{0.20} & \ms{99.20}{0.12} & \ms{86.92}{0.97} & \ms{48.20}{1.05} & \ms{19.13}{7.18} \\
DeepFM \citep{deepfm} & \ms{68.20}{1.15} & \ms{63.60}{0.89} & \ms{97.58}{1.66} & \ms{68.00}{1.28} & \ms{58.92}{0.16} & \ms{98.48}{0.15} & \ms{84.90}{0.86} & \ms{49.62}{0.99} & \ms{22.00}{4.61} \\
DCN \citep{dcn} & \ms{71.60}{0.94} & \ms{63.60}{1.49} & \ms{96.26}{0.80} & \ms{73.20}{1.14} & \ms{64.12}{0.10} & \ms{97.26}{0.25} & \ms{78.68}{0.88} & \ms{48.69}{0.91} & \ms{22.38}{4.79} \\
AutoInt \citep{b40} & \ms{69.00}{1.42} & \ms{66.20}{0.97} & \ms{97.21}{0.51} & \ms{72.30}{0.54} & \ms{59.28}{0.16} & \ms{96.48}{0.15} & \ms{82.76}{1.21} & \ms{49.72}{1.12} & \ms{21.63}{2.97} \\
TabPFN \citep{b4} & - & - & - & \ms{85.80}{0.66} & - & - & - & - & \ms{28.50}{8.14} \\
TabPFNv2 \citep{tabpfnv2} & \ms{80.06}{2.00} & \ms{88.20}{1.93} & \ms{92.56}{1.17} & \ms{95.66}{2.42} & - & - & - & \ms{51.52}{0.84} & \ms{17.50}{12.69} \\
TANGOS \citep{b42} & \ms{66.37}{1.05} & \ms{64.60}{0.78} & \ms{63.20}{1.59} & \ms{74.77}{1.45} & \ms{56.98}{0.18} & \ms{91.20}{0.32} & \ms{73.56}{1.13} & \ms{51.00}{0.95} & \ms{24.00}{7.12} \\
NODE~\citep{b2} & \ms{72.46}{1.18} & \ms{82.30}{1.95} & \ms{90.54}{1.66} & \ms{75.20}{0.97} & \ms{62.56}{0.24} & \ms{96.24}{0.12} & \ms{90.65}{0.72} & \ms{50.95}{0.69} & \ms{16.63}{8.28} \\
TabR \citep{tabr} & \ms{83.21}{1.66} & \ms{98.78}{0.17} & \ms{98.62}{0.68} & \ms{90.25}{1.94} & \ms{64.78}{0.42} & \ms{99.32}{0.18} & \ms{89.88}{0.62} & \ms{49.59}{1.88} & \ms{9.25}{6.53} \\
TabM \citep{tabm} & \ms{82.07}{0.98} & \ms{98.80}{0.20} & \ms{98.92}{0.96} & \ms{88.72}{0.98} & \ms{65.34}{0.24} & \ms{99.36}{0.10} & \ms{89.64}{0.98} & \ms{51.49}{0.92} & \ms{6.94}{2.96} \\
TabICL \citep{tabicl} & \ms{83.30}{0.98} & \ms{98.76}{0.17} & \ms{99.46}{0.10} & \ms{90.05}{2.35} & \ms{66.98}{0.20} & \ms{99.74}{0.16} & \ms{90.75}{0.74} & \ms{51.27}{0.66} & \ms{3.38}{1.43} \\
TANDEM \citep{tandem} & \ms{83.29}{1.53} & \ms{98.37}{0.39} & \ms{99.18}{0.12} & \ms{90.05}{2.07} & \ms{66.96}{0.12} & \ms{99.72}{0.14} & \ms{90.72}{0.32} & \ms{50.26}{1.55} & \ms{6.06}{4.77} \\
TabulaRNN \citep{mamb} & \ms{56.00}{4.79} & \ms{98.72}{0.14} & \ms{97.68}{0.66} & \ms{73.02}{7.85} & \ms{57.82}{0.35} & \ms{98.20}{0.24} & \ms{74.92}{1.36} & \ms{50.34}{0.35} & \ms{20.06}{7.78} \\
ProtoGate \citep{protogate} & \ms{62.96}{4.39} & \ms{94.03}{0.62} & \ms{93.20}{0.38} & \ms{38.42}{6.84} & \ms{56.68}{0.38} & \ms{98.16}{1.02} & \ms{70.46}{1.44} & \ms{50.11}{0.56} & \ms{24.13}{7.64} \\
ZAYAN (ours) & \bms{84.80}{7.10} & \bms{99.69}{0.40} & \bms{99.66}{0.34} & \bms{97.21}{0.45} & \bms{67.00}{0.12} & \bms{99.77}{0.15} & \bms{93.61}{0.47} & \bms{51.55}{0.43} & \bms{1.06}{0.17} \\
\bottomrule
\end{tabular}%
}
\end{table}
\begin{table}[t]
  \centering
  \caption{Left: Ablation on Urban Land Cover (5-fold CV mean accuracy$\pm$std). 
           Right: Wilcoxon signed-rank test comparing ZAYAN to the other top-10 models. 
           $\Delta\text{Acc}$ denotes the mean accuracy difference (ZAYAN $-$ baseline) across the 8 datasets, 
           and $p_{\text{Holm}}$ are Holm-adjusted $p$-values. Here, ns = not significant}
  \label{tab:ablation_wilcoxon}
  \scriptsize

  \begin{minipage}[t]{0.47\textwidth}
    \centering
    \setlength{\tabcolsep}{3pt}
    \textbf{(a) Ablation on Urban Land Cover}\\[2pt]
    \begin{tabular}{@{}>{\raggedright\arraybackslash}p{0.60\textwidth} r@{}}
      \toprule
      Configuration & Acc. (\%) \\
      \midrule
      Full ZAYAN & 84.80$\pm$7.10 \\
      \quad w/o redundancy penalty ($\lambda=0$) & 78.45$\pm$0.72 \\
      \quad w/o contrastive loss (MSE only) & 75.98$\pm$0.88 \\
      \quad no quantile warping & 79.30$\pm$0.65 \\
      \quad no masking & 79.75$\pm$0.60 \\
      \quad no preservation loss ($\gamma=0$) & 79.01$\pm$0.70 \\
      \quad embedding dim $d=64$ & 78.12$\pm$0.80 \\
      \quad embedding dim $d=512$ & 80.05$\pm$0.58 \\
      \bottomrule
    \end{tabular}
  \end{minipage}
  \hfill
  \begin{minipage}[t]{0.47\textwidth}
    \centering
    \setlength{\tabcolsep}{4pt}
    \textbf{(b) Wilcoxon vs. top-10 baselines}\\[2pt]
    \begin{tabular}{@{}lrrrrc@{}}
      \toprule
      Baseline & $\Delta\text{Acc}$ & $n_{\text{ds}}$ & $p_{\text{raw}}$ & $p_{\text{Holm}}$ & Sig \\
      \midrule
      TabICL        &  1.62 & 8 & 0.00781 & 0.0703 & ns \\
      TANDEM        &  1.84 & 8 & 0.00781 & 0.0703 & ns \\
      TabM          &  2.37 & 8 & 0.00781 & 0.0703 & ns \\
      TabR          &  2.36 & 8 & 0.00781 & 0.0703 & ns \\
      GBM           &  7.29 & 8 & 0.00781 & 0.0703 & ns \\
      XGBoost       &  8.25 & 8 & 0.00781 & 0.0703 & ns \\
      CatBoost      &  6.89 & 8 & 0.00781 & 0.0703 & ns \\
      Random Forest &  8.16 & 8 & 0.00781 & 0.0703 & ns \\
      KNN           & 10.90 & 8 & 0.01800 & 0.0703 & ns \\
      \bottomrule
    \end{tabular}
  \end{minipage}

\end{table}
\textbf{F. Comparative Analysis:}
Table~\ref{tab:cv_results} reports 5-fold cross-validation mean accuracy (\%) and standard deviation for ZAYAN and 31 baseline models across the eight remote sensing and flood datasets. All models that required hyperparameter tuning were optimized with Optuna~\cite{optuna} for fair comparison. The benchmarks span a range of difficulty and structure, including noise injection (\$), subsampling (*), class balancing with synthetic samples (+), and PCA-compressed image features (\#). Across all datasets, ZAYAN achieves the highest or tied-highest mean accuracy. It is best on Urban (84.80\%), Wilt (99.69\%), Forest Type (97.21\%), Tree Census (67.00\%), RSI-CB256 (99.77\%), Pluvial Flood (93.61\%), and Indian Flood (51.55\%), and ties KNN on Crop Mapping (99.66\%). These consistent per-dataset wins translate into an average rank of $1.06\pm0.17$, substantially better than the next-best models TabICL, TANDEM, TabM, and TabR, whose average ranks are $3.38$, $6.06$, $6.94$, and $9.25$, respectively. In particular, ZAYAN’s margins over strong baselines become more pronounced on the noisier and more imbalanced settings such as Tree Census and Pluvial Flood, where most other methods lose several percentage points of accuracy. Among classical methods, tree ensembles remain the strongest competitors. CatBoost, GBM, and XGBoost reach high accuracies on several datasets (e.g., $>99\%$ on RSI-CB256 and close to $90\%$ on Pluvial Flood), but their average ranks (11.94–12.13) still lag behind the best tabular deep models. Simpler learners such as logistic regression, SVM, KNN, and Naive Bayes can perform competitively on the easier or more linearly separable tasks (e.g., RSI-CB256, Crop Mapping), yet they degrade more sharply on the harder flood datasets. Neural baselines designed for recommender or tabular data (MLP, 1D CNN, DeepFM, DCN, AutoInt, NODE, TabulaRNN, ProtoGate) generally fall between the tree ensembles and the specialized tabular transformers, but none closes the gap to ZAYAN in average rank. Specialized tabular architectures such as TabNet, FT-Transformer, TabTransformer, TabSeq, SAINT, and TANGOS show mixed performance: they occasionally match ensemble methods on individual datasets but typically underperform both ZAYAN and the strongest trees, with average ranks in the 16–24 range. Probabilistic TabPFN performs well on Forest Type (85.80\%) but can only be evaluated on a subset of datasets due to its sample and feature-count restrictions; TabPFN v2 relaxes these constraints and attains strong scores on Urban and Forest Type (80.06\% and 95.66\%), yet still trails ZAYAN and the best transfer-based baselines in both accuracy and average rank ($17.50$). Overall, the table shows that ZAYAN not only reaches near-ceiling performance on the simpler land-cover tasks where many methods perform well, but also maintains clear advantages on the more challenging noisy, imbalanced, and high-dimensional flood scenarios, leading to the most favorable average rank across all models.\newline
\begin{figure*}[htbp]
  \centering
  \begin{subfigure}[b]{0.49\textwidth}
    \centering
    \includegraphics[width=\linewidth]{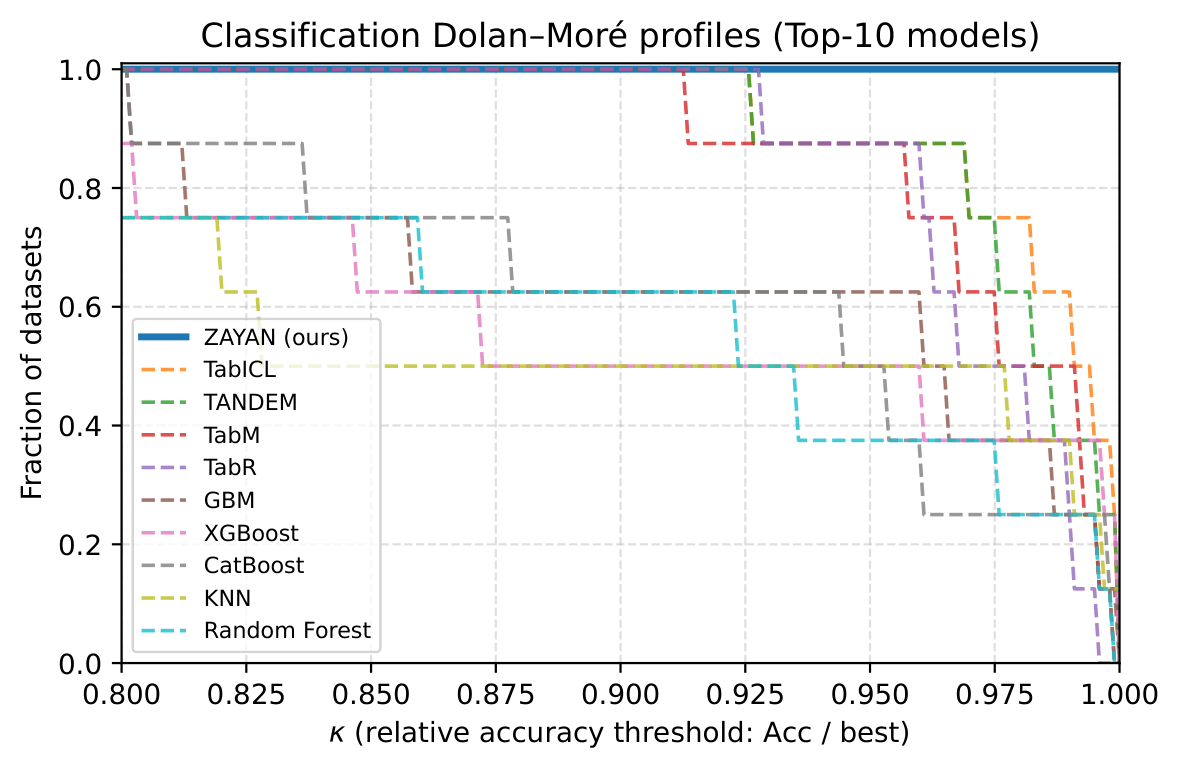}
    \caption{Dolan-Mor\'e profiles (top-10 models).}
    \label{fig:dolan_more}
  \end{subfigure}\hfill
  \begin{subfigure}[b]{0.49\textwidth}
    \centering
    \includegraphics[width=\linewidth]{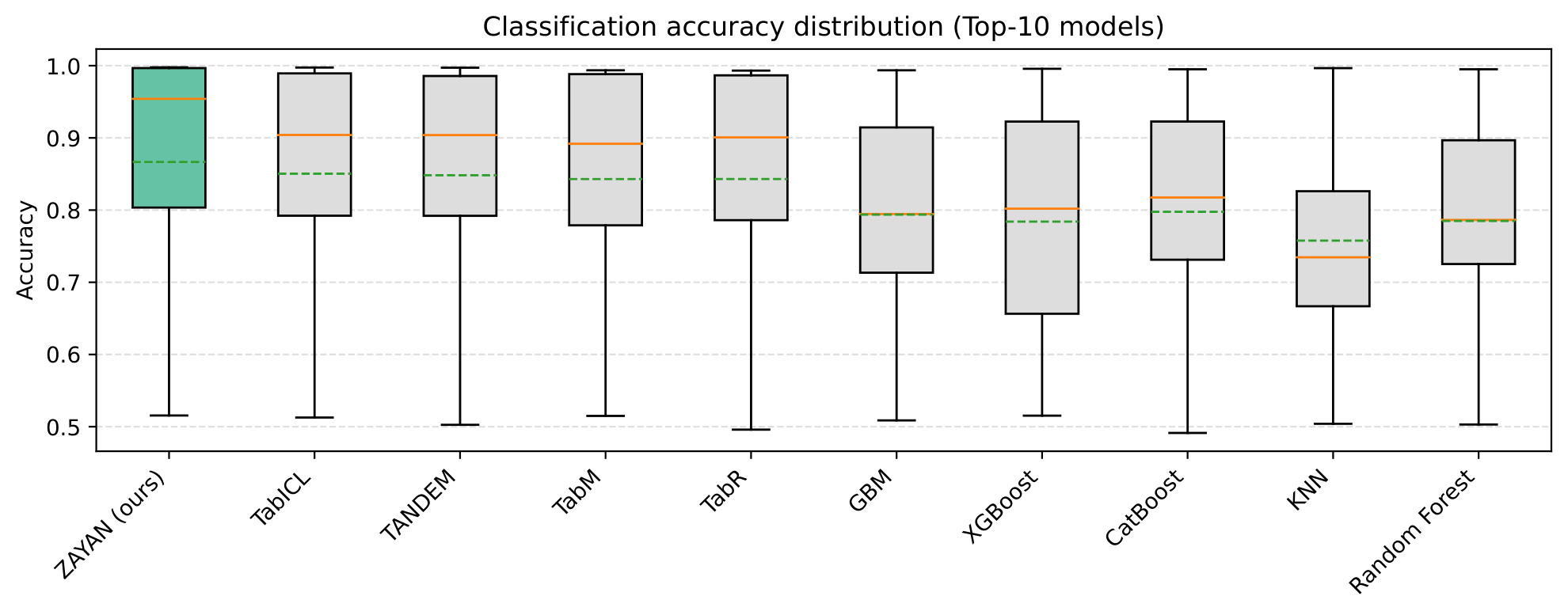}
    \caption{Classification accuracy distribution (top-10 models).}
    \label{fig:acc_boxplot}
  \end{subfigure}
  \vspace{0.4em}
  \begin{subfigure}[b]{0.49\textwidth}
    \centering
    \includegraphics[width=\linewidth]{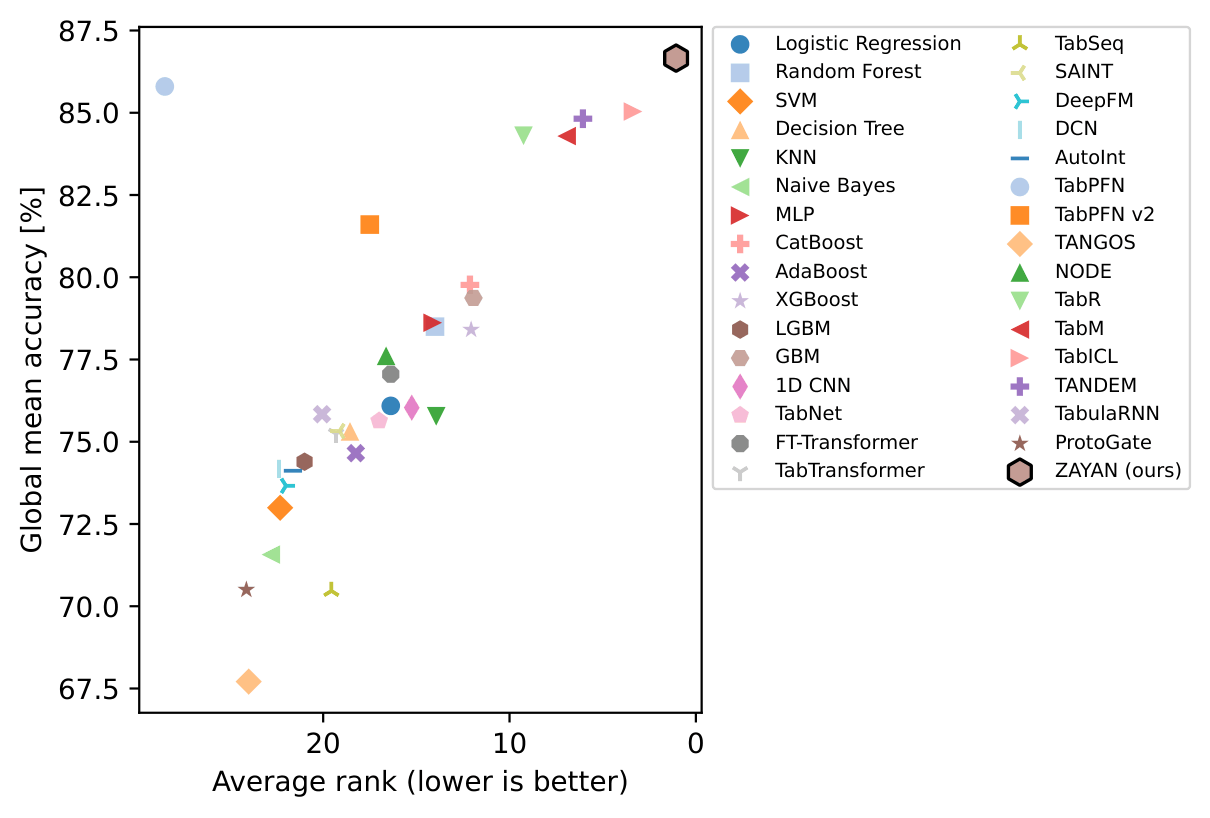}
    \caption{Global mean accuracy vs.\ average rank for all models.}
    \label{fig:rank_vs_accuracy}
  \end{subfigure}\hfill
  \begin{subfigure}[b]{0.49\textwidth}
    \centering
    \includegraphics[width=\linewidth]{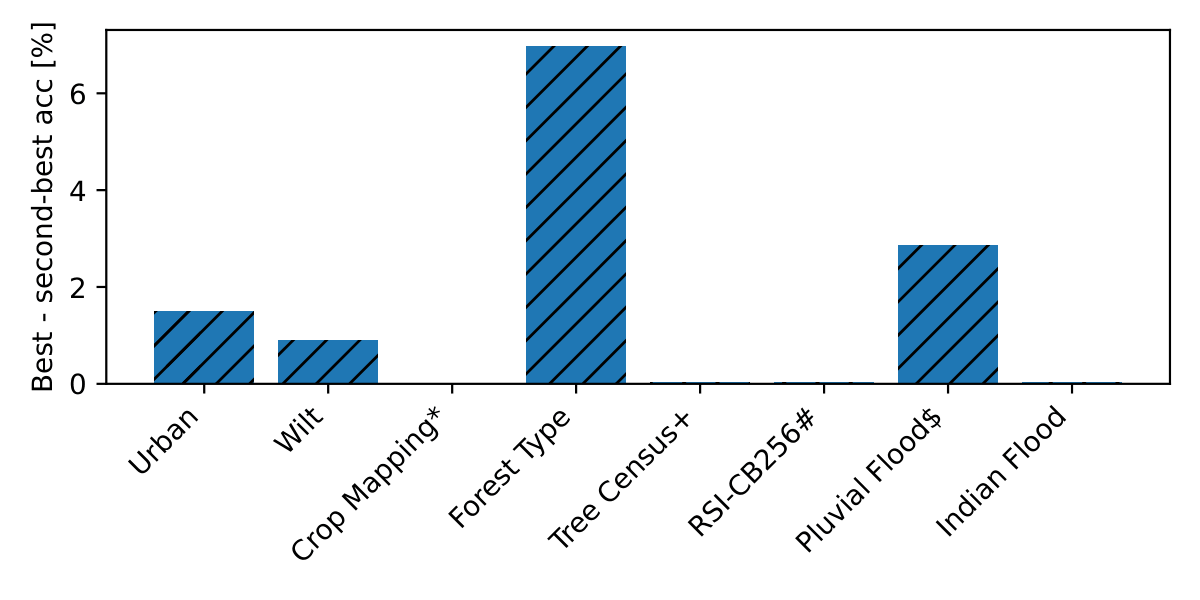}
    \caption{Accuracy margin between best and second-best model per dataset.}
    \label{fig:best_second_margin}
  \end{subfigure}
  \vspace{0.4em}
  \begin{subfigure}[b]{0.49\textwidth}
    \centering
    \includegraphics[width=\linewidth]{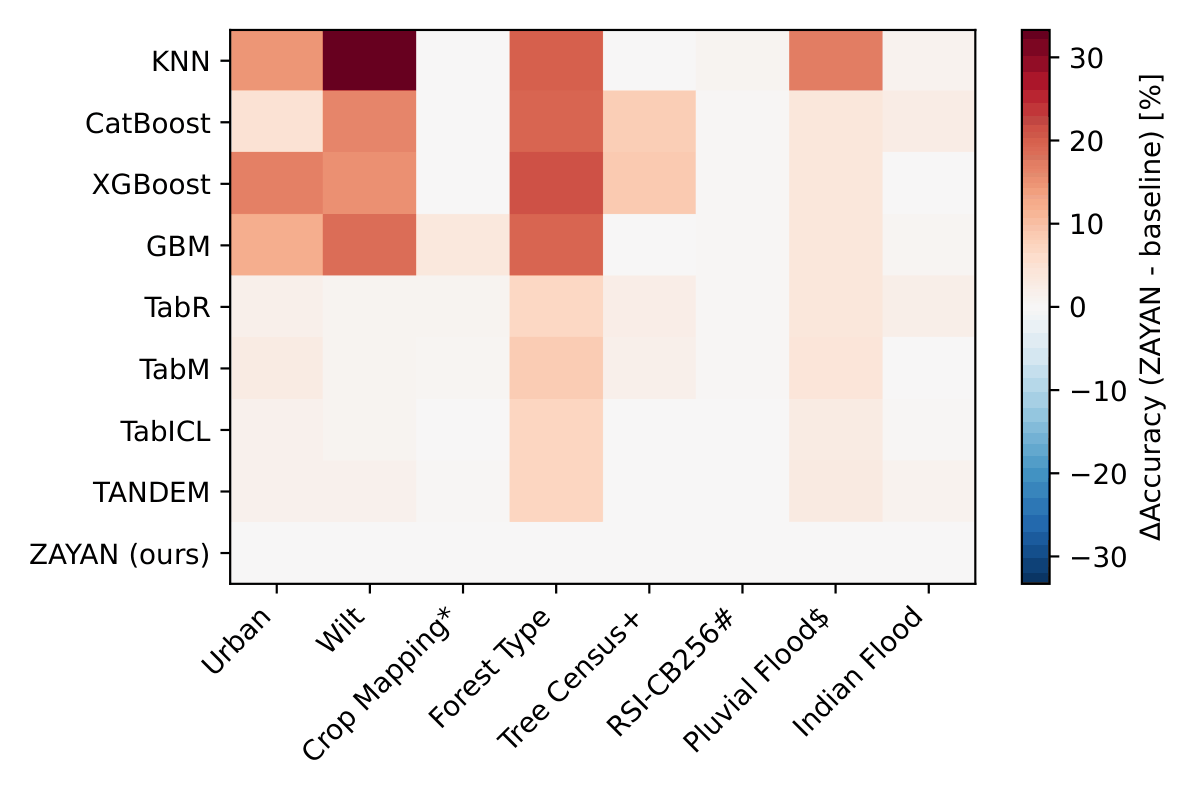}
    \caption{Per-dataset accuracy gains of ZAYAN over other top-10 models.}
    \label{fig:zayan_delta_heatmap}
  \end{subfigure}\hfill
  \begin{subfigure}[b]{0.49\textwidth}
    \centering
    \includegraphics[width=\linewidth]{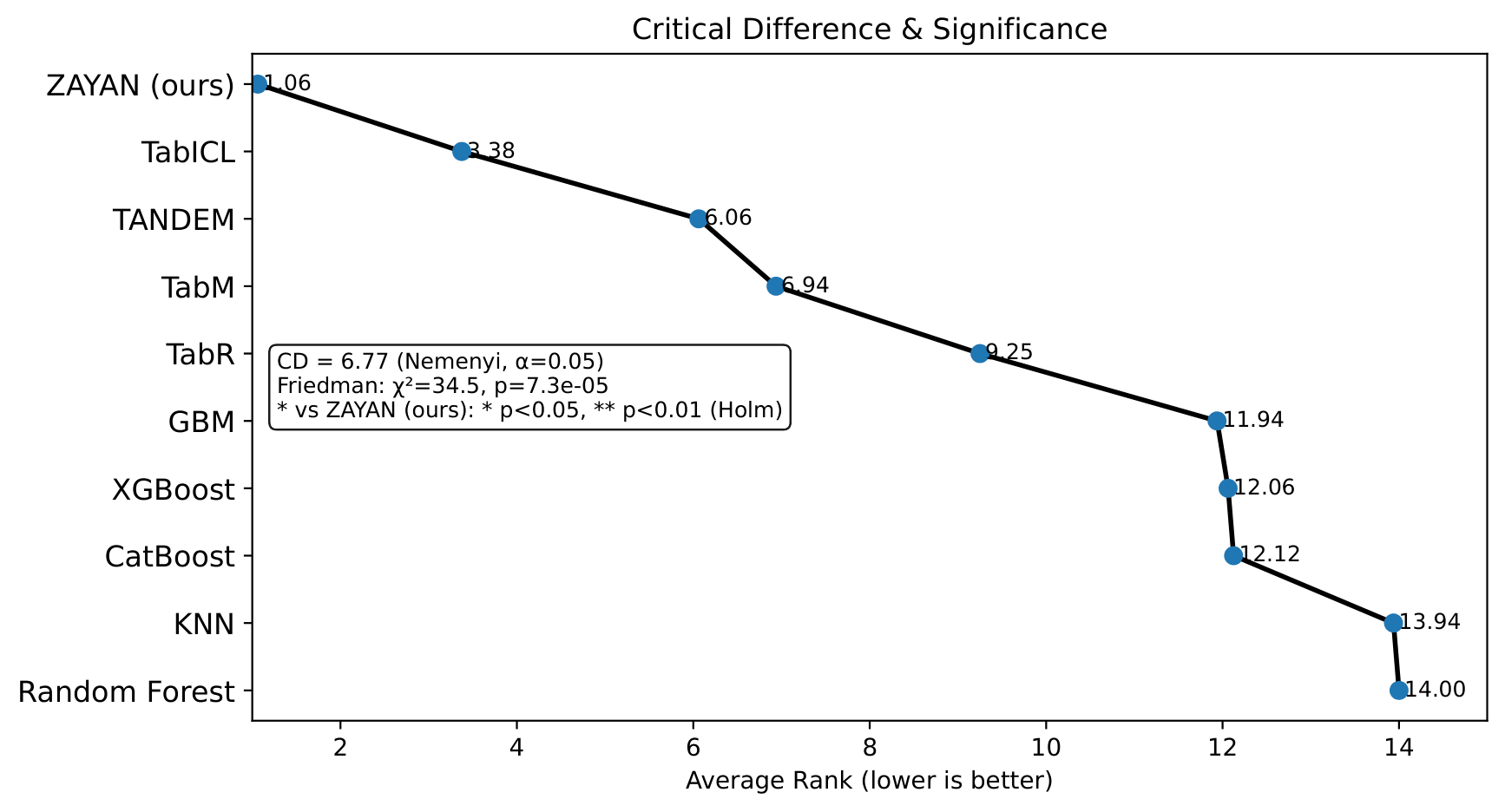}
    \caption{Average ranks of the top-10 models with critical difference summary.}
    \label{fig:zayan_cd_rankline}
  \end{subfigure}
  \caption{
  Comprehensive comparison of ZAYAN and baselines. 
  (a) Dolan-Mor\'e profiles: fraction of datasets where $\mathrm{Acc}/\mathrm{Acc}_{\text{best}} \ge \kappa$.
  (b) Distribution of 5-fold CV accuracies across datasets. 
  (c) Global mean accuracy vs.\ average rank for all models. 
  (d) Per-dataset margin between the best and second-best model. 
  (e) Heatmap of ZAYAN’s per-dataset accuracy gains over the other top-10 models. 
  (f) Average rank plot with critical difference information for the top-10 models.}
  \label{fig:dolan_more_boxplot}
\end{figure*}
\textbf{G. Statistical Significance and Comparative Visualization:}
\label{subsec:stats_viz}
Fig.~\ref{fig:dolan_more_boxplot} and Table~\ref{tab:ablation_wilcoxon}(b) summarize the comparative behaviour of ZAYAN and the strongest baselines.
The Dolan-Mor\'e profiles in Fig.~\ref{fig:dolan_more_boxplot}\subref{fig:dolan_more} plot, for each relative-accuracy threshold $\kappa \in [0,1]$, the fraction of datasets on which a model attains $\mathrm{Acc}/\mathrm{Acc}_{\text{best}} \ge \kappa$. ZAYAN’s curve stays on top over almost the entire range of $\kappa$, indicating that it is both often the best and rarely far from the best model on any dataset. The boxplots in Fig.~\ref{fig:dolan_more_boxplot}\subref{fig:acc_boxplot} show the distribution of 5-fold CV accuracies across datasets: ZAYAN achieves the highest median and upper quartiles, with a relatively tight interquartile range, while several baselines exhibit lower medians and heavier tails. The scatter of global mean accuracy versus average rank in Fig.~\ref{fig:dolan_more_boxplot}\subref{fig:rank_vs_accuracy} places ZAYAN in the top-left corner (best rank and highest global mean), while the margin plot in Fig.~\ref{fig:dolan_more_boxplot}\subref{fig:best_second_margin} and the heatmap in Fig.~\ref{fig:dolan_more_boxplot}\subref{fig:zayan_delta_heatmap} highlight consistent gains over other top-10 competitors. The rank line with CD in Fig.~\ref{fig:dolan_more_boxplot}\subref{fig:zayan_cd_rankline} summarizes the Nemenyi analysis. However, Table~\ref{tab:ablation_wilcoxon}(b) reports Holm-adjusted Wilcoxon $p$-values above 0.05 for all top-10 baselines; thus, while ZAYAN remains the strongest overall model by cross-dataset consistency (profiles and average rank) with uniformly positive gains of 1--11\,pp, confirming these improvements with stronger statistical support would benefit from a larger set of datasets. \newline
\begin{figure*}[t]
  \centering
  \newcommand{\ablimgheight}{2.35cm}

  \begin{subfigure}[t]{0.235\textwidth}
    \centering
    \includegraphics[width=\linewidth,height=\ablimgheight,keepaspectratio]{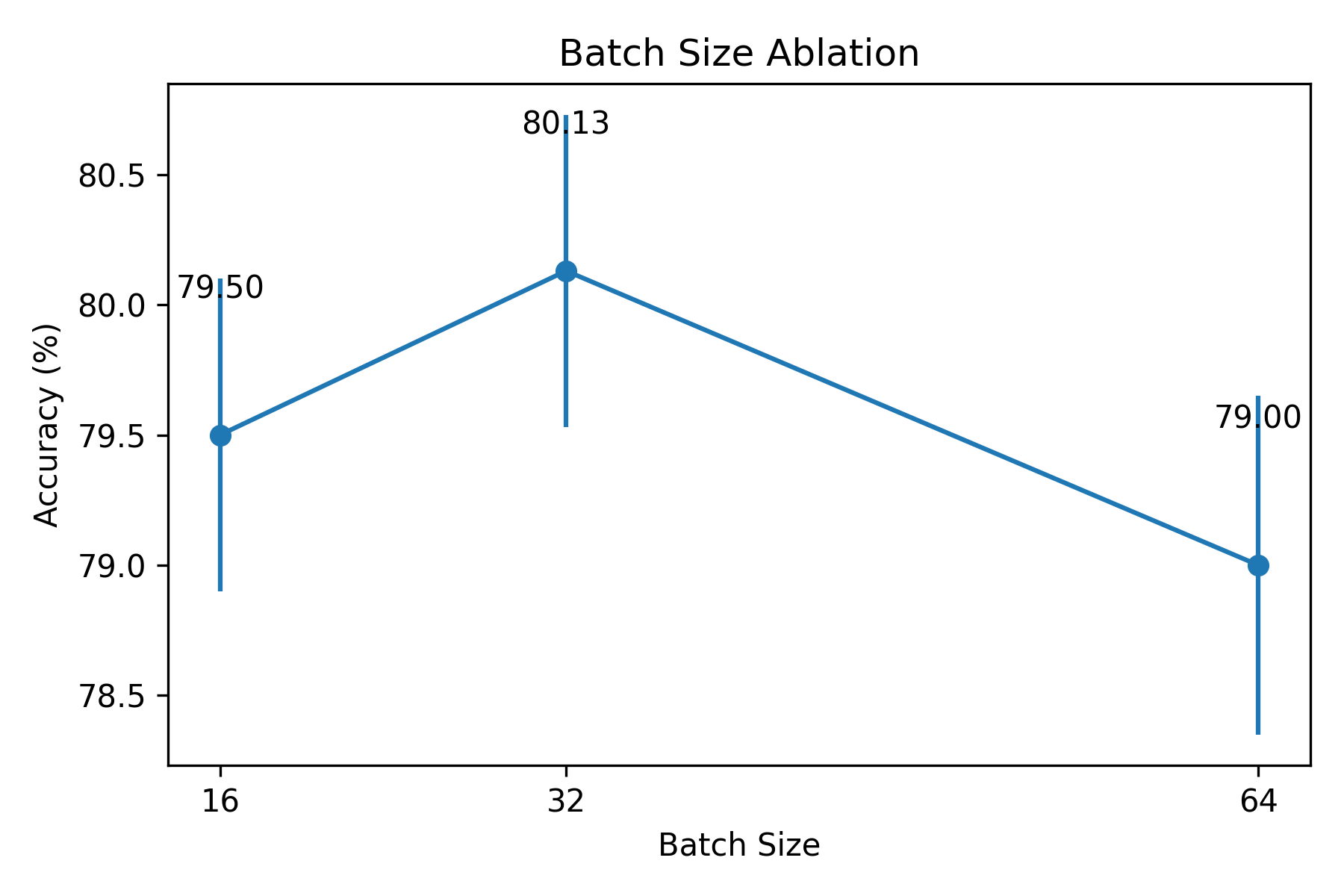}
    \caption{Batch size.}
    \label{fig:abl_batch}
  \end{subfigure}\hfill
  \begin{subfigure}[t]{0.235\textwidth}
    \centering
    \includegraphics[width=\linewidth,height=\ablimgheight,keepaspectratio]{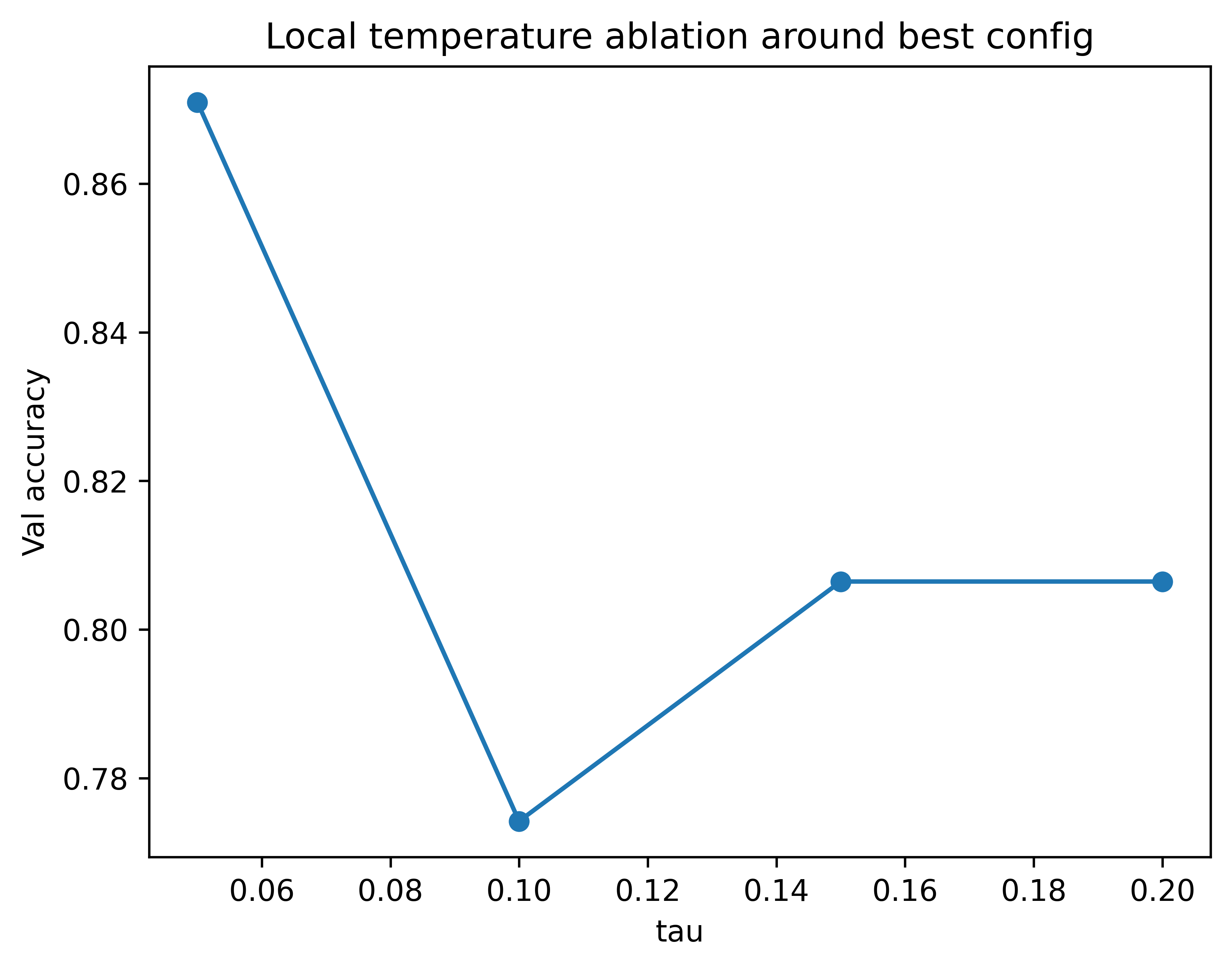}
    \caption{Temperature $\tau$.}
    \label{fig:abl_tau}
  \end{subfigure}\hfill
  \begin{subfigure}[t]{0.235\textwidth}
    \centering
    \includegraphics[width=\linewidth,height=\ablimgheight,keepaspectratio]{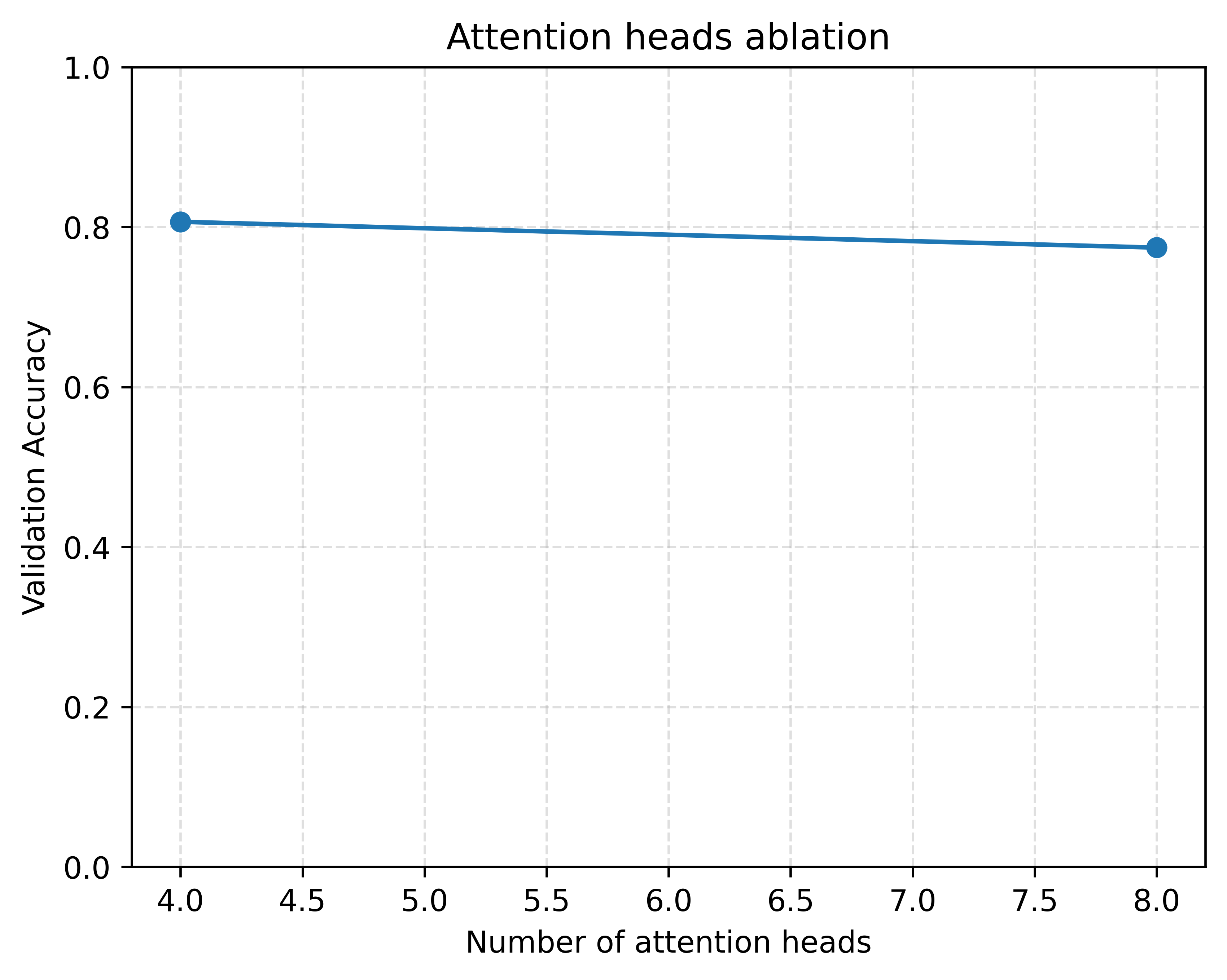}
    \caption{Attention heads.}
    \label{fig:abl_heads}
  \end{subfigure}\hfill
  \begin{subfigure}[t]{0.235\textwidth}
    \centering
    \includegraphics[width=\linewidth,height=\ablimgheight,keepaspectratio]{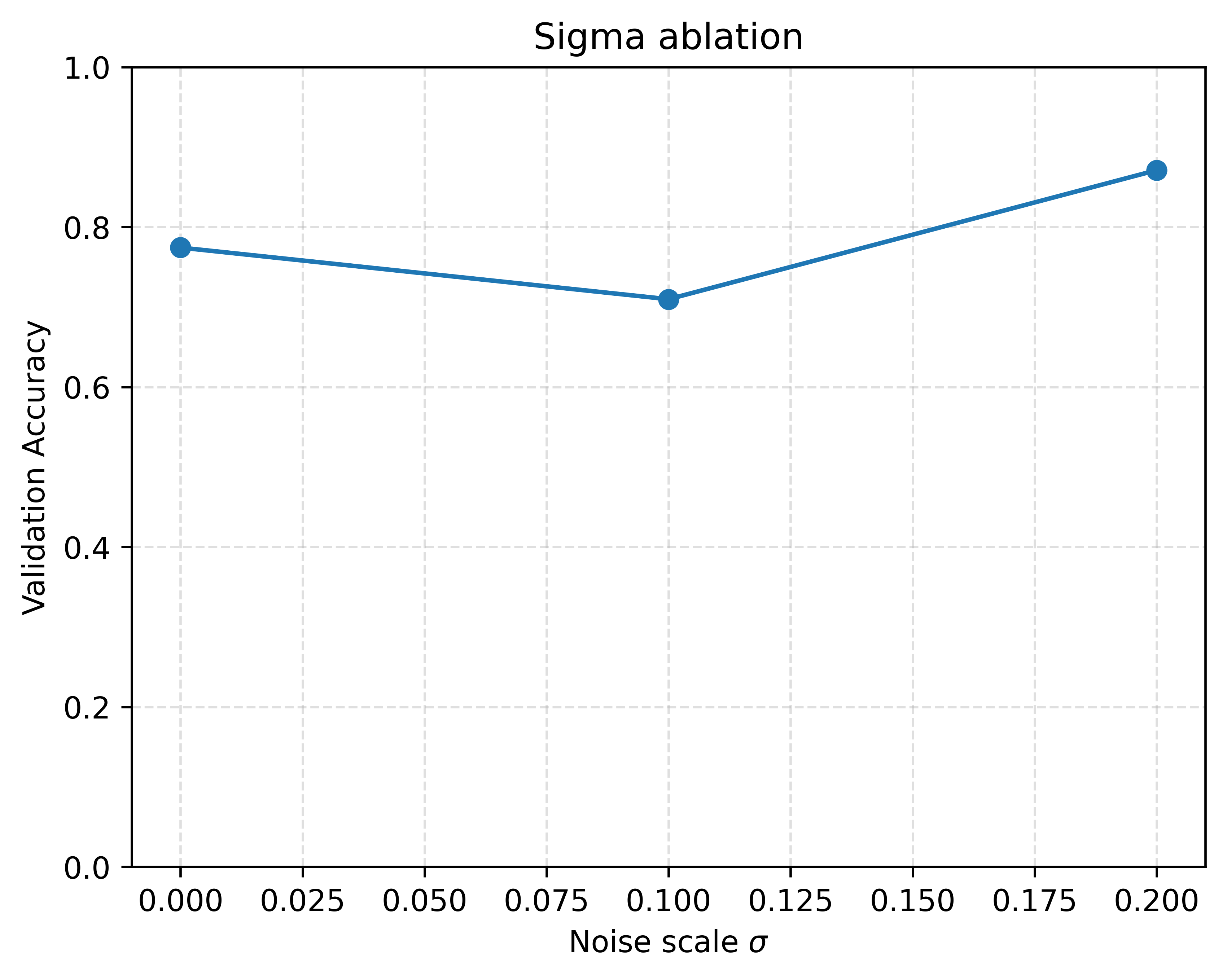}
    \caption{Noise scale $\sigma$.}
    \label{fig:abl_sigma}
  \end{subfigure}

  \vspace{0.25em}

  \begin{subfigure}[t]{0.235\textwidth}
    \centering
    \includegraphics[width=\linewidth,height=\ablimgheight,keepaspectratio]{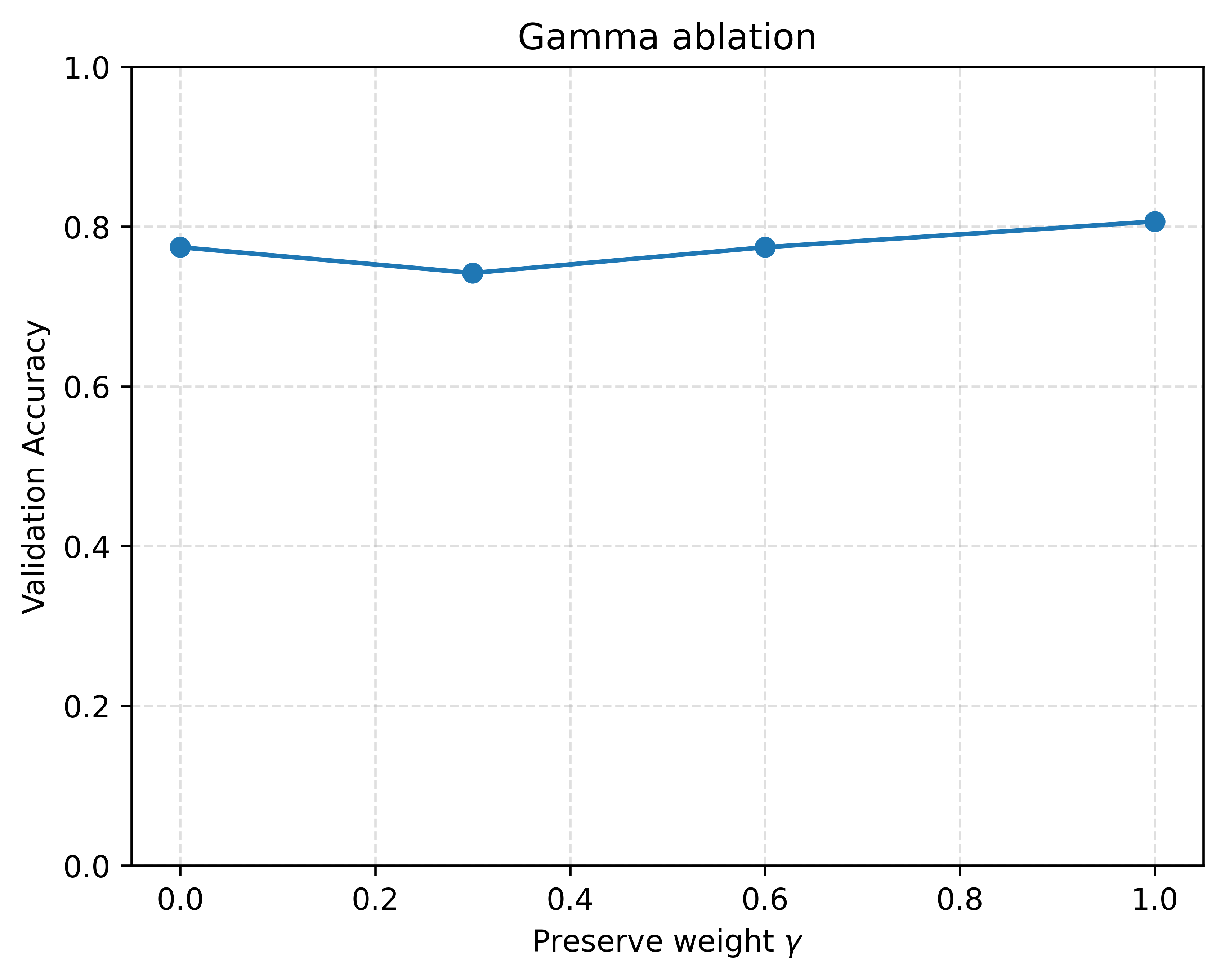}
    \caption{Weight $\gamma$.}
    \label{fig:abl_gamma}
  \end{subfigure}\hfill
  \begin{subfigure}[t]{0.235\textwidth}
    \centering
    \includegraphics[width=\linewidth,height=\ablimgheight,keepaspectratio]{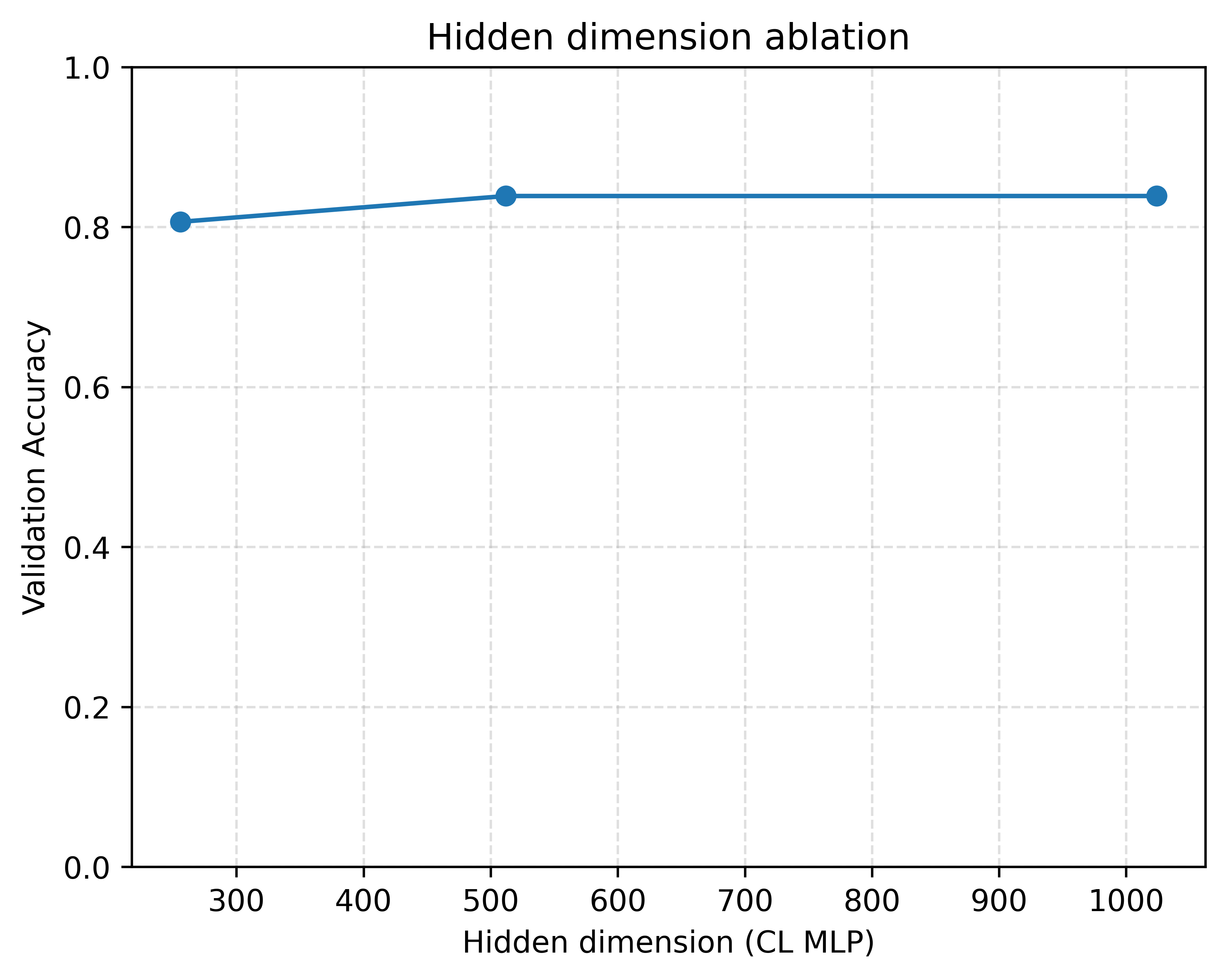}
    \caption{CL hidden dim.}
    \label{fig:abl_hidden}
  \end{subfigure}\hfill
  \begin{subfigure}[t]{0.235\textwidth}
    \centering
    \includegraphics[width=\linewidth,height=\ablimgheight,keepaspectratio]{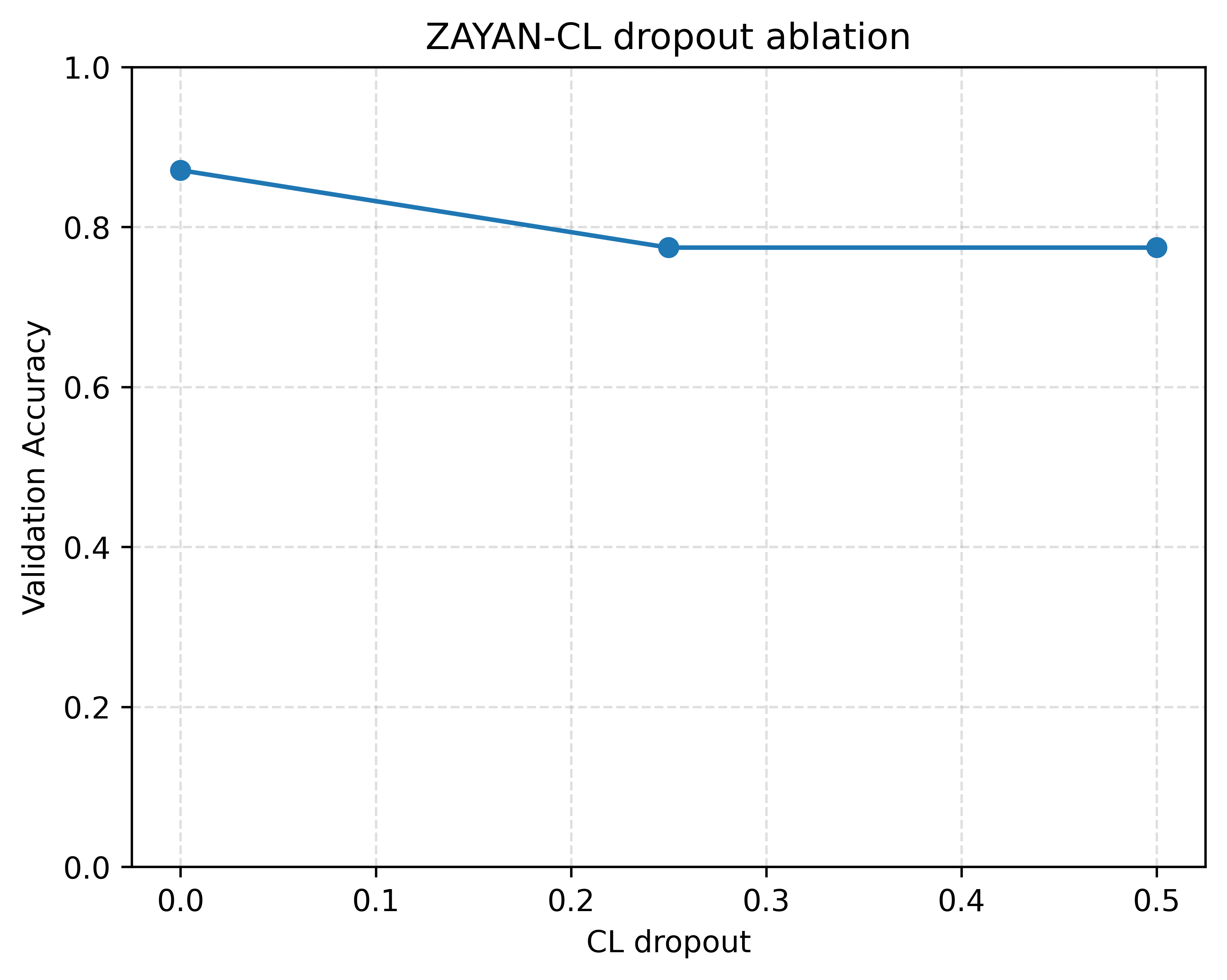}
    \caption{ZAYAN-CL dropout.}
    \label{fig:abl_cl_dropout}
  \end{subfigure}\hfill
  \begin{subfigure}[t]{0.235\textwidth}
    \centering
    \includegraphics[width=\linewidth,height=\ablimgheight,keepaspectratio]{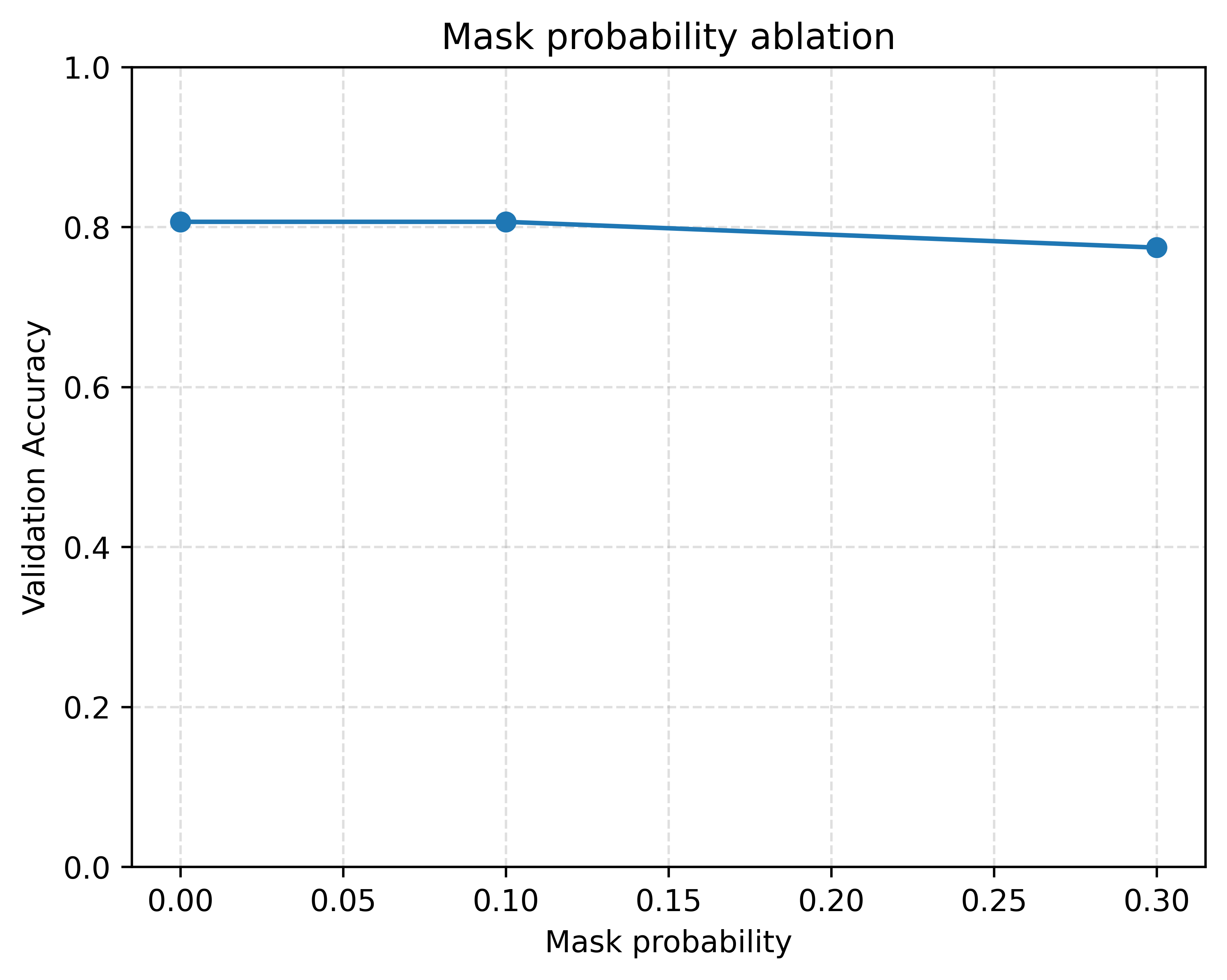}
    \caption{Mask prob.}
    \label{fig:abl_mask}
  \end{subfigure}

  \vspace{0.25em}

  \begin{subfigure}[t]{0.235\textwidth}
    \centering
    \includegraphics[width=\linewidth,height=\ablimgheight,keepaspectratio]{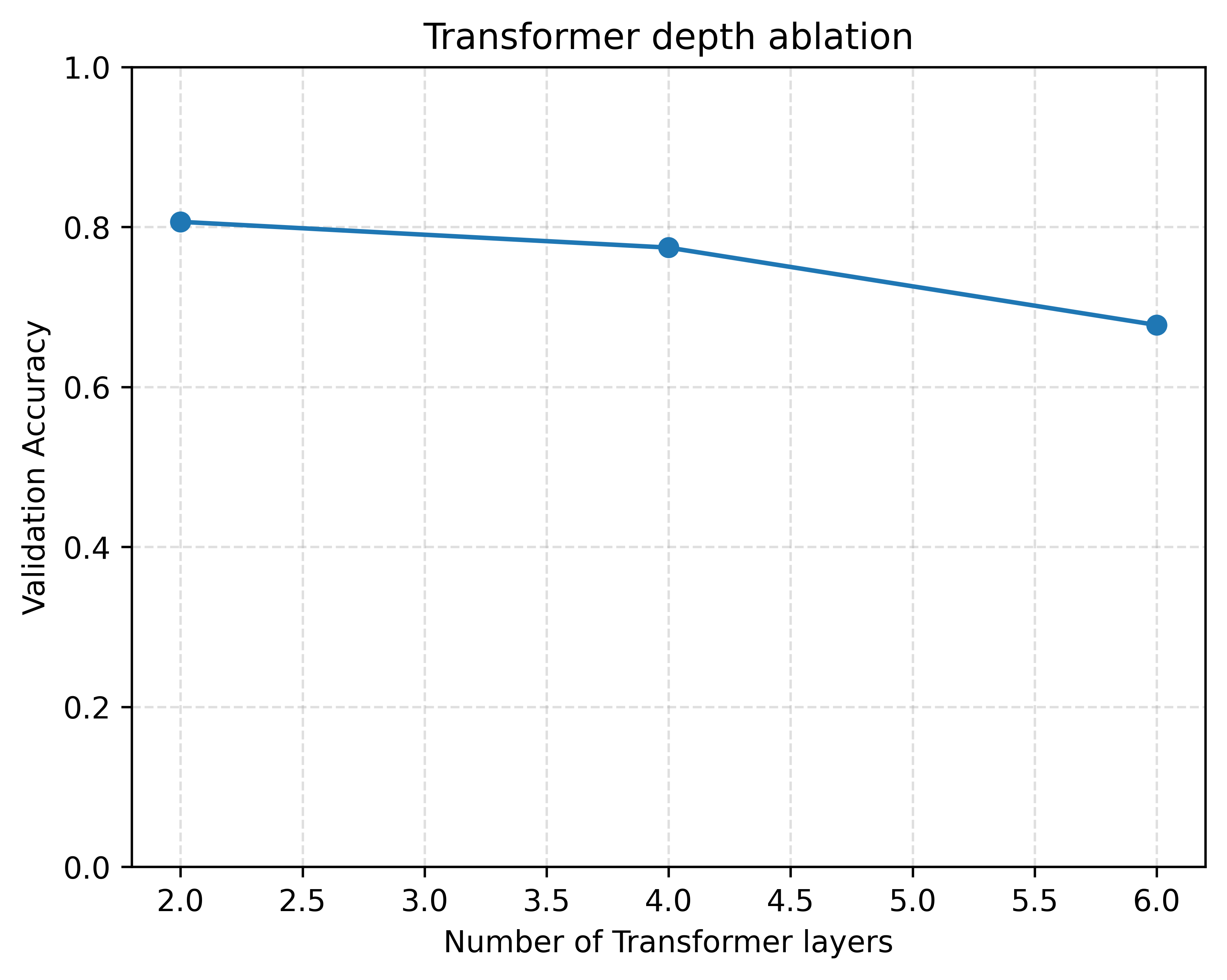}
    \caption{Transformer depth.}
    \label{fig:abl_layers}
  \end{subfigure}\hfill
  \begin{subfigure}[t]{0.235\textwidth}
    \centering
    \includegraphics[width=\linewidth,height=\ablimgheight,keepaspectratio]{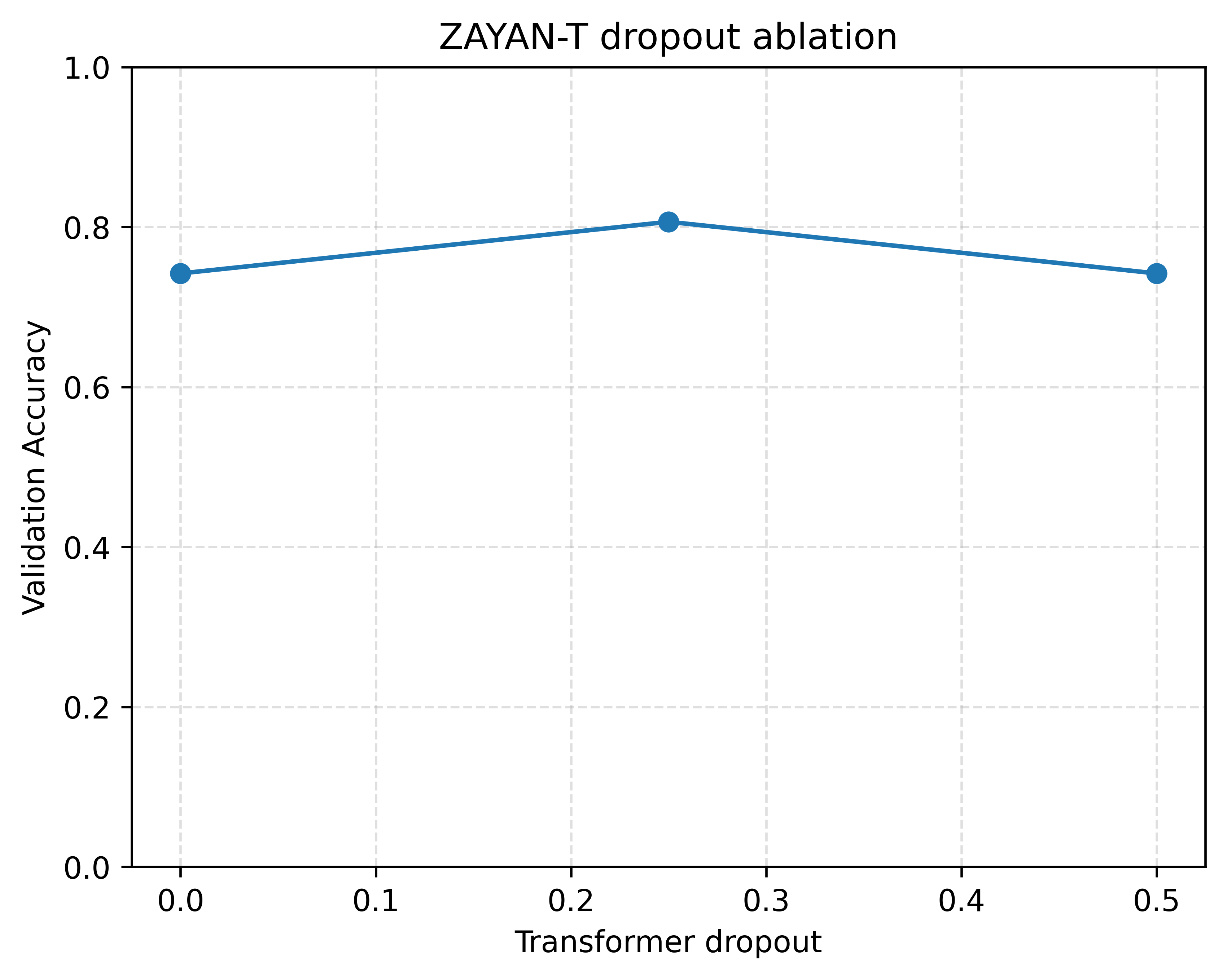}
    \caption{ZAYAN-T dropout.}
    \label{fig:abl_t_dropout}
  \end{subfigure}\hfill
  \begin{subfigure}[t]{0.235\textwidth}
    \centering
    \includegraphics[width=\linewidth,height=\ablimgheight,keepaspectratio]{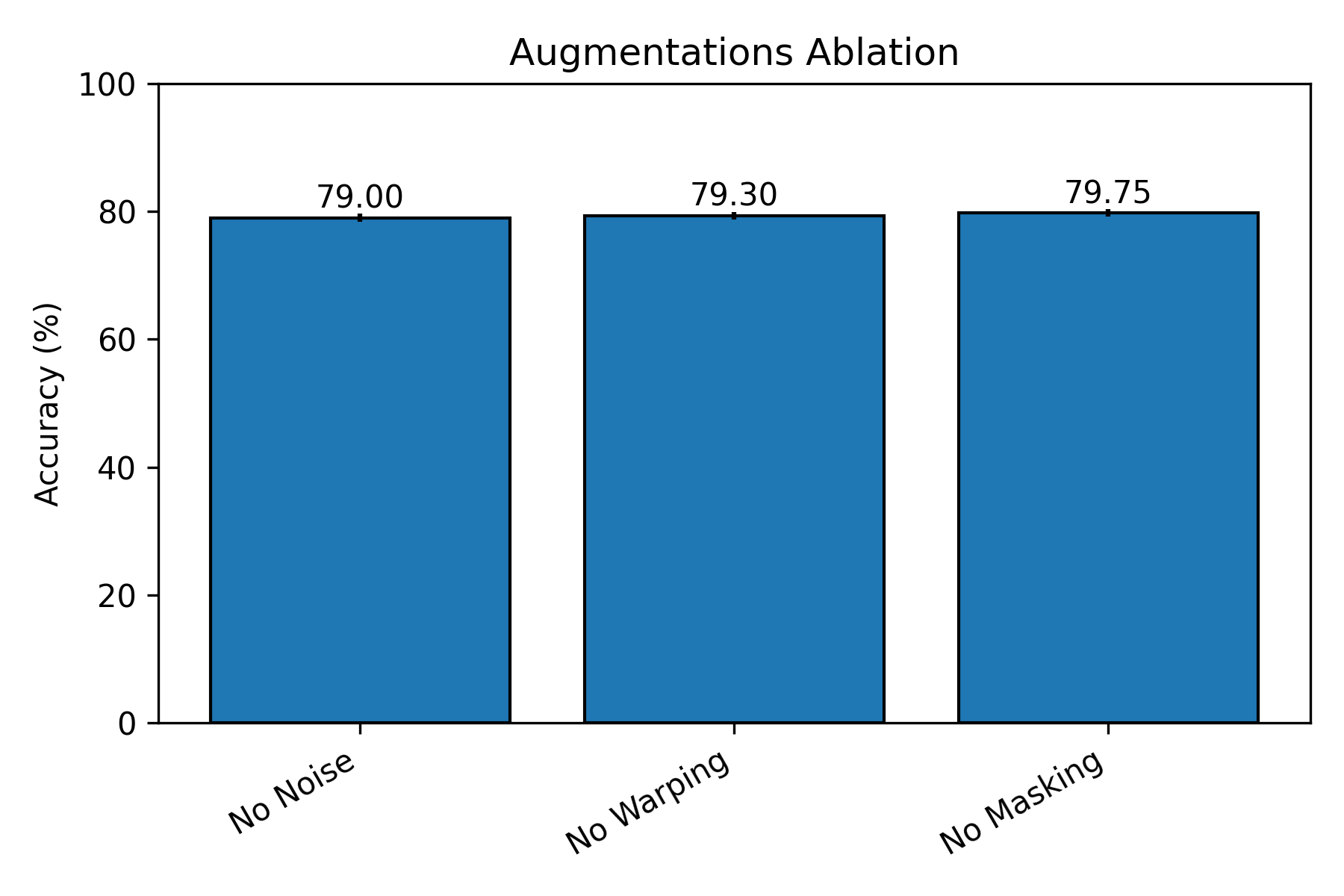}
    \caption{Augmentations.}
    \label{fig:abl_augs}
  \end{subfigure}\hfill
  \begin{subfigure}[t]{0.235\textwidth}
    \centering
    \includegraphics[width=\linewidth,height=\ablimgheight,keepaspectratio]{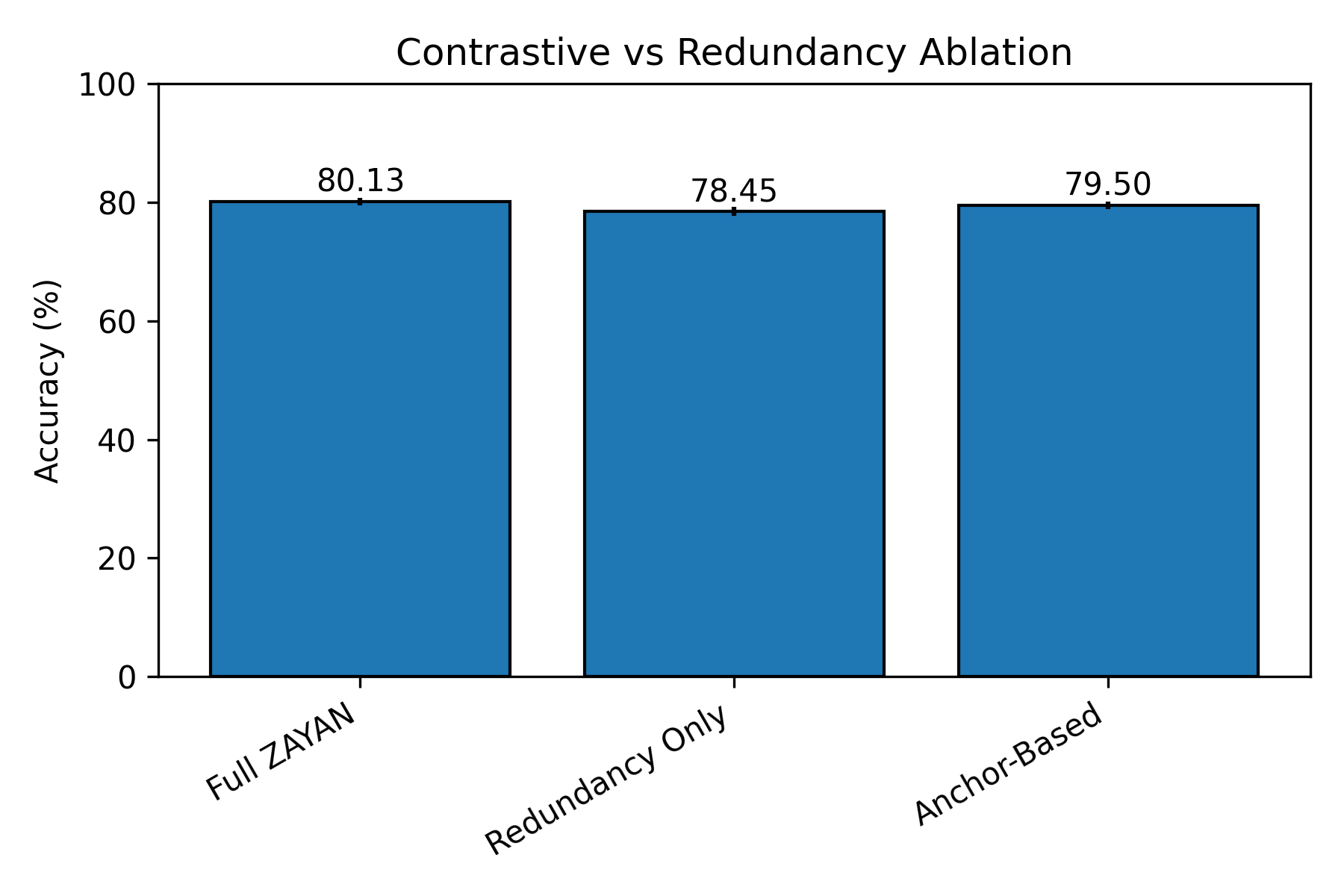}
    \caption{CL vs.\ redund.}
    \label{fig:abl_contrastive}
  \end{subfigure}

  \caption{Ablations for ZAYAN on Urban Land Cover: (a) batch size, (b) temperature $\tau$, (c) attention heads, (d) noise scale $\sigma$, (e) weight $\gamma$, (f) CL hidden dimension, (g) ZAYAN-CL dropout, (h) mask probability, (i) Transformer depth, (j) ZAYAN-T dropout, (k) augmentations, and (l) contrastive/redundancy variants.}
  \label{fig:zayan_ablation_grid}
\end{figure*}
\textbf{H. Ablation on Model Components:}
Fig.~\ref{fig:zayan_ablation_grid} and Table~\ref{tab:ablation_wilcoxon}(a) analyze how individual design choices of ZAYAN influence performance on Urban Land Cover. The batch-size and temperature sweeps in Figs.~\ref{fig:abl_batch}–\ref{fig:abl_tau} show a clear but shallow optimum around batch size $32$ and a moderate contrastive temperature (here $\tau\approx0.05$), with smaller or larger values yielding only modest drops, indicating that the contrastive branch is not overly sensitive to these settings. Architectural ablations in Figs.~\ref{fig:abl_heads}, \ref{fig:abl_hidden}, and \ref{fig:abl_layers} suggest that ZAYAN benefits from a relatively compact Transformer: increasing the number of attention heads or stacking deeper Transformers beyond the default quickly leads to diminished validation accuracy, whereas enlarging the CL-MLP hidden dimension beyond $512$ yields diminishing returns. Noise- and regularization-related ablations in Figs.~\ref{fig:abl_sigma}, \ref{fig:abl_gamma}, \ref{fig:abl_cl_dropout}, and \ref{fig:abl_t_dropout} reveal that moderate stochasticity is helpful: a non-zero input noise scale and transformer dropout ($\approx0.25$) improve generalization over the no-dropout setting, while excessively strong noise or dropout again hurts performance. The preservation weight $\gamma$ is particularly important: pushing $\gamma$ to $1.0$ yields the best validation accuracy, confirming the role of the preservation loss in keeping the learned “cheap” representation faithful to the original features. Masking ablations (Fig.~\ref{fig:abl_mask}) show that a small but non-zero mask probability is beneficial, whereas overly aggressive masking degrades accuracy. Fig.~\ref{fig:abl_augs} and the configuration-level ablation in Table~\ref{tab:ablation_wilcoxon}(a) jointly indicate that the three augmentations (noise injection, quantile warping, and feature masking) act synergistically: removing any one of them, or dropping the redundancy or preservation losses, consistently lowers mean accuracy by $4$–$9$ percentage points relative to the full model. Finally, Fig.~\ref{fig:abl_contrastive} highlights the importance of combining contrastive learning with redundancy regularization: a purely redundancy-based variant or a purely anchor-based contrastive variant both underperform the full ZAYAN configuration, demonstrating that the interplay between redundancy minimization, preservation, and contrastive objectives is critical for achieving the reported gains.\newline
\begin{table*}[htbp]
  \centering
  \caption{Inference-time robustness, OOD behaviour, and deployment-style
           triage diagnostics for ZAYAN on Urban Land Cover (all accuracies in \%).}
  \label{tab:robustness_ood_triage}
  \scriptsize
  \setlength{\tabcolsep}{2pt}

  \begin{minipage}[t]{0.24\textwidth}
    \centering
    \textbf{(a) Robustness to feature perturbations}\\[2pt]
    \begin{tabular}{@{}lccc@{}}
      \toprule
      Frac. & Shuffle & Drop & kNN@5 \\
      pert. & acc & acc & agree \\
      \midrule
      0.00 & 83.87 & 83.87 & 50.97 \\
      0.10 & 83.87 & 80.65 & 50.97 \\
      0.25 & 87.10 & 74.19 & 50.97 \\
      0.50 & 77.42 & 54.84 & 50.97 \\
      0.75 & 38.71 & 22.58 & 50.97 \\
      1.00 & 12.90 & 16.13 & 50.97 \\
      \bottomrule
    \end{tabular}
  \end{minipage}
  \hfill
  \begin{minipage}[t]{0.24\textwidth}
    \centering
    \textbf{(b) OOD confidence and local sensitivity}\\[2pt]
    \begin{tabular}{@{}lcc@{}}
      \toprule
      Regime & max-conf & entropy \\
      \midrule
      ID     & 0.915 & 0.256 \\
      Noise  & 0.850 & 0.491 \\
      Perm.  & 0.815 & 0.537 \\
      Const. & 0.955 & 0.218 \\
      \bottomrule
    \end{tabular}\\[4pt]
    \begin{tabular}{@{}lcc@{}}
      \toprule
      Local sens.\ ($\varepsilon=0.10$) & mean & median \\
      \midrule
      Tabular features & 0.002 & 0.001 \\
      \bottomrule
    \end{tabular}
  \end{minipage}
  \hfill
  \begin{minipage}[t]{0.36\textwidth}
    \centering
    \textbf{(c) Deployment-style triage diagnostics}\\[2pt]
    \begin{tabular}{@{}lc@{}}
      \toprule
      Quantity & Value \\
      \midrule
      Test size $N$              & 31 \\
      Base acc.\ (multi-cl.)     & 83.87 \\
      AUC (class 0 vs.\ rest)    & 1.000 \\
      Best threshold $t^*$       & 0.554 \\
      Sens./spec.\ at $t^*$      & 1.00 / 1.00 \\
      Acc./prec./rec.\ at $t^*$  & 1.00 / 1.00 / 1.00 \\
      Latency mean (ms/batch)    & 5.36 \\
      Latency p50 / p90 / p99    & 5.36 / 5.36 / 5.36 \\
      \bottomrule
    \end{tabular}
  \end{minipage}

\end{table*}
\begin{figure*}[t]
  \centering
  \setlength{\tabcolsep}{2pt}
  \newcommand{\infimgheight}{2.55cm}

  \begin{subfigure}[t]{0.24\textwidth}
    \centering
    \includegraphics[height=\infimgheight]{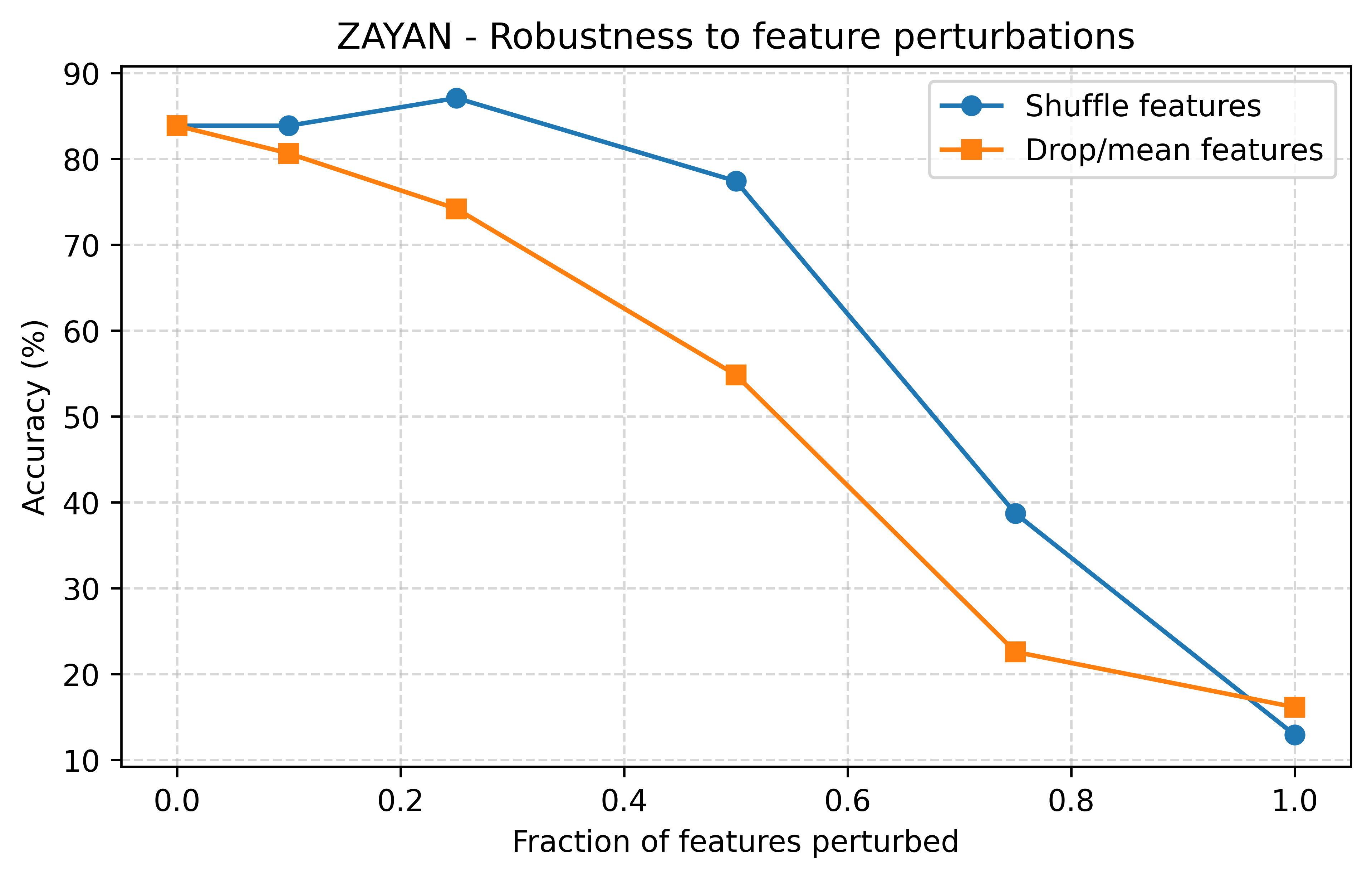}
    \caption{Robustness to perturbations.}
    \label{fig:robustness_perturb}
  \end{subfigure}\hfill
  \begin{subfigure}[t]{0.24\textwidth}
    \centering
    \includegraphics[height=\infimgheight]{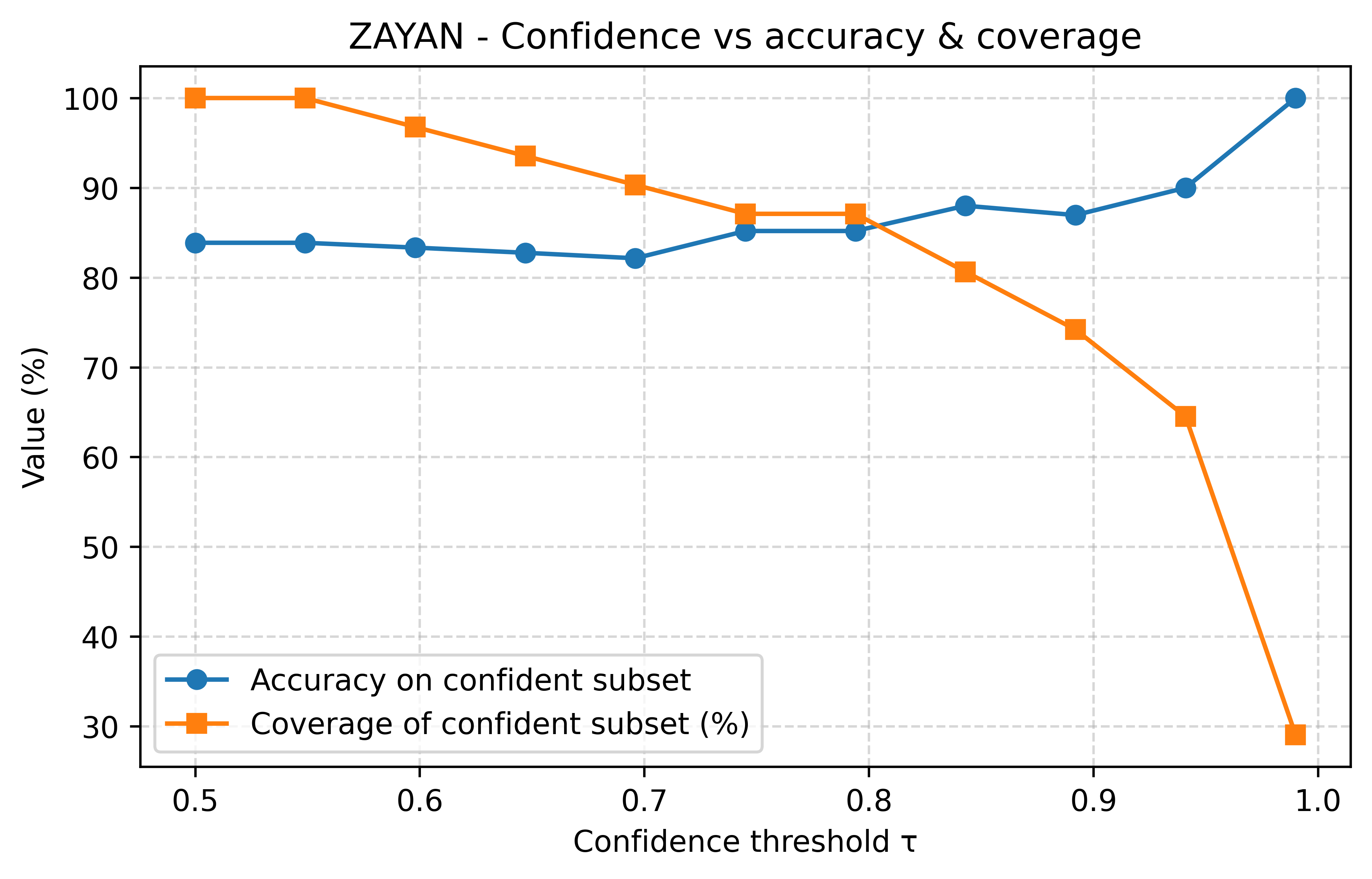}
    \caption{Confidence vs.\ accuracy/coverage.}
    \label{fig:conf_curve}
  \end{subfigure}\hfill
  \begin{subfigure}[t]{0.24\textwidth}
    \centering
    \includegraphics[height=\infimgheight]{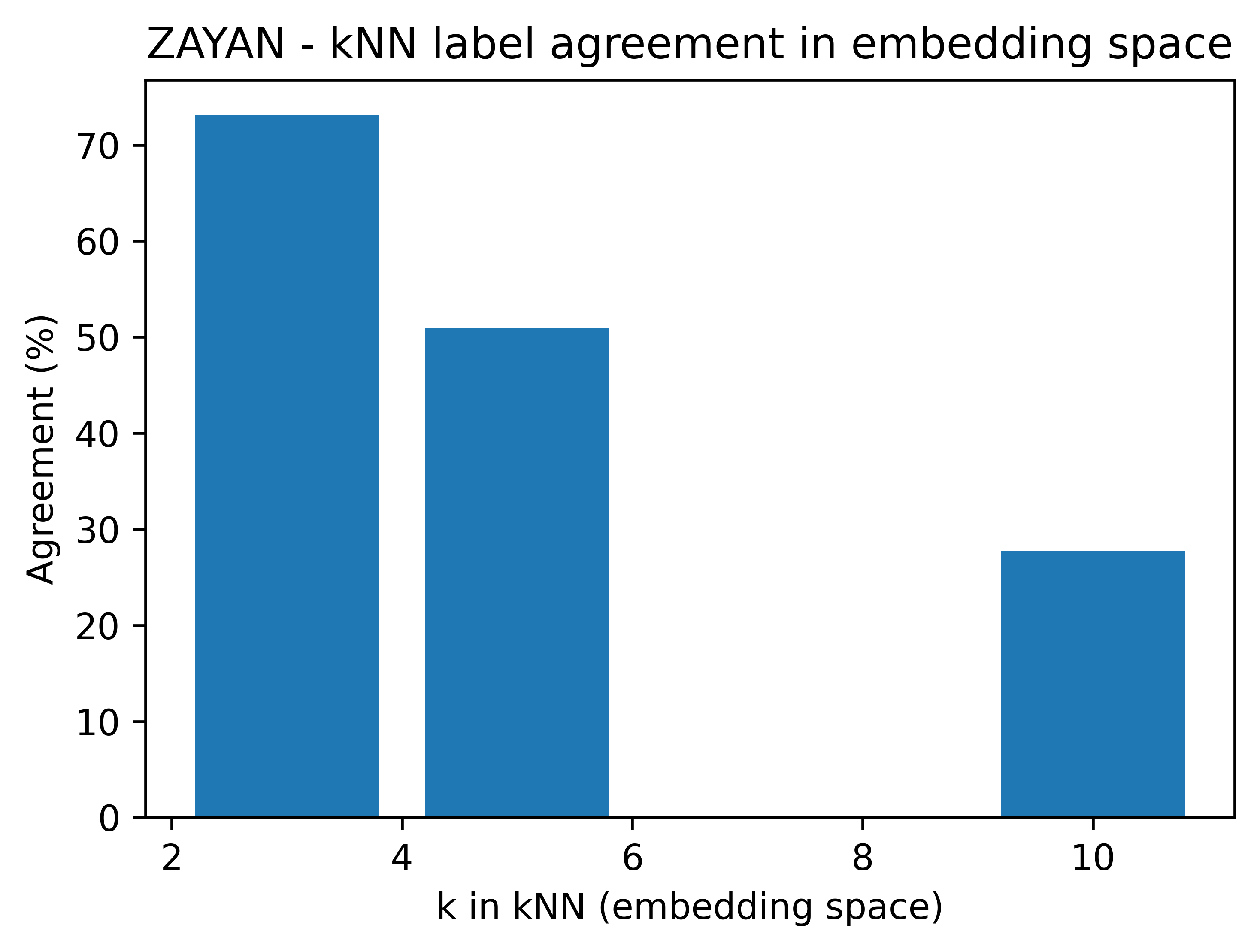}
    \caption{kNN agreement in embedding space.}
    \label{fig:knn_agreement}
  \end{subfigure}\hfill
  \begin{subfigure}[t]{0.24\textwidth}
    \centering
    \includegraphics[height=\infimgheight]{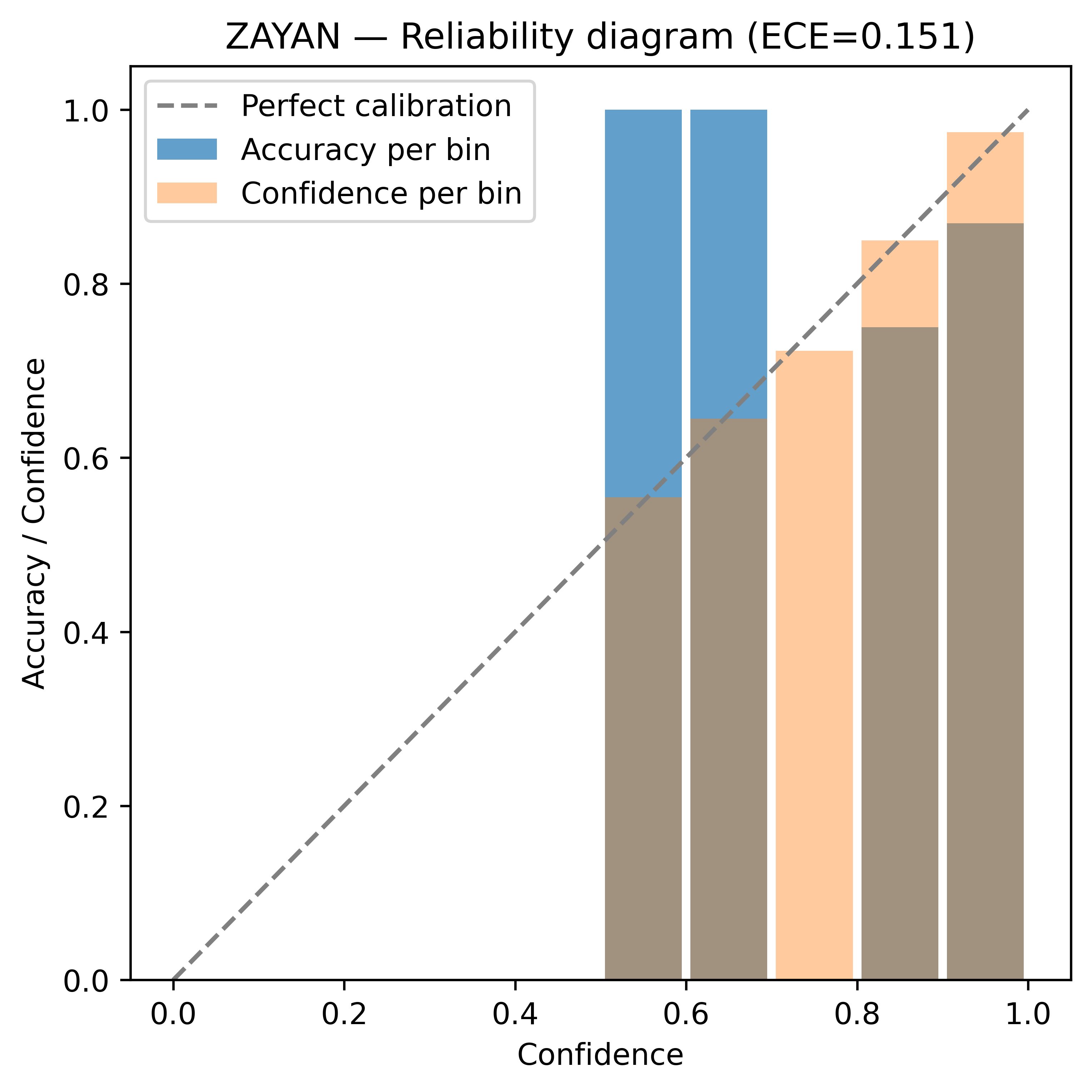}
    \caption{Reliability diagram (ECE $=0.151$).}
    \label{fig:reliability}
  \end{subfigure}

  \vspace{0.35em}

  \begin{subfigure}[t]{0.24\textwidth}
    \centering
    \includegraphics[height=\infimgheight]{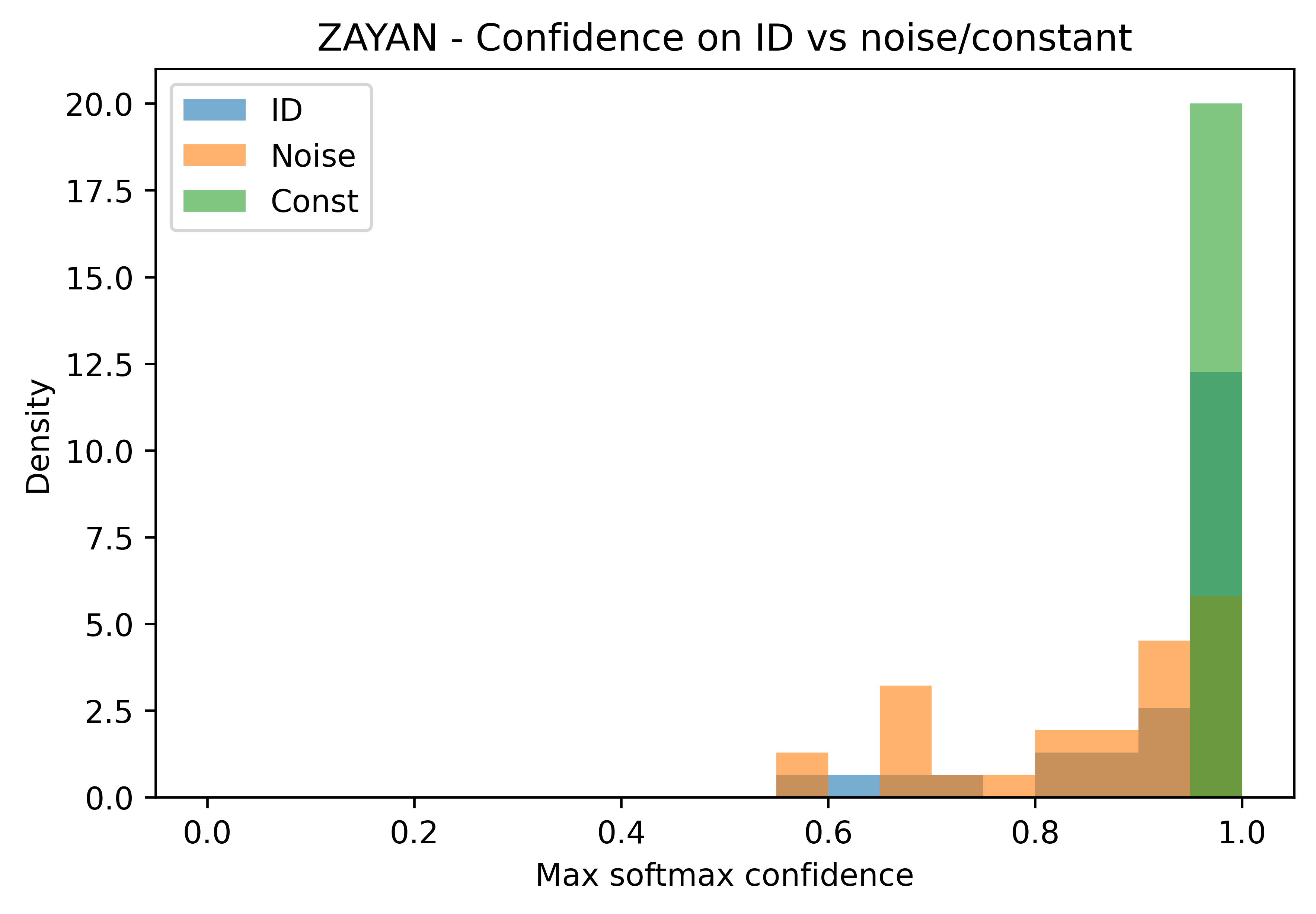}
    \caption{Confidence on ID vs.\ OOD.}
    \label{fig:ood_conf}
  \end{subfigure}\hfill
  \begin{subfigure}[t]{0.24\textwidth}
    \centering
    \includegraphics[height=\infimgheight]{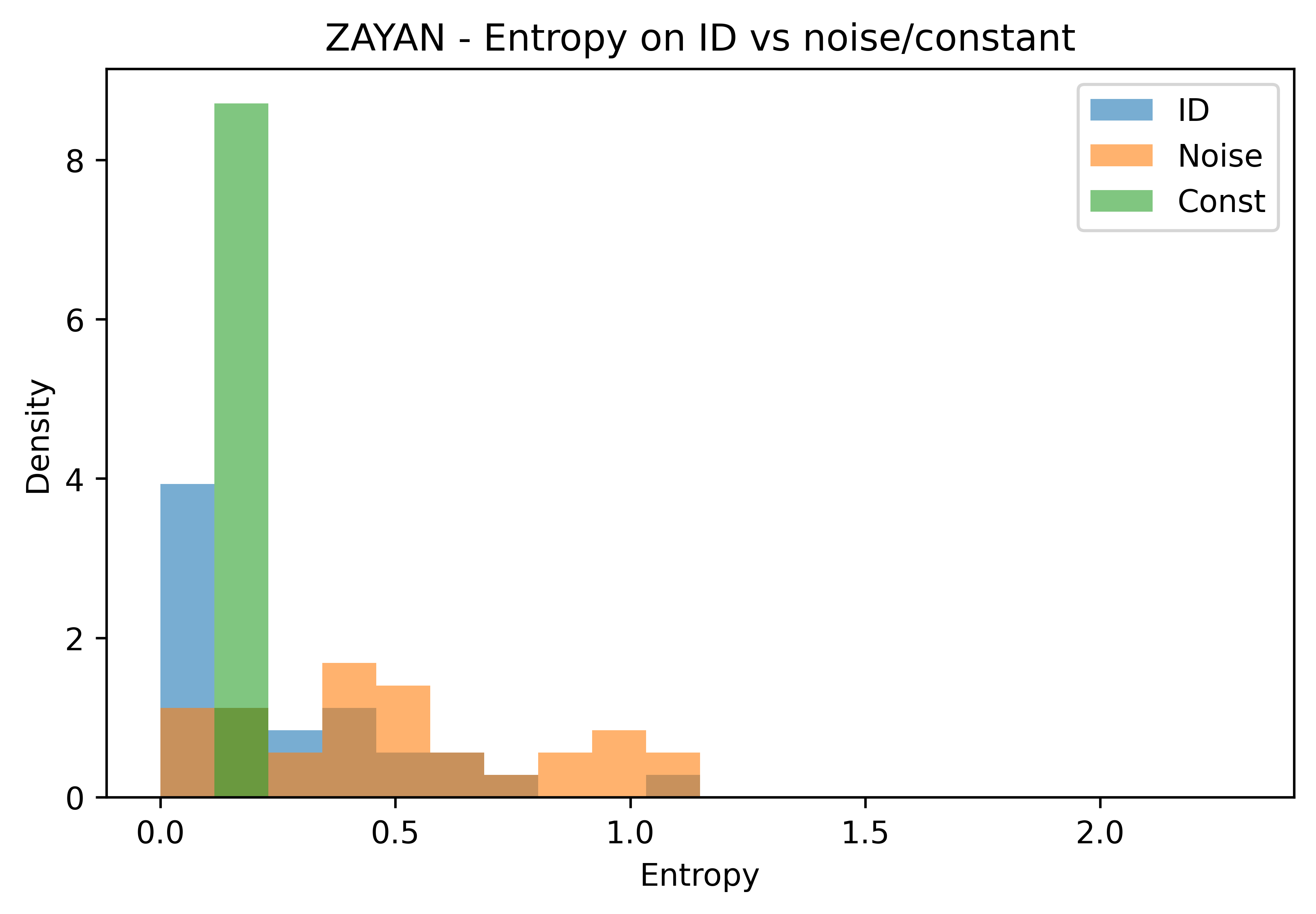}
    \caption{Entropy on ID vs.\ OOD.}
    \label{fig:ood_entropy}
  \end{subfigure}\hfill
  \begin{subfigure}[t]{0.24\textwidth}
    \centering
    \includegraphics[height=\infimgheight]{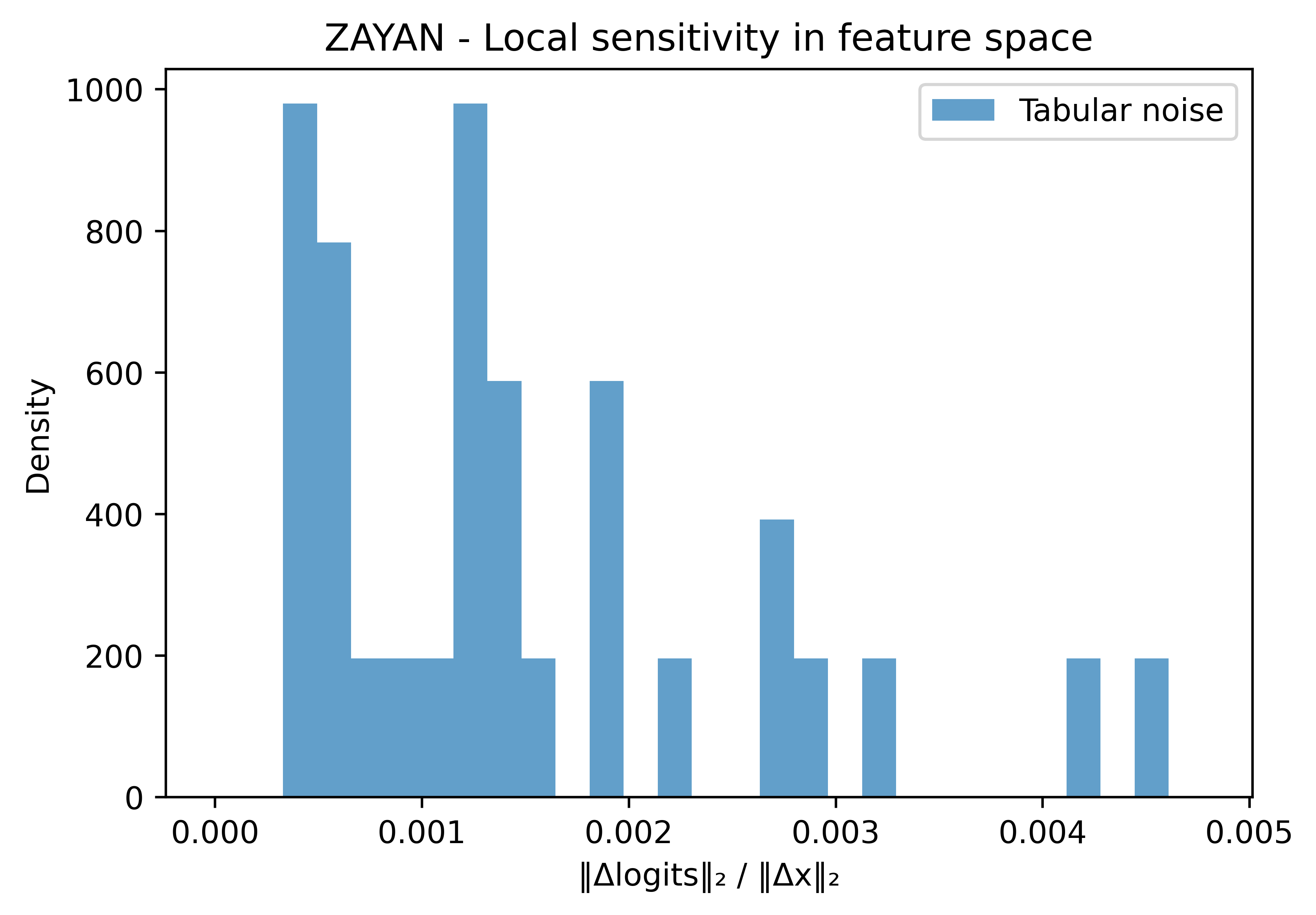}
    \caption{Local sensitivity.}
    \label{fig:local_sens}
  \end{subfigure}\hfill
  \begin{subfigure}[t]{0.24\textwidth}
    \centering
    \includegraphics[height=\infimgheight]{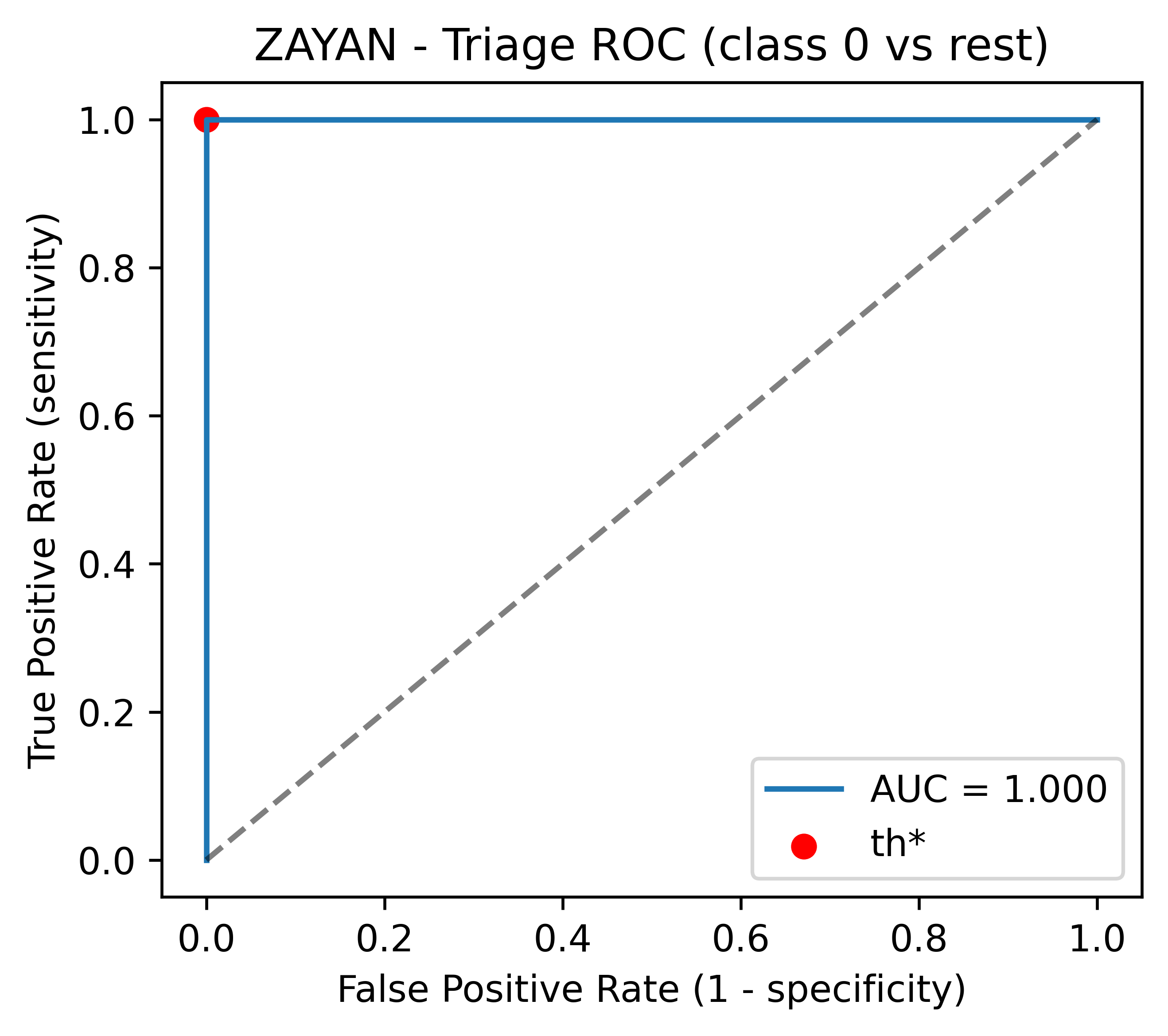}
    \caption{Triage ROC (AUC $=1.0$).}
    \label{fig:triage_roc}
  \end{subfigure}

  \caption{Inference-time diagnostics of ZAYAN on Urban Land Cover. 
  (a) Acc. under feature shuffling and dropping. 
  (b) Acc. and coverage vs.\ confidence threshold. 
  (c) kNN label agreement in the learned embedding space. 
  (d) Calibration reliability diagram. 
  (e)-(g) Confidence, entropy, and local sensitivity under ID and synthetic OOD inputs. 
  (h) ROC curve for the triage task (class~0 vs.\ rest).}
  \label{fig:zayan_inference_ablation}
\end{figure*}
\textbf{I. Inference-Level Ablation on Robustness and Calibration:}
We further probe ZAYAN at inference time to study robustness to feature corruption, confidence-accuracy trade-offs, and calibration (Fig.~\ref{fig:zayan_inference_ablation}, Table~\ref{tab:robustness_ood_triage}a). In Fig.~\ref{fig:zayan_inference_ablation}\subref{fig:robustness_perturb}, we progressively perturb the tabular features by either shuffling them across samples or dropping them and replacing with feature-wise means. Accuracy remains essentially stable up to $25\%$ shuffled features and only starts to drop beyond $50\%$, whereas feature dropping is more damaging, falling below $60\%$ once half of the features are removed. This suggests that ZAYAN can partially compensate for permutation-like noise in the learned representation, but is more sensitive to systematic information loss, consistent with its reliance on ordered meta-features. The robustness table in Tab.~\ref{tab:robustness_ood_triage}(a) also reports a constant kNN-Agree@5 of $\approx 51\%$ across perturbation levels, indicating that the embedding neighbourhood structure remains moderately label-consistent even when raw features are corrupted. Fig.~\ref{fig:zayan_inference_ablation}\subref{fig:conf_curve} studies selective prediction by varying a confidence threshold $\tau$ on the softmax scores. As $\tau$ increases, accuracy on the retained (``confident'') subset rises from $\approx 84\%$ to nearly $100\%$, while coverage monotonically decreases from all test samples to roughly one third of the data. This provides a simple knob to trade coverage for reliability in deployment. To better understand the embedding geometry, Fig.~\ref{fig:zayan_inference_ablation}\subref{fig:knn_agreement} reports kNN label agreement in the latent space; for $k=5$, over half of the neighbours share the same label, supporting the view that ZAYAN learns a class-aware manifold suitable for downstream retrieval or post-hoc kNN smoothing. Finally, the reliability diagram in Fig.~\ref{fig:zayan_inference_ablation}\subref{fig:reliability} yields an expected calibration error of $0.151$, showing that ZAYAN is moderately over-confident but still reasonably calibrated; in particular, the highest-confidence bins are closer to the ideal diagonal, which again aligns with the selective prediction behaviour.\newline
\textbf{J. OOD and Local Sensitivity Diagnostics:}
Table~\ref{tab:robustness_ood_triage}(b) and Fig.~\ref{fig:zayan_inference_ablation}\subref{fig:ood_conf}--\subref{fig:local_sens} analyse ZAYAN’s uncertainty behaviour under synthetic Out-Of-Distribution (OOD) regimes and its local smoothness. Noisy and permuted inputs exhibit the expected drop in mean max-softmax confidence and rise in predictive entropy relative to In-Distribution (ID) data (Table~\ref{tab:robustness_ood_triage}b), visible as a leftward shift in confidence and a rightward shift in entropy in Fig.~\ref{fig:zayan_inference_ablation}\subref{fig:ood_conf}--\subref{fig:ood_entropy}. By contrast, constant-valued inputs form a challenging OOD case, producing spuriously high confidence with low entropy, consistent with known failure modes of confidence-based OOD detection in tabular models. Finally, the local sensitivity distribution in Fig.~\ref{fig:zayan_inference_ablation}\subref{fig:local_sens} together with Table~\ref{tab:robustness_ood_triage}(b) shows that small $\ell_2$ perturbations in feature space ($\varepsilon=0.10$) induce only minor logit changes (mean $2\times10^{-3}$, median $10^{-3}$), suggesting a locally smooth decision surface around typical inputs.\newline
\textbf{K. Deployment-Style Triage Diagnostics:}
\label{subsec:triage_diagnostics}
Table~\ref{tab:robustness_ood_triage}c and Fig.~\ref{fig:zayan_inference_ablation}\subref{fig:triage_roc} report deployment-oriented diagnostics for a binary triage task on Urban Land Cover, where class~0 is treated as the “positive” class and all others are grouped as “negative”. While the base multiclass accuracy on the 31-sample test slice is $83.87\%$, the corresponding one-vs-rest triage problem is nearly separable, yielding an AUC of $1.000$ and an operating point $t^*=0.554$ with perfect sensitivity, specificity, accuracy, precision, and recall. The ROC curve in Fig.~\ref{fig:zayan_inference_ablation}\subref{fig:triage_roc} reflects this behaviour with a point lying on the top-left corner, indicating that ZAYAN can, in principle, support high-stakes triage decisions when restricted to such binary risk stratification. Latency measurements further show that these diagnostics can be computed efficiently: batches of 32 samples require around $5.36$\,ms on the tested GPU, with identical p50/p90/p99 statistics, suggesting stable inference-time behaviour suitable for real-time or near–real-time deployments on moderate-sized tabular workloads. See supplementary material for additional ablation studies and diagnostics.\newline
\textbf{L. Limitations:} The redundancy penalty in ZAYAN-CL currently incurs quadratic complexity ($\mathcal{O}(m^2 d)$ time, $\mathcal{O}(m^2)$ space), and ZAYAN-T may face GPU memory constraints (OOM) with very high-dimensional inputs (e.g., 2048-D ResNet embeddings or large datasets like Crop Mapping with 325,834 samples). Its purely self-supervised augmentations also omit potentially beneficial label information. Future work with sparse attention or supervised-contrastive hybrids could effectively address these scalability issues.
\section{Conclusion}
\label{con}
We have introduced ZAYAN, a self-supervised framework that decouples feature-level representation learning from downstream classification for image-derived tabular remote sensing and environmental data. ZAYAN-CL leverages zero-anchor contrastive learning and a redundancy penalty to produce disentangled, diversity-maximized feature embeddings without labels, while ZAYAN-T incorporates these embeddings into a Transformer with ZAYAN-aware positional encodings and a structure-preserving loss to improve task performance. Extensive experiments on eight benchmarks show that ZAYAN consistently outperforms both classical and modern tabular models in accuracy, robustness, and deployment-oriented diagnostics (calibration, OOD behaviour, and triage). At the same time, its quadratic feature complexity and Transformer resource requirements on very high-dimensional inputs highlight opportunities for future work on sparse or low-rank attention and hybrid supervised–contrastive objectives to scale ZAYAN to even larger tabular sensing datasets.
\section*{Acknowledgement}
This work was supported in part by the International Association for Pattern Recognition (IAPR) through ICPR 2026 registration support.
%
%
%
\bibliographystyle{splncs04}
\bibliography{ref}
\input{sup}
\end{document}

%% file: sup.tex
\clearpage
\maketitlesupplementary

\setcounter{section}{0}
\renewcommand{\thesection}{A1}
\renewcommand{\thesubsection}{A1.\arabic{subsection}}
\setcounter{figure}{0}\renewcommand{\thefigure}{A.\arabic{figure}}
\setcounter{table}{0}\renewcommand{\thetable}{A.\arabic{table}}


This supplementary document supports our main paper \textit{ZAYAN: Disentangled Contrastive Transformer for Tabular Remote Sensing Data} (Submitted to the 28\textsuperscript{th} International Conference on Pattern Recognition (ICPR) 2026). Specifically, it includes:
\begin{itemize}
    \item ZAYAN Hyperparameters in Sec.~\ref{ab2}
    \item Baseline Details and Tuned Hyperparameters for Selected Ones in Sec.~\ref{ab3}
    \item More on Model Components Ablation Studies in Sec.~\ref{ab4}
    \item Representation Quality via t-SNE in Sec.~\ref{ab7}
    \item Sanity and Stress Diagnostics in Sec.~\ref{ab9}
    \item Additional Reliability and Interpretability Diagnostics in Sec.~\ref{ab10}
    \item Theory-Inspired Representation Diagnostics in Sec.~\ref{ab11}
    \item Turing-Style Human-Model Evaluation in Sec.~\ref{ab12}
    \item Optuna-Level Diagnostics (Global Search Behavior) in Sec.~\ref{ab15}
\end{itemize}

\section{ZAYAN Hyperparameters}
\label{ab2}
ZAYAN has a compact but expressive hyperparameter set spanning the contrastive pretraining stage (ZAYAN‐CL) and the Transformer fine–tuning stage (ZAYAN‐T). For each dataset, we tune the contrastive learning rate (cl\_lr), weight decay (cl\_weight\_decay), temperature $\tau$, redundancy weight $\lambda$, perturbation scale $\sigma$, masking probability (mask\_prob), and encoder dropout (cl\_dropout), which jointly control the geometry and regularization of the feature–level embedding space. The downstream Transformer is tuned via its learning rate (t\_lr), weight decay (t\_weight\_decay), embedding and hidden dimensions (emb\_dim, hidden\_dim), attention configuration (nhead, num\_layers), and dropout (t\_dropout), as well as the preservation weight $\gamma$ that balances classification and structure–preserving losses. As shown in Table~\ref{tab:zayan_hparams}, ZAYAN typically favors moderate embedding sizes (64–256), shallow to medium‐depth Transformers (2–6 layers), and nontrivial redundancy penalties and preservation weights, while dataset–specific learning rates and masking/perturbation strengths adapt to differences in sample size, dimensionality, and label scarcity across the benchmarks.
\begin{table}[htbp]
  \centering
  \footnotesize
  \caption{Optuna-tuned ZAYAN hyperparameters for all datasets.}
  \label{tab:zayan_hparams}
  \begin{tabular}{@{}lccccccc@{}}
    \toprule
    \textbf{Param.}   & \textbf{Urban} & \textbf{Forest} & \textbf{Wilt} & \textbf{Indian} & \textbf{Pluvial} & \textbf{Crop} & \textbf{RSI‐CB256} \\
                      & \textbf{Land}  & \textbf{Type}   &               & \textbf{Flood}  & \textbf{Flood}   & \textbf{Mapping} & \\
    \midrule
    cl\_lr            & 2.06e-3 & 1.03e-4 & 5.78e-3 & 5.42e-3 & 3.85e-4 & 4.17e-4 & 6.05e-4 \\
    cl\_weight\_decay & 1.08e-4 & 1.58e-5 & 4.62e-4 & 9.19e-4 & 2.84e-5 & 1.50e-5 & 1.12e-4 \\
    t\_lr             & 4.23e-4 & 4.92e-4 & 1.62e-5 & 2.20e-5 & 4.18e-5 & 2.21e-4 & 2.66e-4 \\
    t\_weight\_decay  & 7.11e-4 & 1.94e-5 & 6.47e-4 & 7.97e-5 & 5.51e-4 & 3.10e-5 & 2.03e-5 \\
    emb\_dim          & 128     & 256     & 256     & 64      & 128     & 64      & 128     \\
    hidden\_dim       & 512     & 1024    & 512     & 256     & 256     & 256     & 256     \\
    $\tau$            & 0.160   & 0.0673  & 0.0806  & 0.0682  & 0.0874  & 0.0782  & 0.0752  \\
    $\lambda$         & 0.304   & 1.615   & 1.115   & 0.568   & 0.728   & 0.641   & 0.474   \\
    $\sigma$          & 0.0866  & 0.187   & 0.109   & 0.0643  & 0.0158  & 0.0838  & 0.0490  \\
    mask\_prob        & 0.215   & 0.232   & 0.263   & 0.154   & 0.1735  & 0.1346  & 0.0155  \\
    cl\_dropout       & 0.095   & 0.427   & 0.350   & 0.100   & 0.344   & 0.112   & 0.271   \\
    t\_dropout        & 0.397   & 0.260   & 0.105   & 0.239   & 0.248   & 0.119   & 0.198   \\
    $\gamma$          & 0.226   & 0.811   & 0.553   & 0.422   & 0.323   & 0.215   & 0.507   \\
    batch\_size       & 32      & 32      & 64      & 64      & 64      & 16      & 16      \\
    nhead             & 8       & 8       & 8       & 8       & 4       & 4       & 4       \\
    num\_layers       & 4       & 6       & 5       & 4       & 3       & 2       & 2       \\
    \bottomrule
  \end{tabular}
\end{table}
\renewcommand{\thesection}{A2}
\renewcommand{\thesubsection}{A2.\arabic{subsection}}
\section{Baseline Details and Tuned Hyperparameters for Selected Ones}
\renewcommand{\thesection}{\arabic{section}}
\label{ab3}
We summarize the baseline models used in this study, spanning classical linear methods, tree ensembles, deep neural networks, attention-based transformers, and recent tabular architectures (including foundation-style models). For each model, we briefly state its goal, key mechanism, and typical strengths/limitations. Implementation sources are listed in Table~\ref{tab:model_sources}.

\paragraph{Naive Bayes}
A generative classifier assuming conditional independence of features given the class. It is extremely fast and data-efficient, but its independence assumption can be badly violated on real tabular data.

\paragraph{$k$-NN}
A non-parametric method that predicts from the labels of the $k$ nearest training points in feature space. It can be intuitive and competitive on low-dimensional problems, but distance metrics become unreliable and inference costly as dimensionality grows.

\paragraph{Logistic Regression}
A linear discriminative model that maps features to class probabilities via a sigmoid/softmax link. It is simple, interpretable, and strong when the decision boundary is close to linear, but struggles with complex nonlinear interactions unless features are engineered.

\paragraph{SVM}
Maximum-margin classifiers (or regressors) with optional kernels to capture nonlinear structure. They often generalize well on moderate-sized datasets, but require careful tuning of kernel and regularization parameters and can scale poorly with many samples.

\paragraph{Decision Tree}
A single axis-aligned tree that recursively partitions the feature space. Trees are easy to interpret and capture feature interactions, yet they are high-variance models and tend to overfit without pruning or regularization.

\paragraph{MLP}
Fully connected feed-forward networks with nonlinear activations. They are flexible universal approximators, but in tabular domains they lack strong inductive bias and can be sensitive to feature scaling, regularization, and distribution shift.

\paragraph{1D-CNN}
Convolutional networks applied along the feature dimension to exploit local patterns in an assumed feature ordering. They can perform well when the feature order is meaningful (e.g., time-series), but performance is order-dependent and less robust when no natural ordering exists.

\paragraph{Random Forest}
An ensemble of decision trees trained on bootstrapped samples with random feature subsets. RFs reduce variance, are comparatively robust, and deliver strong performance with limited tuning on many tabular tasks.

\paragraph{AdaBoost}
A boosting method that sequentially reweights samples so later weak learners focus on previously misclassified points. It can be highly accurate on clean data, but is notably sensitive to label noise and outliers.

\paragraph{GBM}
Gradient boosting builds an additive ensemble where each new tree fits the residuals of the previous ensemble under a differentiable loss. It is very expressive, but requires careful control of tree depth, learning rate, and regularization to avoid overfitting.

\paragraph{LGBM}
A highly optimized gradient boosted tree implementation with histogram-based splits and leaf-wise growth. It scales well to large datasets, handles categorical features, and is a strong general-purpose baseline for tabular prediction.

\paragraph{XGBoost}
A widely used gradient boosting library with regularization, sparsity-aware split finding, and efficient quantile sketching. It remains a standard reference point for structured prediction benchmarks.

\paragraph{CatBoost}
A gradient boosting method tailored for categorical-rich datasets, using ordered boosting and target statistics that reduce target leakage. It is typically very strong when categorical features dominate.

\paragraph{TabNet}
A deep architecture that uses sparse attentive masks to select features at each decision step. It jointly learns which features to attend to and when, providing some interpretability through the learned masks.

\paragraph{TabTransformer}
A transformer-based model that embeds categorical features as tokens and processes them (with continuous features) via multi-head self-attention. Contextualized feature embeddings help capture high-order dependencies across heterogeneous attributes.

\paragraph{FT-Transformer}
A streamlined transformer variant for tabular data using learned embeddings for numerical features and simple feature encodings. It retains the expressiveness of attention while reducing architectural complexity relative to earlier tabular transformers.

\paragraph{TabSeq}
A sequence-style model that learns feature permutations and then applies sequential backbones (transformers or RNNs) over the ordered features. Treating features as tokens allows position-aware modeling of interactions along the learned sequence.

\paragraph{TANGOS}
A regularization scheme that encourages gradients for different units to be approximately orthogonal and specialized. This promotes disentangled representations and can improve robustness to noisy or redundant features in tabular networks.

\paragraph{TabPFN}
A meta-learned transformer pretrained on millions of synthetic tabular tasks to approximate Bayesian inference. At test time it performs near one-shot classification via a single forward pass, eliminating per-task training but being limited to small-scale problems.

\paragraph{NODE}
Neural Oblivious Decision Ensembles replace hard decision trees by differentiable “oblivious” trees whose splits are shared at each depth. The resulting model is end-to-end trainable while retaining tree-like inductive bias for tabular data.

\paragraph{SAINT}
A transformer-style architecture that combines row-wise and column-wise attention with contrastive pretraining and strong data augmentations. It is particularly effective in low-label or class-imbalanced regimes.

\paragraph{DeepFM}
An architecture that combines a factorization machine component for low-order feature interactions with an MLP for higher-order patterns. It is widely used for CTR and recommendation problems where cross-feature effects are crucial.

\paragraph{DCN}
The Deep \& Cross Network stacks explicit “cross” layers that construct feature interactions together with a deep tower network. This yields a hybrid of manually constructed and learned high-order crosses for tabular prediction.

\paragraph{AutoInt}
A self-attention–based model that automatically learns high-order feature interactions via multi-head attention over feature embeddings, reducing the need for manual feature engineering in recommender-style tasks.

\paragraph{TabPFN v2}
A refined TabPFN variant with improved calibration and uncertainty estimation. It aims to provide more reliable probabilities and better behavior under distribution shift while preserving near–zero-shot inference.

\paragraph{TabR}
A semi-parametric method that augments tabular networks with nearest-neighbor retrieval. By mixing parametric predictions with local neighbor information, it can better handle rare patterns and tail examples.

\paragraph{ProtoGate}
A prototype-based tabular network that learns global prototypes and uses sparse gating to select relevant features for each prototype. This offers a trade-off between accuracy and interpretability via prototype explanations and feature gates.

\paragraph{TabM}
An architecture that shares a backbone across multiple lightweight heads, effectively implementing a parameter-efficient ensemble. This reduces variance relative to a single network while keeping compute overhead modest.

\paragraph{TabulaRNN}
A recurrent model that scans along the feature dimension with dynamic positional encodings. It is designed to exploit sequential dependencies among features when an ordering (learned or given) is available.

\paragraph{TabICL}
A tabular foundation model trained across many heterogeneous datasets to support in-context learning. At test time, a small set of labeled rows is concatenated with a query row and processed by a transformer, enabling adaptation to new tasks without gradient updates. It is powerful in few-shot regimes but demands substantial pretraining resources and careful prompt (context) construction.

\paragraph{TANDEM}
A hybrid self-supervised autoencoder for tabular data that couples a neural encoder with an oblivious soft decision tree (OSDT) encoder, both feeding a shared decoder. Sample-specific stochastic gating networks create encoder-specific masked views, and reconstruction plus alignment losses encourage complementary yet consistent representations. At inference, only the neural encoder is used, yielding a lean prediction path while benefiting from tree-like inductive biases learned during pretraining.

\paragraph{Selected Baseline Hyperparameters.}
Tree and $k$NN baselines are tuned \emph{per dataset} with Optuna over standard ranges, while deep baselines share a fixed configuration across all datasets (Table~\ref{tab:baseline_hparams}). For example, on the Urban land cover dataset, XGBoost selects \texttt{n\_estimators} $=69$ and \texttt{learning\_rate} ${\approx}0.0627$, CatBoost uses \texttt{iterations} $=176$ with \texttt{learning\_rate} ${\approx}0.0666$, and the scikit-learn GBM favors \texttt{n\_estimators} $=92$ with \texttt{max\_depth} $=17$. Random Forest and $k$NN retain near-default settings, with $k$ chosen from a small grid on standardized features. For deep tabular baselines, we adhere to the publicly released TabICL configuration, and use a unified, code-aligned setup for TANDEM, TabM, and TabR (hidden dimensions, depths, optimization schedule, and retrieval hyperparameters), ensuring that ZAYAN is compared against strong and consistently tuned models.

\begin{table*}[htbp]
\centering
\small
\setlength{\tabcolsep}{3pt}
\renewcommand{\arraystretch}{1.1}
\caption{Selected hyperparameters for the baselines used in our experiments. Tree / $k$NN models are tuned per dataset with Optuna; deep baselines share the listed configurations across datasets.}
\label{tab:baseline_hparams}
\begin{tabular}{|p{0.16\textwidth}|p{0.16\textwidth}|p{0.63\textwidth}|}
\hline
\textbf{Model} & \textbf{Family} & \textbf{Key hyperparameters / tuning} \\
\hline
Random Forest
& Tree ensemble
& For each dataset, Optuna tunes \texttt{n\_estimators}, \texttt{max\_depth}, minimum samples per leaf, and maximum features per split; we use the best configuration per dataset. \\
\hline
GBM
& Tree ensemble
& scikit-learn gradient boosting classifier with tuned \texttt{n\_estimators}, \texttt{learning\_rate}, base tree depth (\texttt{max\_depth}), and (where applicable) subsampling / column-sampling ratios. Hyperparameters are selected via Optuna on validation folds and applied per dataset (e.g., Urban land cover: \texttt{n\_estimators} $=92$, \texttt{max\_depth} $=17$). \\
\hline
XGBoost
& Tree ensemble
& Tree booster with tuned \texttt{n\_estimators}, \texttt{max\_depth}, \texttt{learning\_rate} (\(\eta\)), \texttt{subsample}, \texttt{colsample\_bytree}, and \texttt{reg\_lambda}/\texttt{reg\_alpha}. We follow standard tabular search ranges and keep the Optuna-best setting per dataset (e.g., Urban land cover: \texttt{n\_estimators} $=69$, \texttt{learning\_rate} ${\approx}0.0627$). \\
\hline
CatBoost
& Tree ensemble
& Gradient boosting with ordered boosting for categorical features. For each dataset, Optuna tunes \texttt{iterations}, \texttt{learning\_rate}, tree \texttt{depth}, and \texttt{l2\_leaf\_reg}; we fix the best configuration (e.g., Urban land cover: \texttt{iterations} $=176$, \texttt{learning\_rate} ${\approx}0.0666$). \\
\hline
$k$NN
& Distance-based
& Standard $k$-nearest neighbors classifier with Euclidean distance. Optuna selects \texttt{n\_neighbors} and \texttt{weights} $\in\{\texttt{uniform},\texttt{distance}\}$ per dataset on z-score–standardized features; the best pair $(k,\texttt{weights})$ is used for each dataset. \\
\hline
TabICL
& Foundation / transformer
& Tabular foundation model evaluated in in-context mode. We use the public TabICL implementation without changing architecture-level hyperparameters (layers, heads, embedding size). Dataset-specific inference settings (number of in-context examples, maximum sequence length) follow the official defaults; our code release logs the exact configurations. \\
\hline
TANDEM
& Hybrid autoencoder
& Implementation following the public TANDEM design. Shared configuration:
\texttt{hidden\_layers} $=[256,128]$, \texttt{num\_trees} $=1$, \texttt{osdt\_depth} $=7$ (latent dim $2^{7}{=}128$),
\texttt{use\_gating} $=\text{True}$, \texttt{gating\_hidden\_dim} $=128$, \texttt{gating\_sigma} $=0.5$;
optimizer: Adam (\texttt{lr} $=10^{-3}$, \texttt{weight\_decay} $=10^{-5}$);
training: \texttt{batch\_size} $=64$, \texttt{max\_epochs} $=100$, \texttt{patience} $=10$,
reconstruction weight \texttt{rec\_weight} $=1.0$, alignment weight \texttt{align\_weight} $=0.1$. \\
\hline
TabM
& Mixture-of-experts DL
& Mixture-of-experts MLP shared across datasets:
\texttt{d\_hidden} $=128$, \texttt{n\_layers} $=2$, \texttt{n\_experts} $=8$, \texttt{dropout} $=0.2$;
optimizer: Adam (\texttt{lr} $=10^{-3}$, \texttt{weight\_decay} $=10^{-5}$);
training: \texttt{batch\_size} $=64$, \texttt{max\_epochs} $=50$, \texttt{patience} $=10$.
The gating network and experts are trained jointly under cross-entropy with gradient clipping (\texttt{max\_norm} $=1.0$). \\
\hline
TabR
& Retrieval-augmented DL
& Encoder MLP plus retrieval head. Shared configuration:
encoder \texttt{emb\_dim} $=64$, \texttt{d\_hidden} $=128$, \texttt{n\_layers} $=2$, \texttt{dropout} $=0.2$;
optimizer: Adam (\texttt{lr} $=10^{-3}$, \texttt{weight\_decay} $=10^{-5}$);
training: \texttt{batch\_size} $=64$, \texttt{max\_epochs} $=50$, \texttt{patience} $=10$.
At inference, classifier probabilities are combined with $k$NN over train embeddings using
\texttt{k\_neighbors} $=5$ and mixing coefficient \texttt{retrieval\_alpha} $=0.5$. \\
\hline
\end{tabular}
\end{table*}
\renewcommand{\thesection}{A3}
\renewcommand{\thesubsection}{A3.\arabic{subsection}}
\section{More on Model Components Ablation Studies}
\begin{figure*}[htbp]
    \centering
    \begin{subfigure}[t]{0.32\textwidth}
        \centering
        \includegraphics[width=\linewidth]{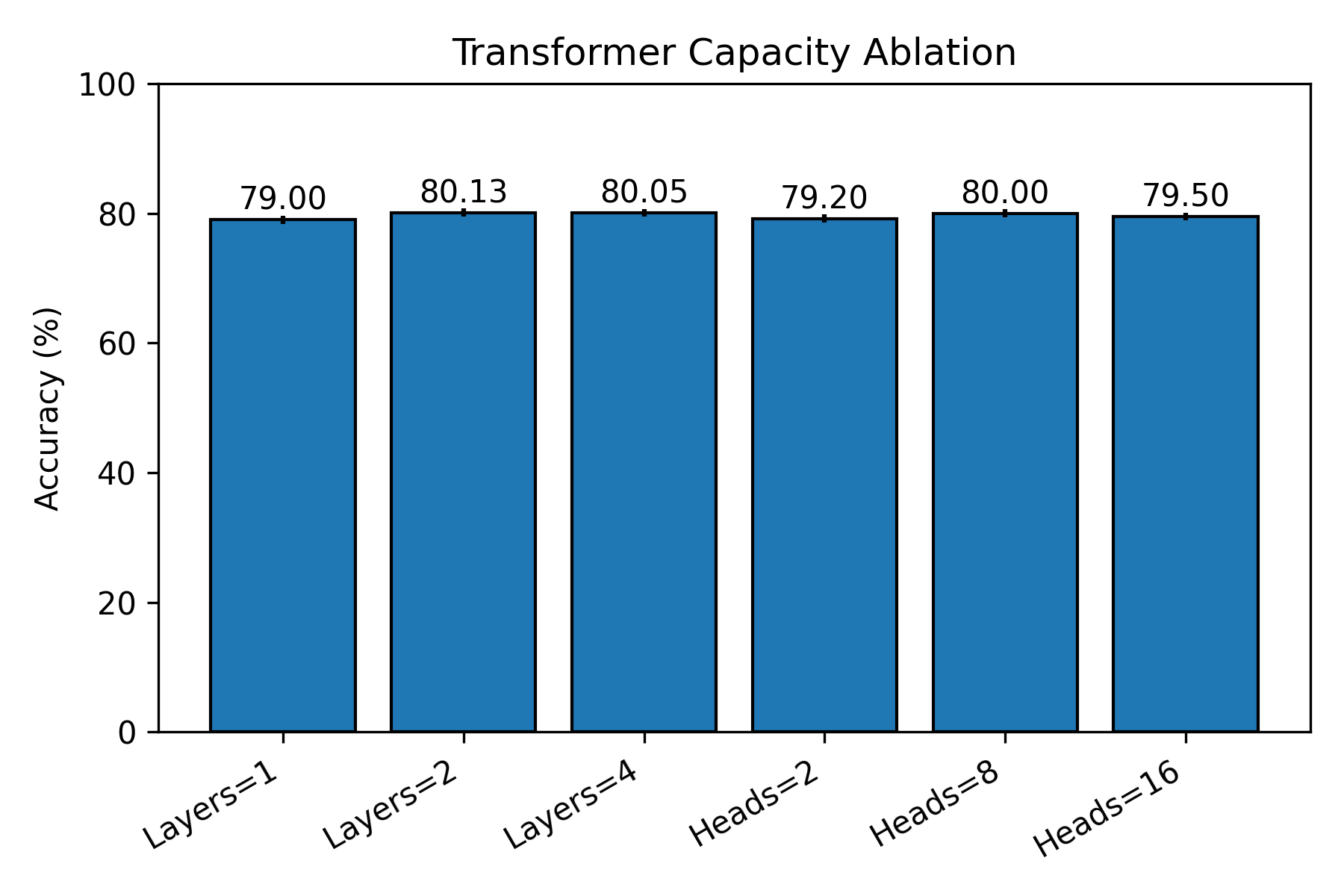}%
        \caption{Transformer depth and head count.}
        \label{fig:abl_transformer_capacity}
    \end{subfigure}
    \hfill
    \begin{subfigure}[t]{0.32\textwidth}
        \centering
        \includegraphics[width=\linewidth]{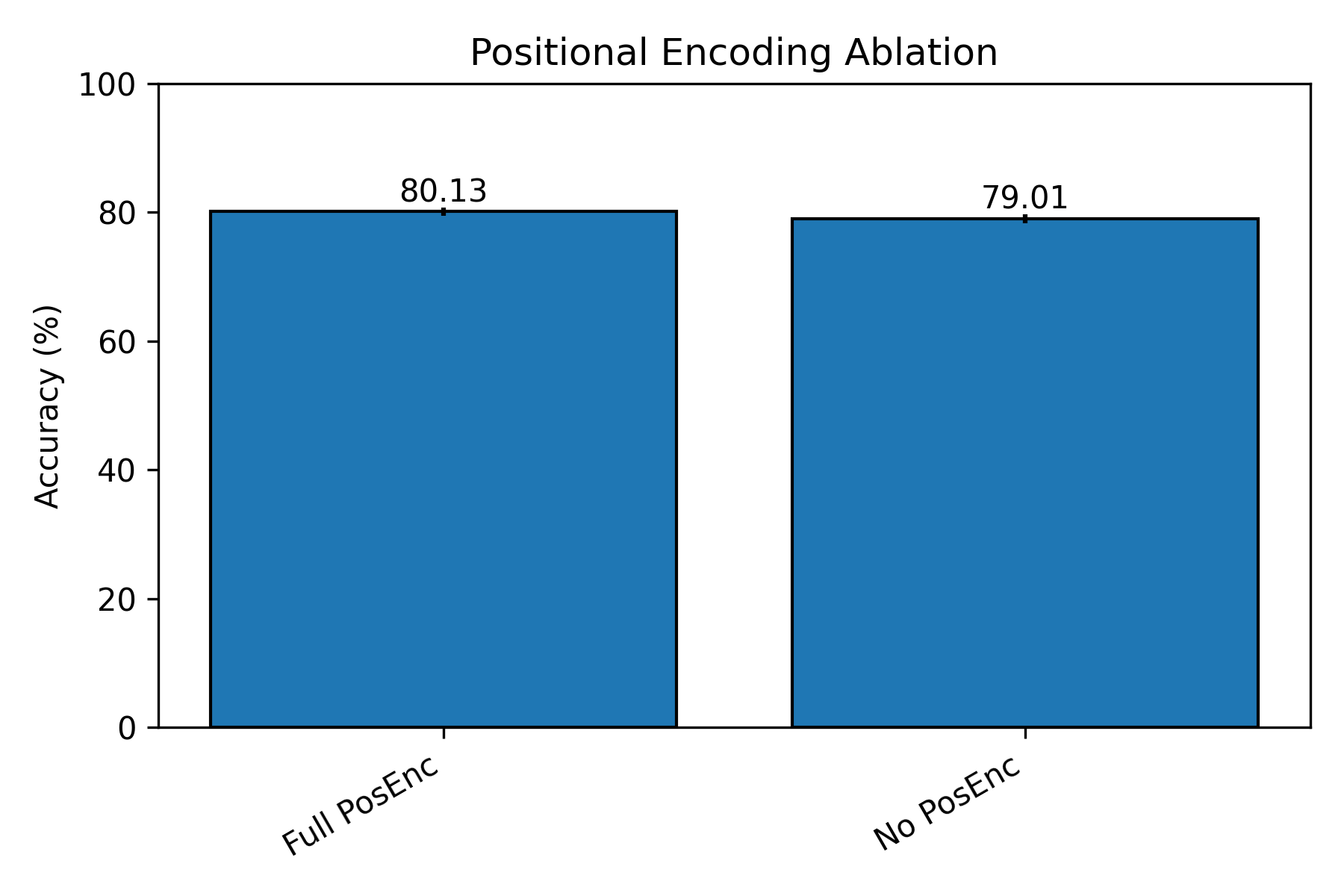}%
        \caption{Effect of positional encodings.}
        \label{fig:abl_posenc}
    \end{subfigure}
    \hfill
    \begin{subfigure}[t]{0.32\textwidth}
        \centering
        \includegraphics[width=\linewidth]{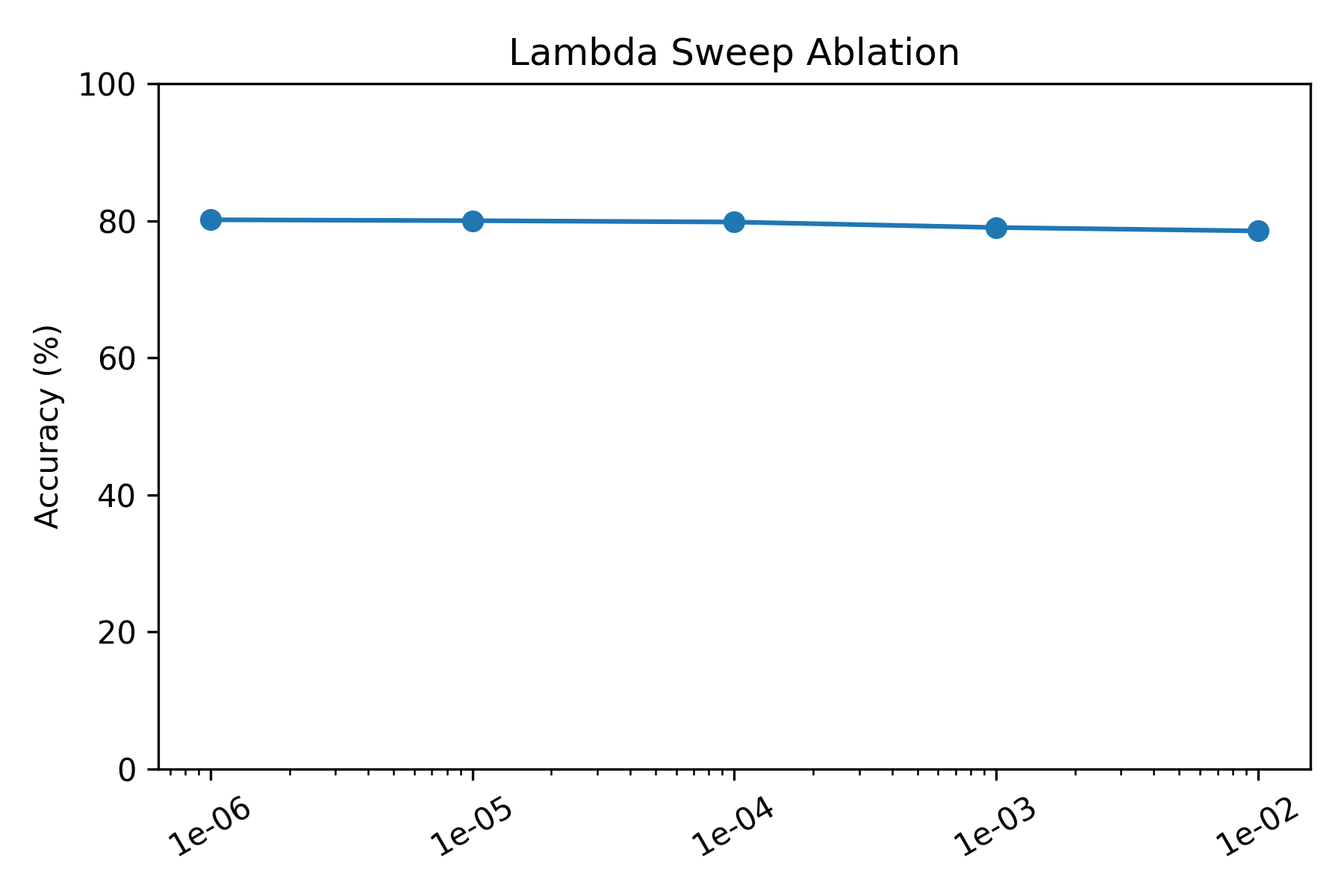}%
        \caption{Redundancy penalty weight $\lambda$.}
        \label{fig:abl_lambda}
    \end{subfigure}
    \caption{Additional ablations of ZAYAN on the Urban land cover dataset.
(a) Accuracy changes only slightly when varying the number of Transformer layers and heads, indicating limited sensitivity to encoder capacity.
(b) Removing positional encodings causes a small but consistent drop in accuracy, showing the benefit of feature-order information.
(c) Sweeping the redundancy penalty $\lambda$ over several orders of magnitude keeps accuracy near $79$--$80\%$ except for the largest values, suggesting a robust regularizer and a reasonable default choice of $\lambda$.}
    \label{fig:more_ablation}
\end{figure*}

\renewcommand{\thesection}{\arabic{section}}
\label{ab4}
Figure~\ref{fig:more_ablation} further probes the robustness of ZAYAN’s components on the Urban land cover dataset. The transformer capacity study shows that varying the encoder depth (1–4 layers) and the number of heads (2–16) keeps accuracy in a narrow band around 79–80\%, suggesting that ZAYAN does not rely on a carefully tuned, high-capacity transformer and is relatively insensitive to moderate architectural changes. The positional encoding ablation indicates that removing feature-wise positional embeddings yields a small but consistent drop in performance (from about 80.1\% to 79.0\%), confirming that positional information is beneficial but not the sole driver of the gains. Finally, sweeping the redundancy penalty $\lambda$ over several orders of magnitude produces only mild fluctuations in accuracy (staying near 79–80\% except at the largest values), which implies that the contrastive redundancy regularizer is stable and that our default choice of $\lambda$ strikes a reasonable balance between disentanglement and predictive accuracy.
\renewcommand{\thesection}{A4}
\renewcommand{\thesubsection}{A4.\arabic{subsection}}
\section{Representation Quality via t-SNE}
\begin{figure*}[htbp]
\centering
\subfloat[t-SNE of ZAYAN-CL feature embeddings.\label{fig:tsne_zayan_cl}]{
  \includegraphics[width=0.45\textwidth]{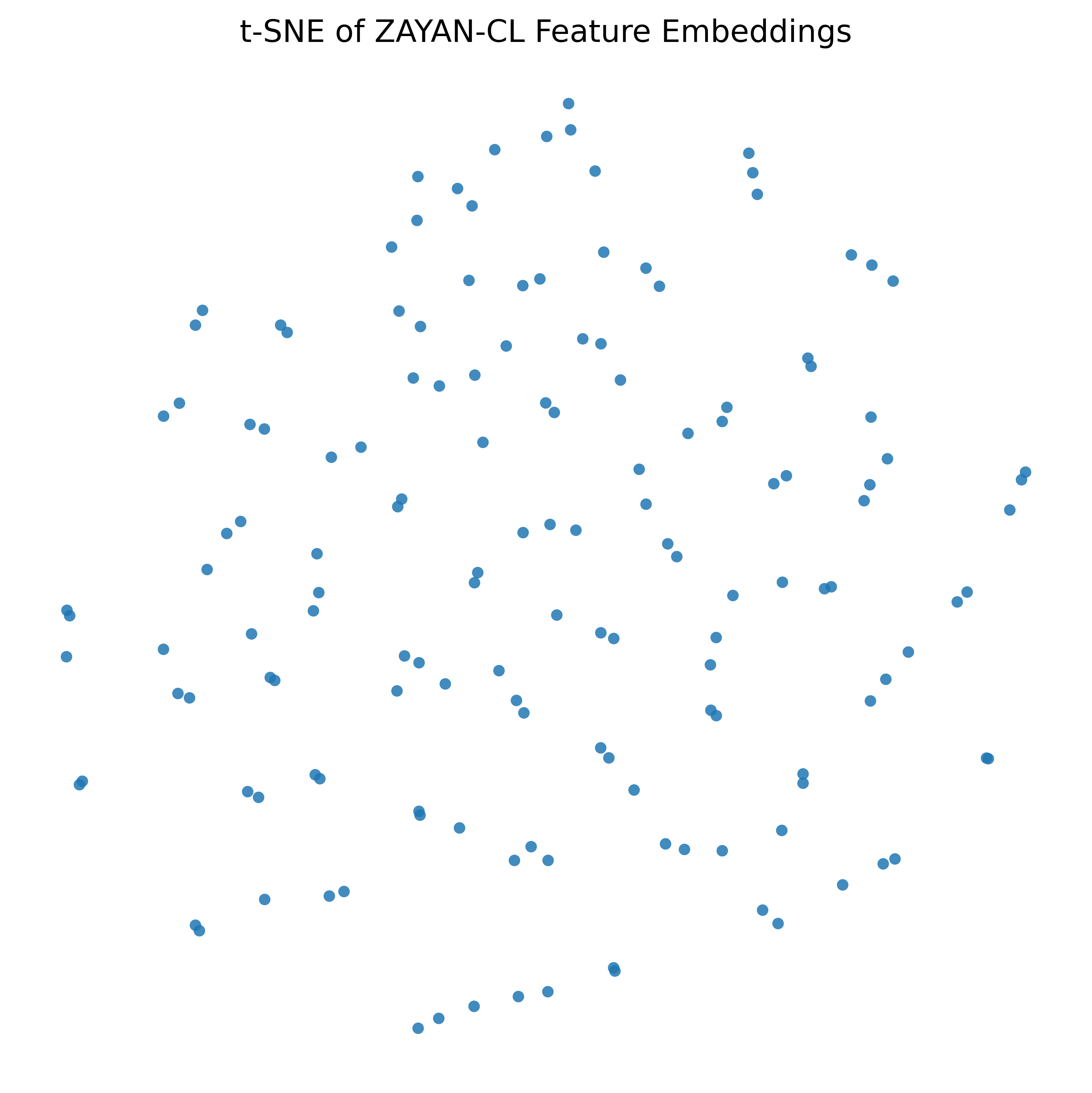}
}
\hfill
\subfloat[t-SNE of ZAYAN-T sample embeddings on the test set.\label{fig:tsne_zayan_t}]{
  \includegraphics[width=0.45\textwidth]{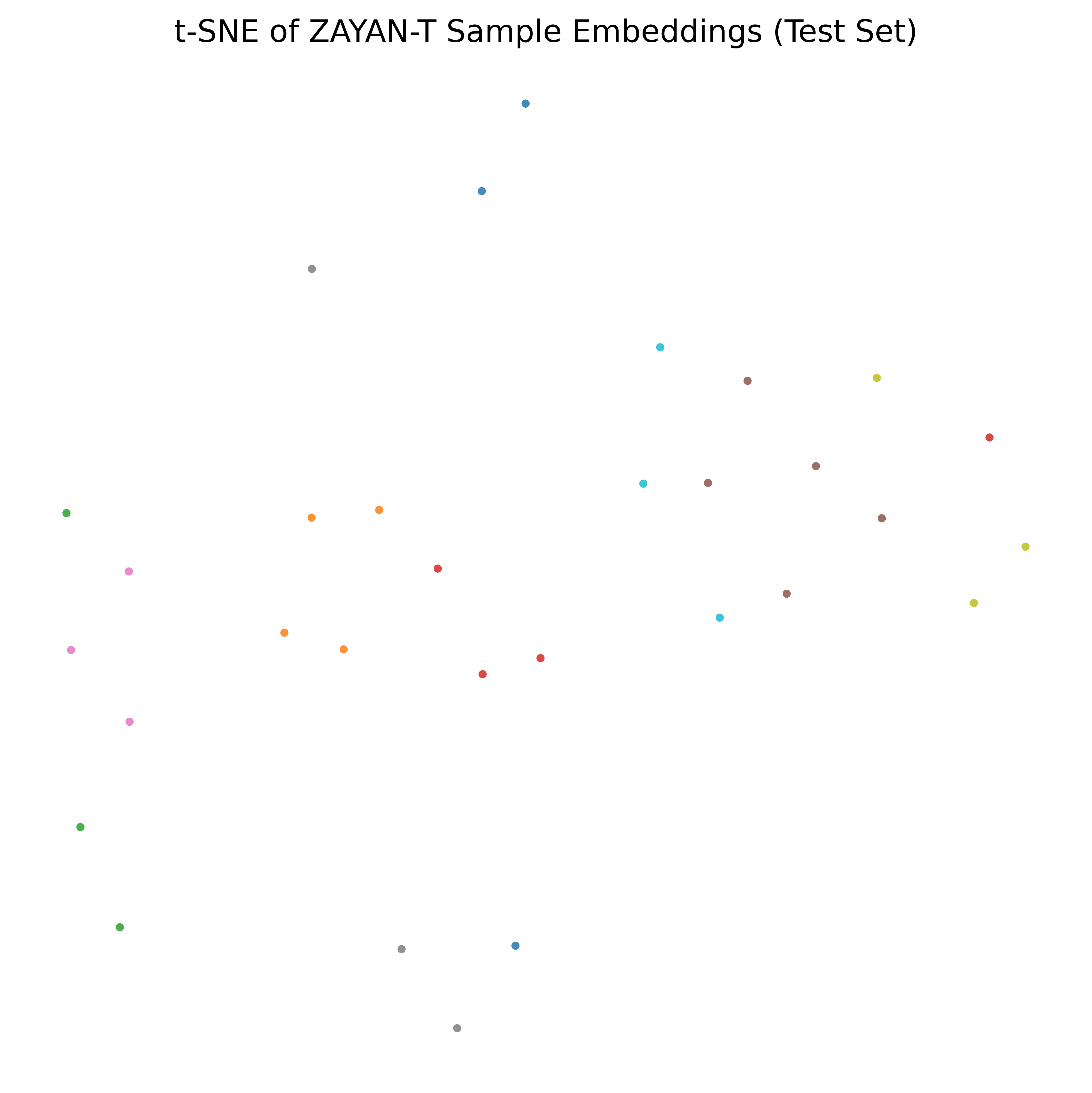}
}
\caption{t-SNE visualizations of ZAYAN feature and sample embeddings.}
\label{fig:tsne_zayan}
\end{figure*}

\renewcommand{\thesection}{\arabic{section}}
\label{ab7}
Figure~\ref{fig:tsne_zayan} compares the geometry of ZAYAN’s feature and sample representations on the Urban land cover dataset. Panel~\ref{fig:tsne_zayan_cl} shows the t-SNE projection of ZAYAN-CL feature embeddings, which are well spread in the plane without obvious collapse, suggesting that the contrastive objective encourages diverse, disentangled feature directions. Panel~\ref{fig:tsne_zayan_t} depicts the t-SNE projection of ZAYAN-T test-sample embeddings, colored by the ground-truth classes; despite some overlap at the boundaries, points from the same class tend to occupy nearby regions, indicating that the Transformer head preserves the diversity of the feature space while organizing samples into class-aware clusters that support accurate downstream prediction.
\renewcommand{\thesection}{A5}
\renewcommand{\thesubsection}{A5.\arabic{subsection}}
\section{Sanity and Stress Diagnostics}
\begin{figure*}[htbp]
  \centering
  \begin{minipage}[t]{0.32\textwidth}
    \centering
    \includegraphics[width=\linewidth]{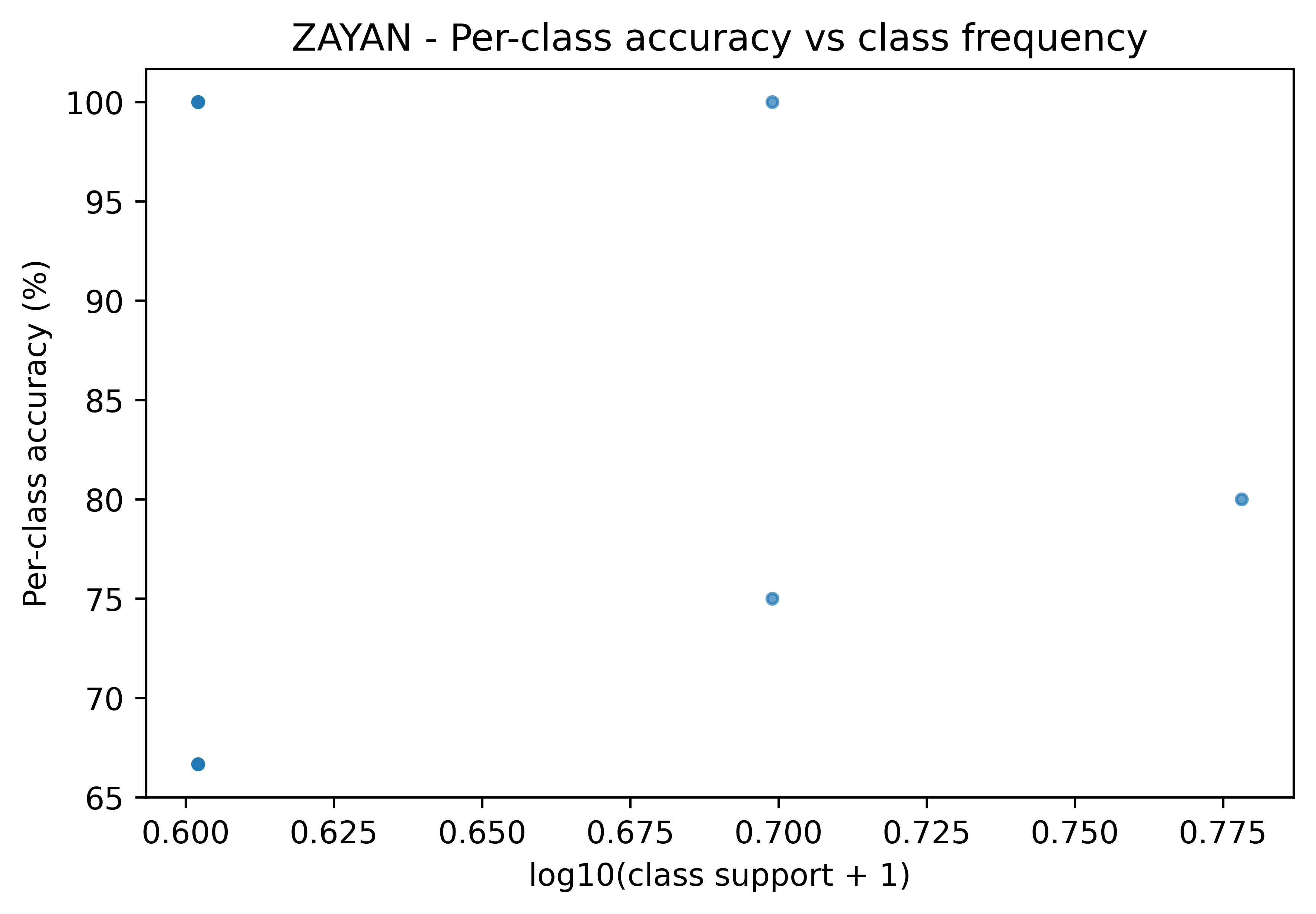}
  \end{minipage}\hfill
  \begin{minipage}[t]{0.32\textwidth}
    \centering
    \includegraphics[width=\linewidth]{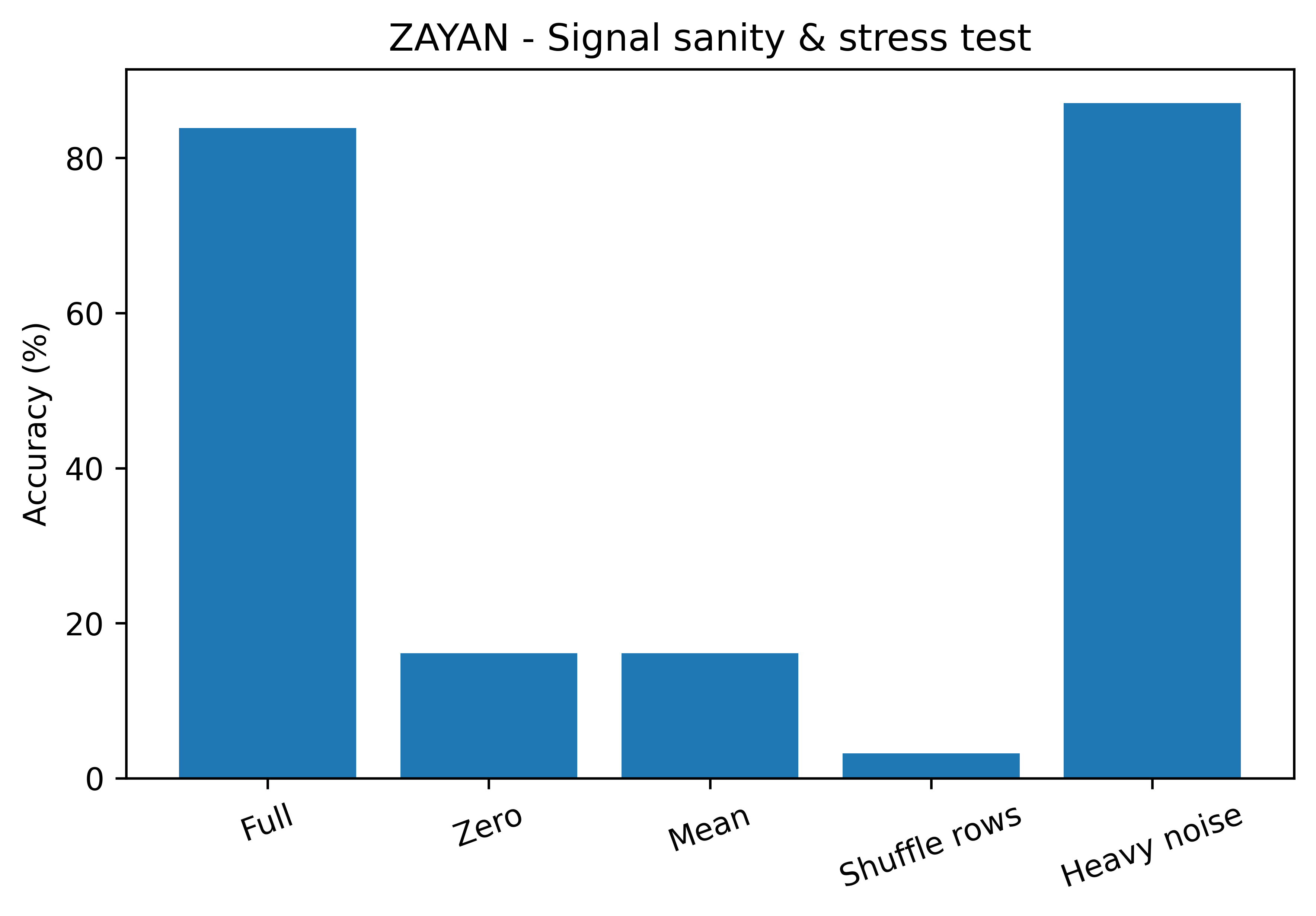}
  \end{minipage}\hfill
  \begin{minipage}[t]{0.32\textwidth}
    \centering
    \includegraphics[width=\linewidth]{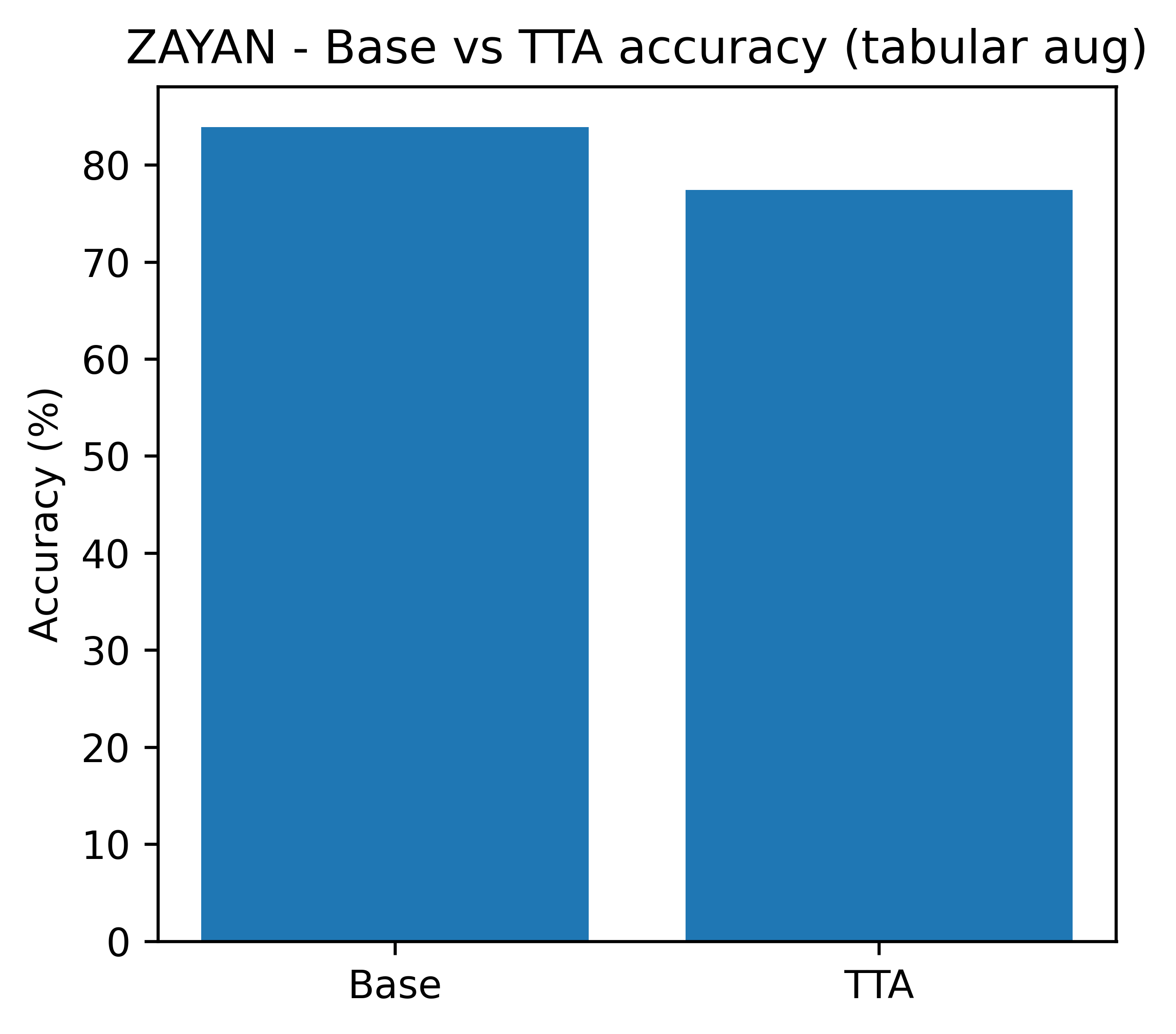}
  \end{minipage}
  \caption{Sanity and stress diagnostics for ZAYAN on the Urban land cover dataset.}
  \label{fig:zayan_sanity_stress}
\end{figure*}

\renewcommand{\thesection}{\arabic{section}}
\label{ab9}
Figure~\ref{fig:zayan_sanity_stress} and Tables~\ref{tab:zayan_signal_modes}-\ref{tab:zayan_per_class} summarize a set of sanity and stress tests for ZAYAN on the Urban land cover dataset. When evaluated on intact features, ZAYAN attains $83.87\%$ accuracy, whereas zeroing out all features, replacing them with global means, or shuffling rows across the test set collapses performance to near-chance levels ($16.13\%$, $16.13\%$, and $3.23\%$,
respectively), confirming that the model relies on meaningful tabular structure rather than label leakage or spurious priors. Injecting heavy Gaussian noise degrades the signal but still yields $87.10\%$ accuracy, suggesting that the learned representations retain a degree of robustness to perturbations. Test-time augmentation (TTA) with five tabular perturbations and majority voting produces a slightly lower accuracy ($77.42\%$) than the single-pass baseline, but only $12.90\%$ of samples change their predicted label under augmentation, indicating that ZAYAN’s decisions are largely stable. Finally, the per-class analysis shows that both head and medium-frequency classes achieve comparable mean accuracies (approximately $85\%$ and $83\%$, respectively), with minority classes occasionally dropping to $\sim 67\%$ accuracy but no systematic collapse on low-support categories. Together, these diagnostics support that ZAYAN is genuinely exploiting the tabular sensing signal, is reasonably robust to noise, and maintains balanced performance across classes despite the moderate label imbalance in Urban land cover.
\begin{table}[htbp]
\centering
\small
\caption{Signal sanity and stress tests for ZAYAN on the Urban land cover test set.}
\label{tab:zayan_signal_modes}
\begin{tabular}{lcc}
\toprule
Mode & Accuracy (\%) & $n$ \\
\midrule
Full (original features) & 83.87 & 31 \\
Zero (all features set to 0) & 16.13 & 31 \\
Mean (features replaced by global mean) & 16.13 & 31 \\
Shuffle rows (row-wise permutation) & 3.23 & 31 \\
Heavy noise (strong Gaussian perturbation) & 87.10 & 31 \\
\bottomrule
\end{tabular}
\end{table}
\begin{table}[htbp]
\centering
\small
\caption{Augmentation consistency and test-time augmentation (TTA) for ZAYAN.}
\label{tab:zayan_tta}
\begin{tabular}{lc}
\toprule
Quantity & Value \\
\midrule
Base accuracy (no TTA) & 83.87\% \\
TTA accuracy (5 tabular augmentations, majority vote) & 77.42\% \\
Fraction of samples with any label change across augmentations & 12.90\% \\
\bottomrule
\end{tabular}
\end{table}
\begin{table}[htbp]
\centering
\small
\caption{Per-class accuracy and bucketed summary for ZAYAN on Urban land cover.}
\label{tab:zayan_per_class}
\begin{tabular}{ccc}
\toprule
Class ID & Support & Accuracy (\%) \\
\midrule
0 & 3 & 100.00 \\
1 & 4 & 100.00 \\
2 & 3 & 66.67 \\
3 & 4 & 75.00 \\
4 & 5 & 80.00 \\
5 & 3 & 100.00 \\
6 & 3 & 100.00 \\
7 & 3 & 66.67 \\
8 & 3 & 66.67 \\
\bottomrule
\end{tabular}

\vspace{0.5em}
\small
Bucketed by class support:
\emph{head} (3 classes) - mean accuracy $85.00\%$;
\emph{medium} (6 classes) - mean accuracy $83.33\%$.
\end{table}
\renewcommand{\thesection}{A6}
\renewcommand{\thesubsection}{A6.\arabic{subsection}}
\section{Additional Reliability and Interpretability Diagnostics }
\begin{figure*}[htbp]
\centering

\begin{minipage}{0.48\textwidth}
  \centering
  \includegraphics[width=\linewidth]{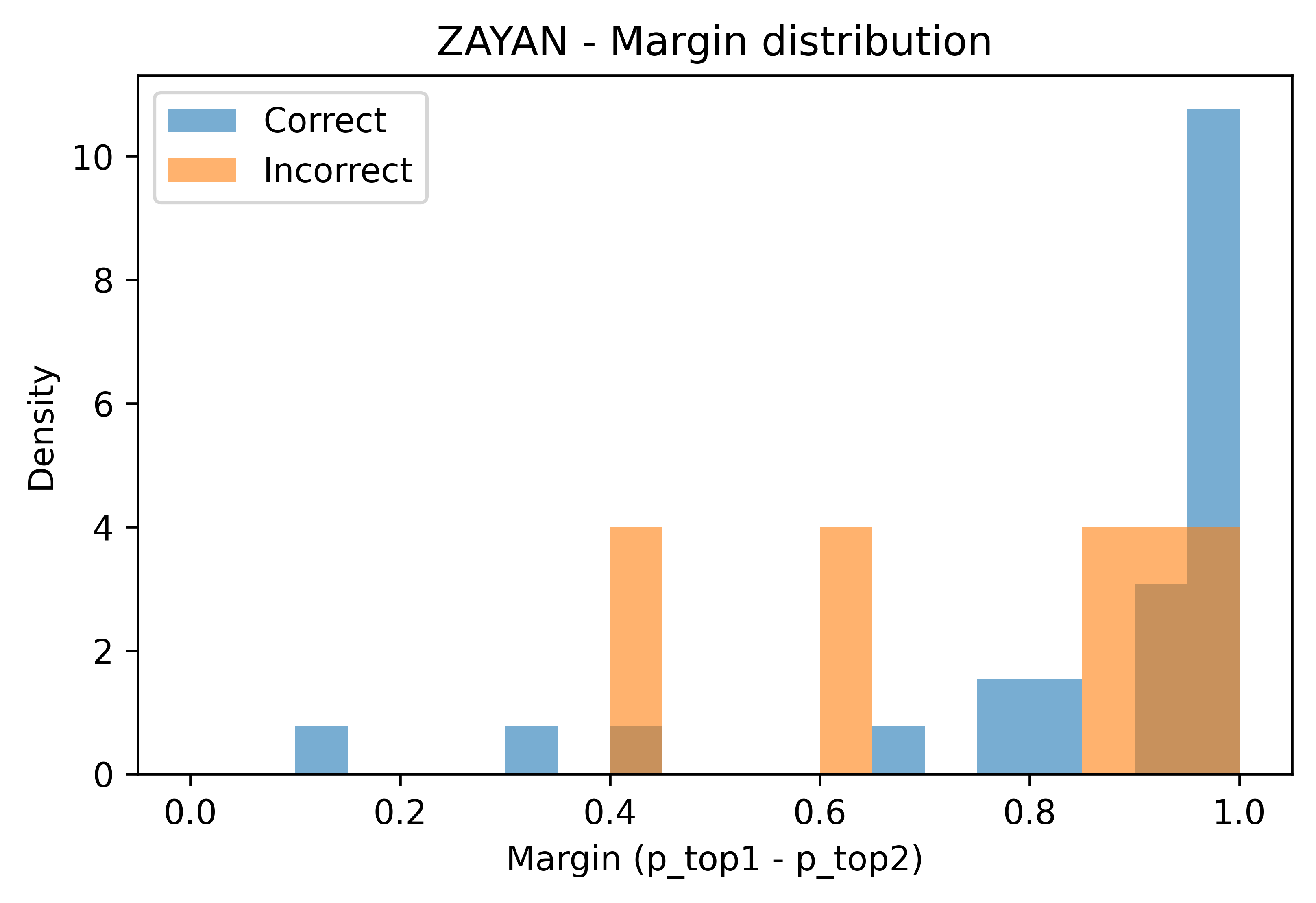}
\end{minipage}\hfill
\begin{minipage}{0.48\textwidth}
  \centering
  \includegraphics[width=\linewidth]{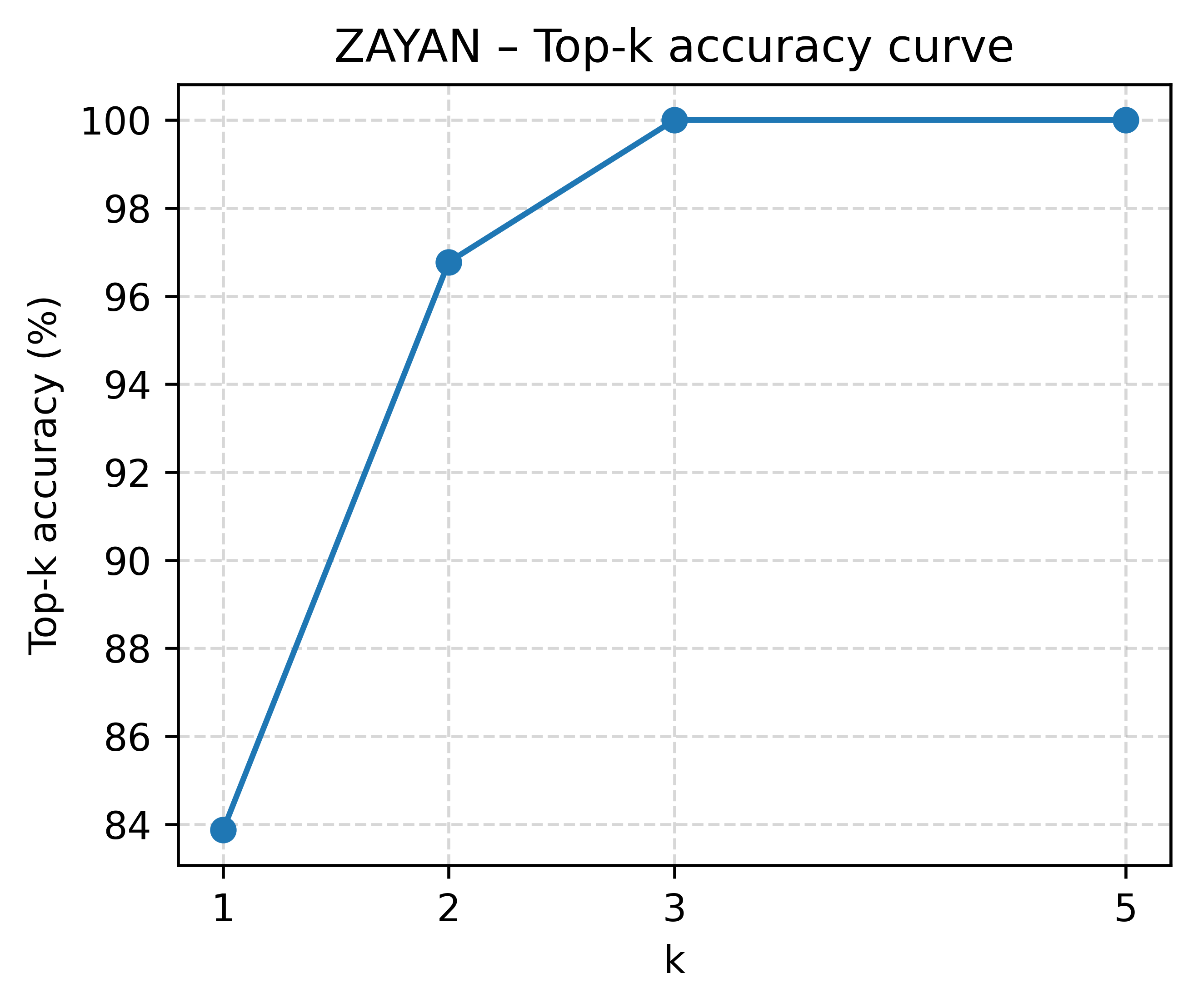}
\end{minipage}

\medskip

\begin{minipage}{0.48\textwidth}
  \centering
  \includegraphics[width=\linewidth]{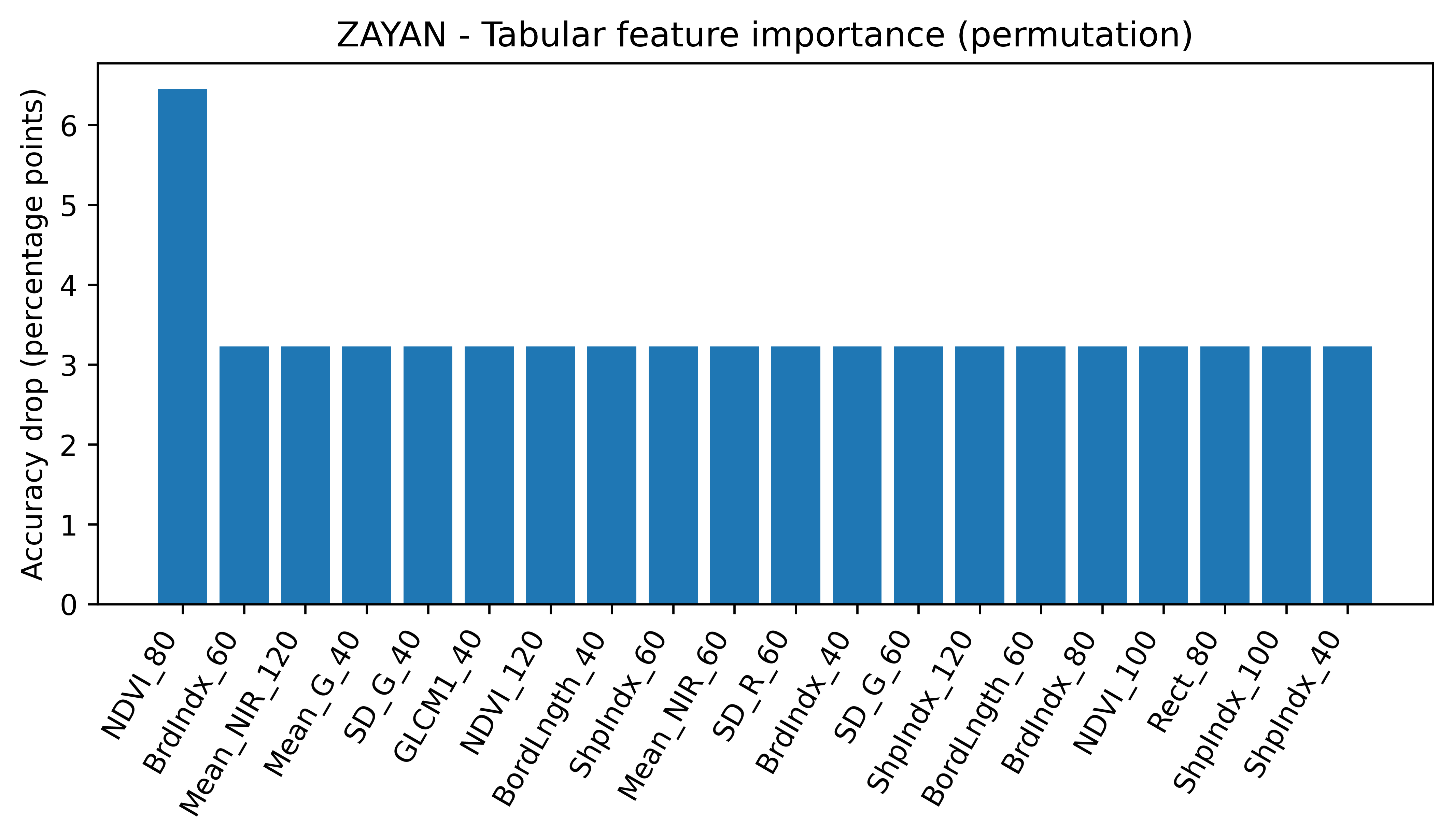}
\end{minipage}\hfill
\begin{minipage}{0.48\textwidth}
  \centering
  \includegraphics[width=\linewidth]{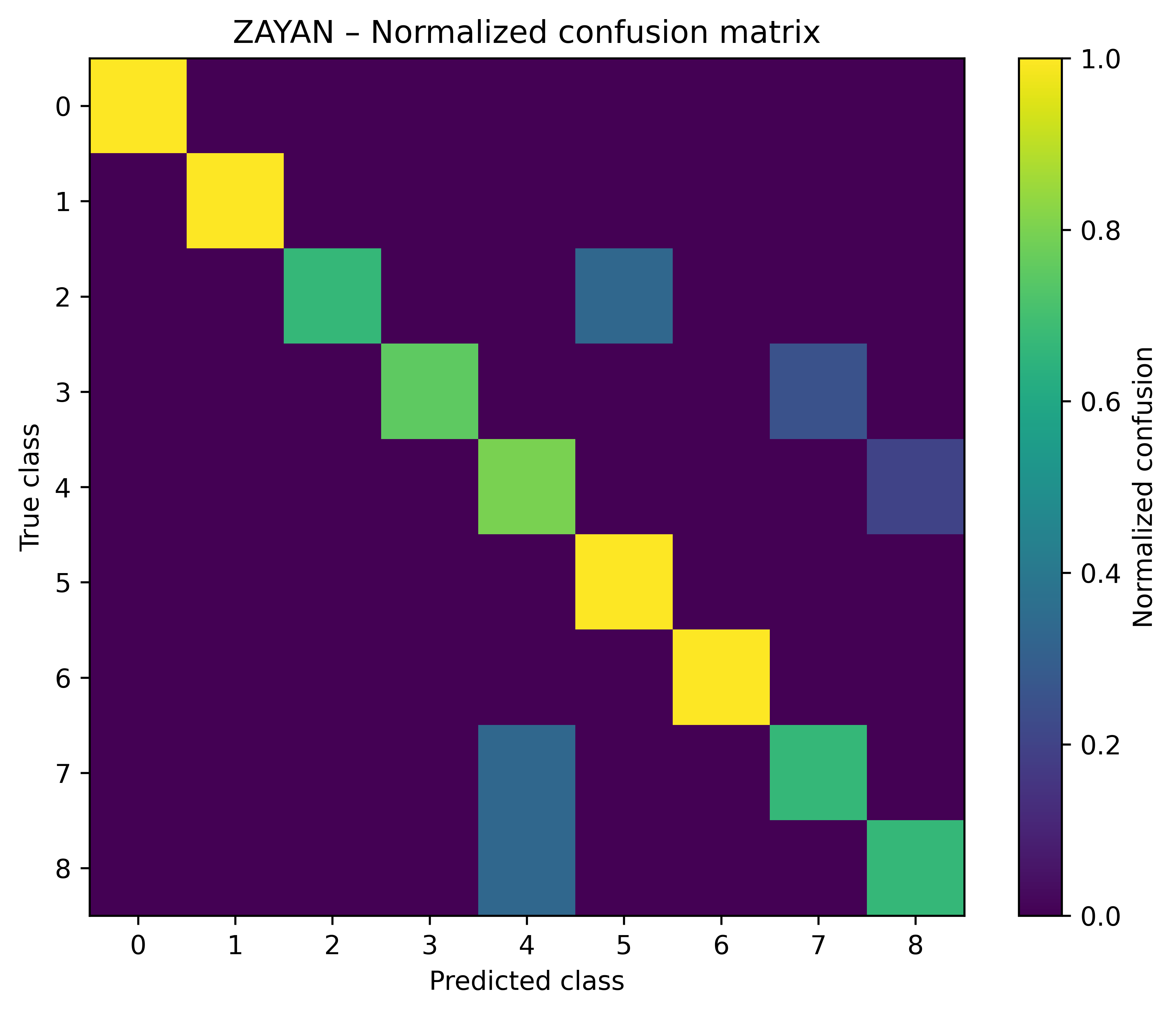}
\end{minipage}

\caption{Additional reliability and interpretability diagnostics for ZAYAN on the
Urban land cover test set. (a) Margin distribution comparing correct vs.\ incorrect
predictions. (b) Top-$k$ accuracy curve. (c) Permutation-based tabular feature
importance for the top 15 features by accuracy drop. (d) Normalized confusion
matrix highlighting the main confusions across the nine land-cover classes.}
\label{fig:zayan_rel_diagnostics}
\end{figure*}

\renewcommand{\thesection}{\arabic{section}}
\label{ab10}
Figure~\ref{fig:zayan_rel_diagnostics}(a) shows the margin distribution ($p_{\text{top1}} - p_{\text{top2}}$) for correct versus incorrect predictions. Correct test samples concentrate near high margins (mean $0.858$), while the errors exhibit slightly smaller but still reasonably separated margins (mean $0.769$), suggesting that ZAYAN is generally well-calibrated and that a simple margin-based threshold could flag a subset of uncertain cases. The top-$k$ accuracy curve in Fig.~\ref{fig:zayan_rel_diagnostics}(b) further confirms that most mistakes are ``near misses'': the Top-1 accuracy is $83.87\%$, but the ground-truth label appears within the Top-2 predictions $96.77\%$ of the time and within the Top-3 predictions for \emph{all} test samples (Table~\ref{tab:zayan_reliability_summary}). This indicates that ZAYAN's ranking over classes is highly reliable even when the top prediction is wrong.

Figure~\ref{fig:zayan_rel_diagnostics}(c) reports permutation-based feature importance for the 15 most influential tabular features. The dominant role of NDVI at the 80\,m scale (NDVI\_80; $6.45$ percentage-point accuracy drop) is consistent with the domain expectation that vegetation indices are crucial for land-cover discrimination. Several additional multi-scale descriptors-border indices (BrdIndx\_40/60), shape indices (ShpIndx\_60/120), NIR band statistics (Mean\_NIR\_60/120), green-band and texture features (Mean\_G\_40, SD\_G\_40, GLCM1\_40), and border length (BordLngth\_40/60)-also incur non trivial drops ($3.23$ points each), highlighting that ZAYAN leverages a diverse set of physically meaningful features rather than overfitting to a single cue (Table~\ref{tab:zayan_perm_importance}).

The normalized confusion matrix in Fig.~\ref{fig:zayan_rel_diagnostics}(d) shows that most probability mass lies on the diagonal, with only a few isolated off-diagonal entries. The top-5 confusions summarized in Table~\ref{tab:zayan_confusions} involve pairs such as class~8$\rightarrow$4 and class~4$\rightarrow$8, or class~2$\rightarrow$5 and class~3$\rightarrow$7, which correspond to land-cover types that are visually and spectrally similar. Together, these diagnostics indicate that ZAYAN achieves strong overall accuracy on the Urban land cover dataset, makes mostly structured and interpretable errors, and relies on domain-plausible tabular features across multiple spatial scales.
\begin{table}[htbp]
\centering
\small
\caption{Summary of reliability diagnostics for ZAYAN on the Urban land cover test set.}
\label{tab:zayan_reliability_summary}
\begin{tabular}{lc}
\toprule
Metric & Value \\
\midrule
\# test samples $N$         & 31 \\
\# tabular features $m$     & 147 \\
Top-1 accuracy              & 83.87\% \\
Top-2 accuracy              & 96.77\% \\
Top-3 accuracy              & 100.00\% \\
Top-5 accuracy              & 100.00\% \\
Mean margin (correct)       & 0.858 \\
Mean margin (incorrect)     & 0.769 \\
\bottomrule
\end{tabular}
\end{table}
\begin{table}[htbp]
\centering
\small
\caption{Most frequent confusions for ZAYAN on the Urban land cover test set 
(true class $\rightarrow$ predicted class).}
\label{tab:zayan_confusions}
\begin{tabular}{ccc}
\toprule
True class & Predicted class & Count \\
\midrule
8 & 4 & 1 \\
7 & 4 & 1 \\
4 & 8 & 1 \\
3 & 7 & 1 \\
2 & 5 & 1 \\
\bottomrule
\end{tabular}
\end{table}
\begin{table*}[htbp]
\centering
\small
\caption{Top-15 most important tabular features for ZAYAN on the Urban land cover
test set under permutation testing. Baseline accuracy is 83.87\%; we report the
absolute accuracy drop in percentage points when each feature is permuted.}
\label{tab:zayan_perm_importance}
\begin{tabular}{clc}
\toprule
Rank & Feature      & Accuracy drop (pts) \\
\midrule
1  & NDVI\_80      & 6.45 \\
2  & BrdIndx\_60   & 3.23 \\
3  & Mean\_NIR\_120& 3.23 \\
4  & Mean\_G\_40   & 3.23 \\
5  & SD\_G\_40     & 3.23 \\
6  & GLCM1\_40     & 3.23 \\
7  & NDVI\_120     & 3.23 \\
8  & BordLngth\_40 & 3.23 \\
9  & ShpIndx\_60   & 3.23 \\
10 & Mean\_NIR\_60 & 3.23 \\
11 & SD\_R\_60     & 3.23 \\
12 & BrdIndx\_40   & 3.23 \\
13 & SD\_G\_60     & 3.23 \\
14 & ShpIndx\_120  & 3.23 \\
15 & BordLngth\_60 & 3.23 \\
\bottomrule
\end{tabular}
\end{table*}
\renewcommand{\thesection}{A7}
\renewcommand{\thesubsection}{A7.\arabic{subsection}}
\section{Theory-Inspired Representation Diagnostics}
\begin{figure*}[htbp]
\centering
\begin{subfigure}{0.32\textwidth}
    \centering
    \includegraphics[width=\linewidth]{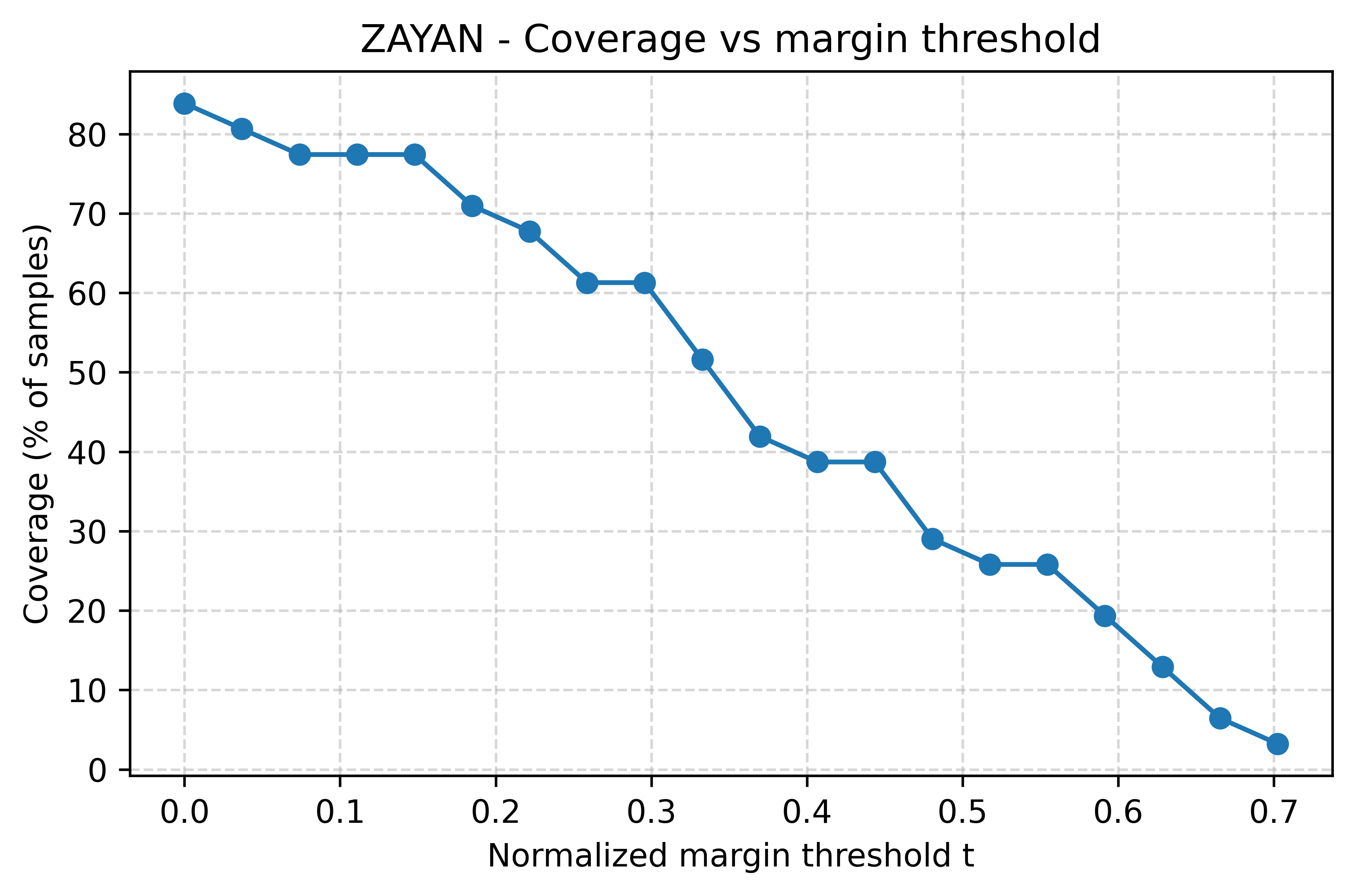}
    \caption{Coverage vs.\ margin threshold.}
    \label{fig:zayan_cov_margin}
\end{subfigure}
\begin{subfigure}{0.32\textwidth}
    \centering
    \includegraphics[width=\linewidth]{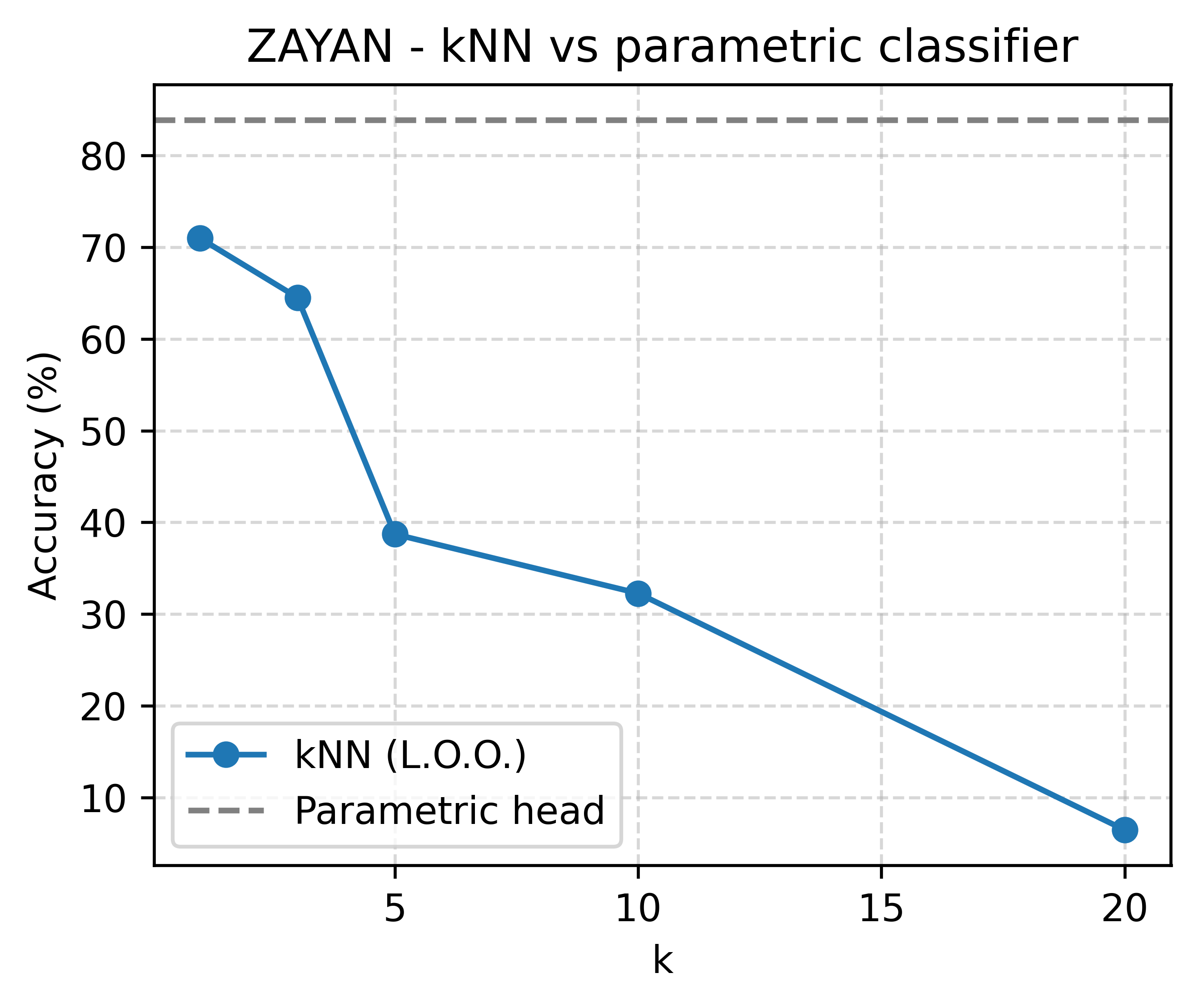}
    \caption{$k$NN vs.\ parametric head.}
    \label{fig:zayan_knn_vs_head}
\end{subfigure}
\begin{subfigure}{0.32\textwidth}
    \centering
    \includegraphics[width=\linewidth]{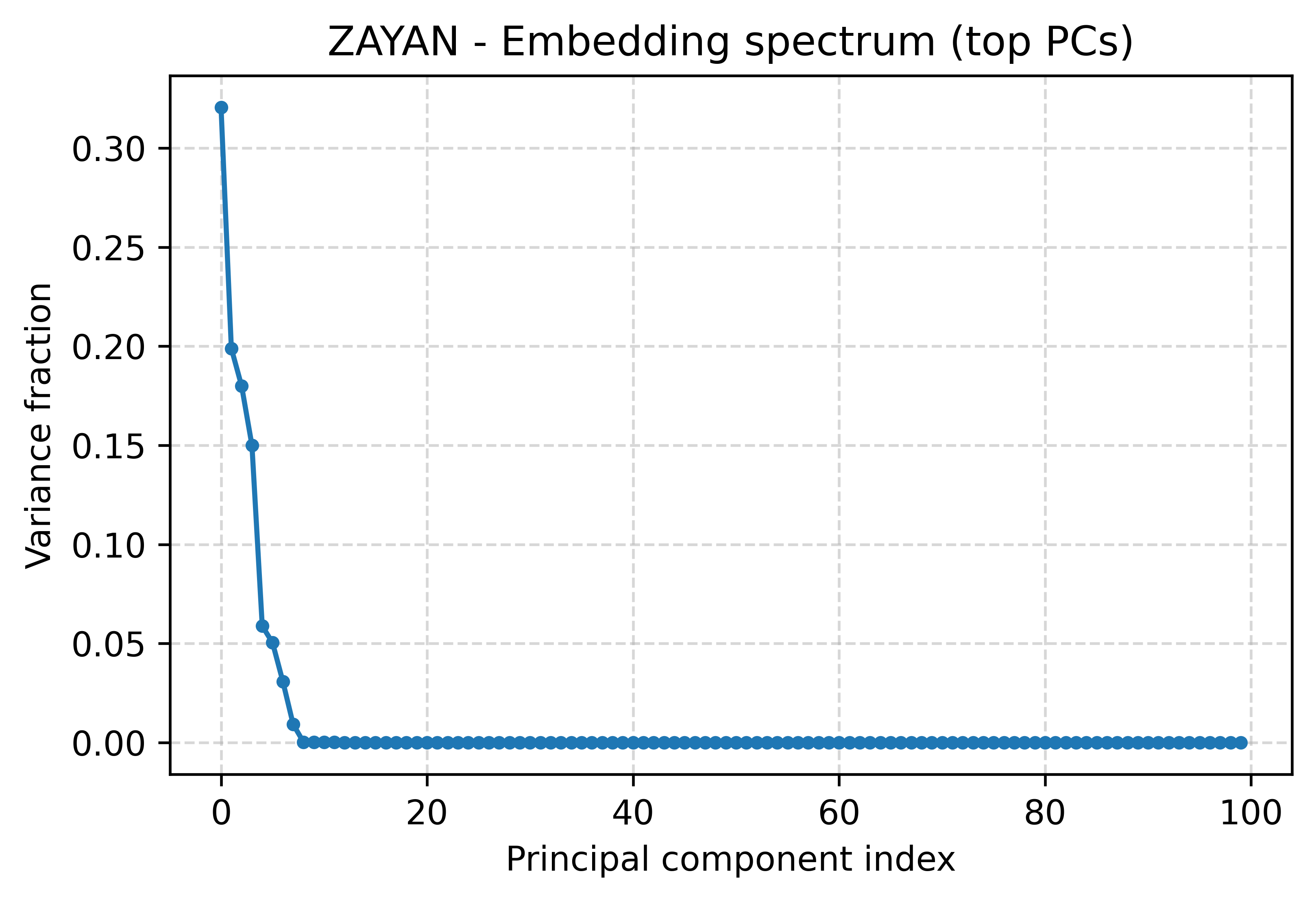}
    \caption{Embedding spectrum (top PCs).}
    \label{fig:zayan_spectrum}
\end{subfigure}

\vspace{0.5em}

\begin{subfigure}{0.32\textwidth}
    \centering
    \includegraphics[width=\linewidth]{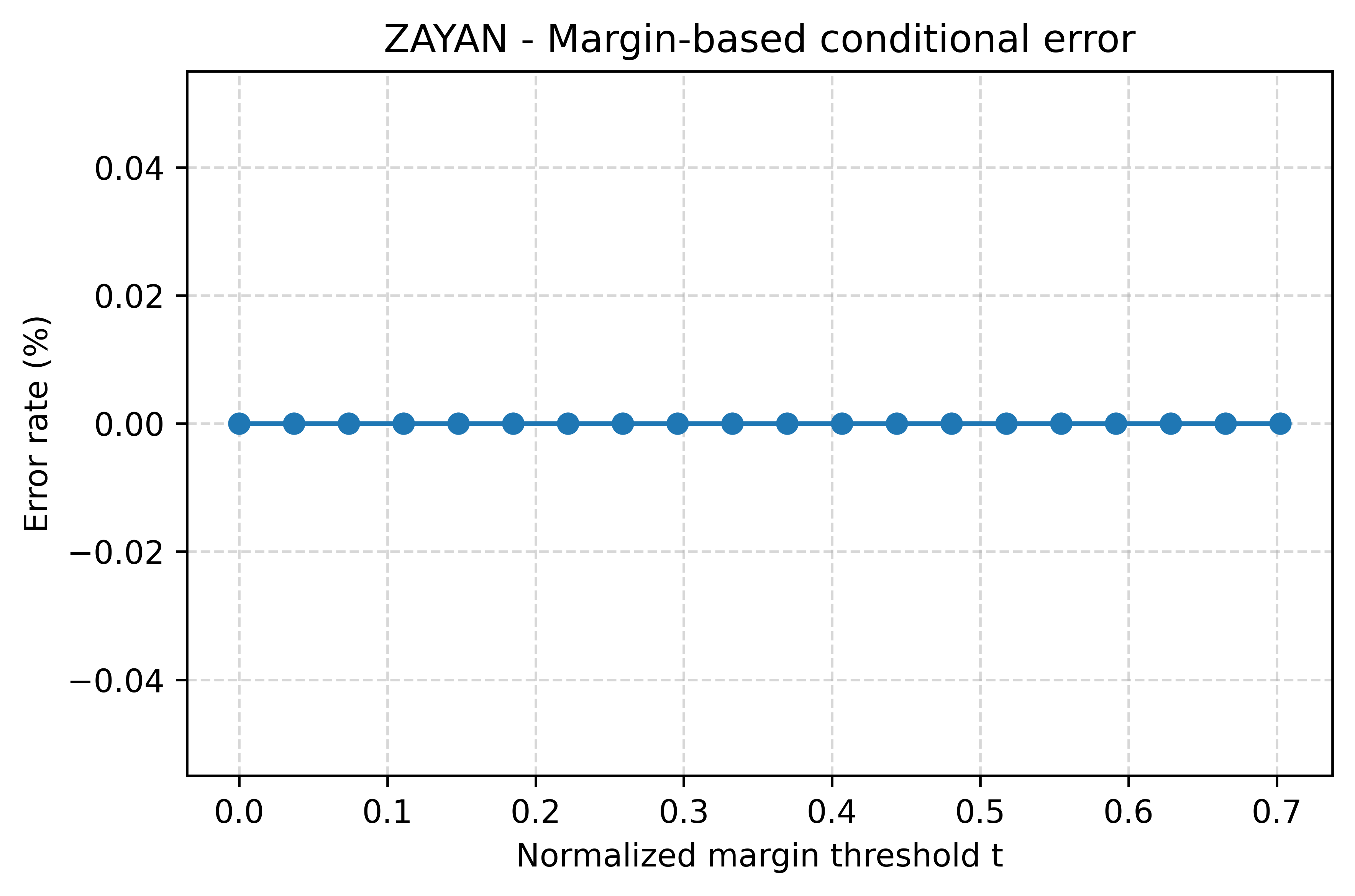}
    \caption{Margin-based conditional error.}
    \label{fig:zayan_margin_error}
\end{subfigure}
\begin{subfigure}{0.32\textwidth}
    \centering
    \includegraphics[width=\linewidth]{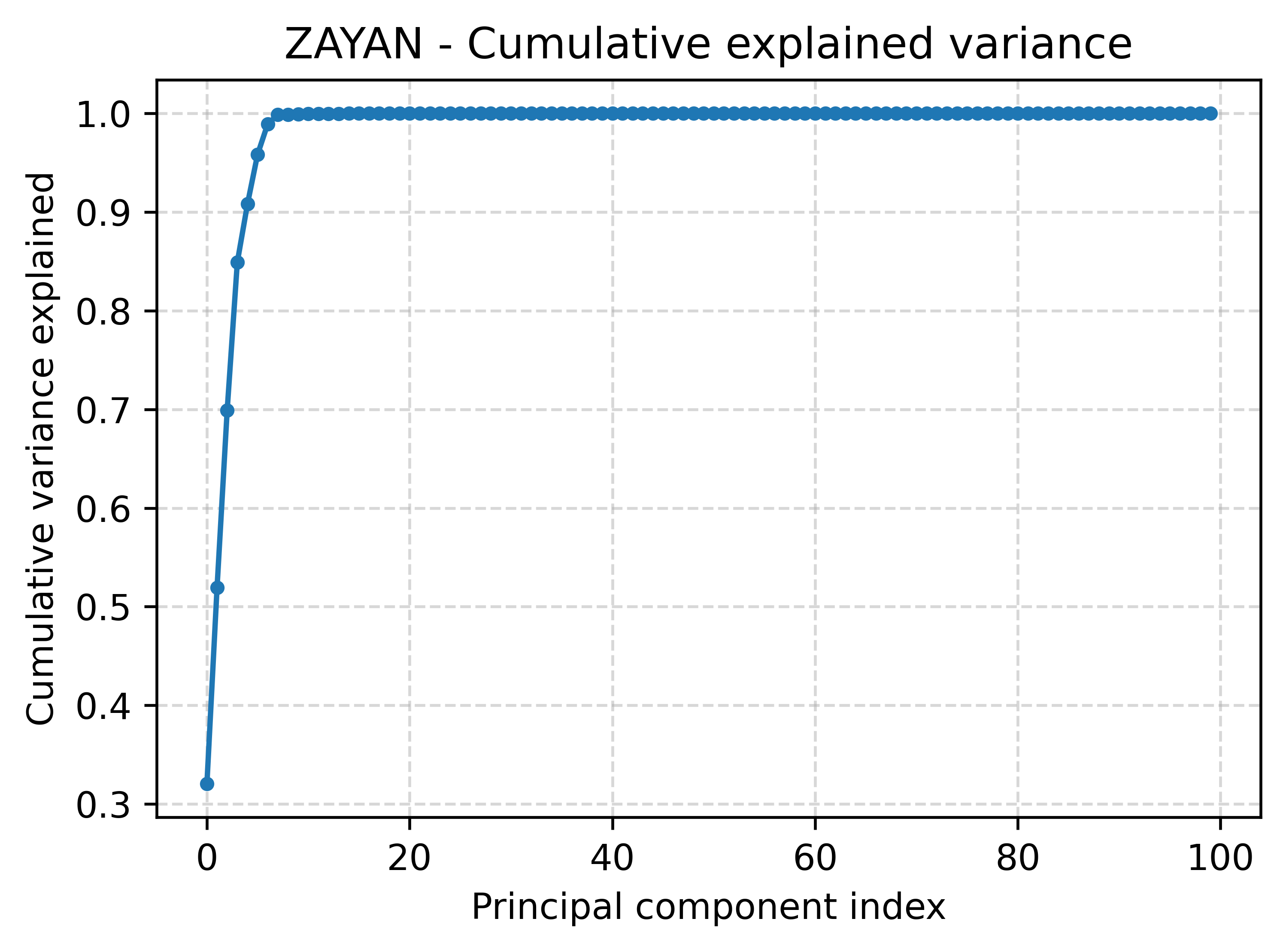}
    \caption{Cumulative explained variance.}
    \label{fig:zayan_cumvar}
\end{subfigure}
\caption{Theory-inspired diagnostics for ZAYAN on the Urban land cover test set ($N{=}31$, embedding dimension $128$). Panels (a,b) relate prediction margins to selective prediction and non-parametric baselines, while (c–e) analyze the geometry of the learned embedding via PCA.}
\label{fig:zayan_theory_diagnostics}
\end{figure*}

\renewcommand{\thesection}{\arabic{section}}
\label{ab11}
We examine ZAYAN through a set of theory-inspired diagnostics on the Urban land cover test set ($N{=}31$). Figure~\ref{fig:zayan_theory_diagnostics}\subref{fig:zayan_cov_margin} shows a coverage-margin curve, where we retain only predictions whose normalized margin $m = (p_{\text{top1}} - p_{\text{top2}})$ exceeds a threshold $t$. As $t$ increases, coverage drops smoothly from about $84\%$ to below $10\%$, but in this small test set the empirical conditional error above each threshold is essentially $0\%$ (Fig.~\ref{fig:zayan_theory_diagnostics}\subref{fig:zayan_margin_error}), consistent with a large separation between the correct and competing classes (mean normalized margins $0.40$ vs.\ $-0.29$; Table~\ref{tab:zayan_knn_margin}). Comparing the parametric classifier head to leave-one-out $k$NN in the embedding space (Fig.~\ref{fig:zayan_theory_diagnostics}\subref{fig:zayan_knn_vs_head}), we observe that $k$NN accuracies degrade quickly as $k$ grows (from $70.97\%$ at $k{=}1$ to $6.45\%$ at $k{=}20$), and remain consistently below the parametric top-1 accuracy of $83.87\%$, indicating that ZAYAN learns a geometry that benefits from a global decision surface rather than a purely local neighborhood rule. PCA on the 128-dimensional embeddings reveals a very low effective dimension: the participation-ratio effective dimension is $d_{\text{eff}}{\approx}4.9$, and only $k_{\text{PC}}{=}2,4,5,6,8$ components are needed to explain ${\approx}50\%,80\%,90\%,95\%,99\%$ of the variance, respectively (Table~\ref{tab:zayan_geometry}, Fig.~\ref{fig:zayan_theory_diagnostics}\subref{fig:zayan_spectrum},\subref{fig:zayan_cumvar}). Together, these results suggest that ZAYAN compresses the high-dimensional tabular features into a low-dimensional, well-separated manifold where a simple linear head achieves strong performance, while confidence margins provide a meaningful knob for selective prediction without sacrificing reliability on the retained samples.
\begin{table}[htbp]
\centering
\small
\caption{Geometric diagnostics for ZAYAN embeddings on the Urban land cover test set. The effective dimension is given by the participation ratio; $k_{\text{PC}}$ is the number of principal components required to exceed the target explained variance.}
\label{tab:zayan_geometry}
\begin{tabular}{lcc}
\toprule
Metric & Value & Notes \\
\midrule
Embedding dimension $d$ & $128$ & Transformer feature size \\
Effective dimension $d_{\text{eff}}$ & $4.9$ & Participation ratio \\
\midrule
Target variance & $k_{\text{PC}}$ & \\
\midrule
$0.50$ & $2$ & \\
$0.80$ & $4$ & \\
$0.90$ & $5$ & \\
$0.95$ & $6$ & \\
$0.99$ & $8$ & \\
\bottomrule
\end{tabular}
\end{table}
\begin{table}[htbp]
\centering
\small
\caption{Neighborhood-based and margin-based diagnostics for ZAYAN on the Urban land cover test set. Leave-one-out ($\mathrm{LOO}$) $k$NN is applied in the learned embedding space.}
\label{tab:zayan_knn_margin}
\begin{tabular}{lcc}
\toprule
Metric & Value & Notes \\
\midrule
Top-1 accuracy (\%) & $83.87$ & Parametric classifier head \\
Mean normalized margin (correct) & $0.40$ & Higher is better \\
Mean normalized margin (incorrect) & $-0.29$ & Lower is better \\
\midrule
$k$ (LOO $k$NN) & Accuracy (\%) & In embedding space \\
\midrule
$1$  & $70.97$ & Best $k$ among tested \\
$3$  & $64.52$ &  \\
$5$  & $38.71$ &  \\
$10$ & $32.26$ &  \\
$20$ & $6.45$  &  \\
\bottomrule
\end{tabular}
\end{table}
\renewcommand{\thesection}{A8}
\renewcommand{\thesubsection}{A8.\arabic{subsection}}
\section{Turing-Style Human-Model Evaluation}
\renewcommand{\thesection}{\arabic{section}}
\label{ab12}
To facilitate human-in-the-loop assessment of ZAYAN, we generate a small ``Turing-test'' kit on the Urban land cover test set. Starting from the $N{=}31$ held-out samples, we randomly sample up to $60$ indices; in this case all $31$ unique test points are included in the Turing subset (Table~\ref{tab:zayan_turing_config}). For each selected instance we export an answer-sheet CSV containing the global index, true class label, ZAYAN's predicted label and confidence, and the full set of $147$ standardized tabular features (Table~\ref{tab:zayan_turing_example}). The sheet also contains empty columns (\texttt{human\_label\_id} or \texttt{human\_label\_name}) that can be filled by a human expert. The intended protocol is as follows: (i) the experimenter hides the model-prediction columns if desired, (ii) a human annotator inspects only the feature values and assigns a label to each row, and (iii) our scoring script reloads the completed CSV and automatically computes human accuracy, human-model agreement rates, and confusion statistics on the same subset. On the current run, before collecting any human labels, ZAYAN achieves a top-1 accuracy of $83.87\%$ on the Turing subset, matching its overall test accuracy (Table~\ref{tab:zayan_turing_config}). The exported answer sheet and scoring code therefore provide a lightweight mechanism for future user studies where domain experts can be compared directly against ZAYAN on exactly the same tabular instances, in a Turing-style setting where only the feature vectors, not the underlying images or geospatial context are visible.
\begin{table}[htbp]
\centering
\small
\caption{Configuration of the Turing-style human-model evaluation for ZAYAN on the Urban land cover test set.}
\label{tab:zayan_turing_config}
\begin{tabular}{lcc}
\toprule
Quantity & Value & Notes \\
\midrule
\# test samples $N$ & $31$ &
Size of held-out set \\
Turing subset size & $31$ &
Sampled from test set (target $60$ indices) \\
Embedding dimension & $128$ &
Transformer feature size \\
Answer sheet path &
\texttt{zayan\_turing\_test/zayan\_turing\_sheet.csv} &
CSV for human annotators \\
Model top-1 accuracy (\%) &
$83.87$ &
On the Turing subset ($N{=}31$) \\
\bottomrule
\end{tabular}
\end{table}
\begin{table*}[htbp]
\centering
\small
\caption{Excerpt from the ZAYAN Turing-test answer sheet. For each sampled test instance we store the global index, true class, model prediction and confidence, a column to be filled by the human annotator, and the normalized tabular features.}
\label{tab:zayan_turing_example}
\begin{tabular}{ccccccccc}
\toprule
global\_index & true\_label & model\_pred & model\_conf. &
human\_label & BrdIndx & Area & Round & Bright \\
\midrule
0 & class\_1 & class\_1 & 0.994 & (to fill) & $-1.10$ & $1.60$ & $-0.93$ & $0.64$ \\
1 & class\_5 & class\_5 & 0.994 & (to fill) & $-0.68$ & $-0.61$ & $-0.97$ & $0.08$ \\
2 & class\_7 & class\_4 & 0.924 & (to fill) & $0.65$ & $0.19$ & $0.23$ & $-0.19$ \\
3 & class\_7 & class\_7 & 0.975 & (to fill) & $-0.15$ & $0.08$ & $-0.41$ & $0.92$ \\
4 & class\_4 & class\_8 & 0.723 & (to fill) & $-0.93$ & $-0.37$ & $-0.10$ & $-0.17$ \\
\bottomrule
\end{tabular}
\end{table*}
\renewcommand{\thesection}{A9}
\renewcommand{\thesubsection}{A9.\arabic{subsection}}
\section{Optuna-Level Diagnostics (Global Search Behavior)}
\renewcommand{\thesection}{\arabic{section}}
\label{ab15}
We ran a 150-trial Optuna search over the ZAYAN hyperparameters on the Urban land cover dataset. As summarized in Table~\ref{tab:optuna_global_summary}, the objective (5-fold CV accuracy) spans a wide range from $0.39$ to $0.85$, with a mean of $0.76$ and median of $0.79$, indicating that the search space contains both clearly suboptimal and near-optimal configurations. Individual trials complete in roughly $20$-$86$ seconds (mean $\approx 60$\,s), making it feasible to explore a reasonably rich configuration space. The best trial achieves a mean CV accuracy of $0.848$ with fold-wise scores between $0.77$ and $0.93$, yielding a $95\%$ confidence interval of $[0.788, 0.908]$ (Table~\ref{tab:optuna_best_fold}), which suggests that the optimum is not driven by a single lucky split.

\begin{figure*}[htbp]
\centering
\includegraphics[width=0.95\textwidth]{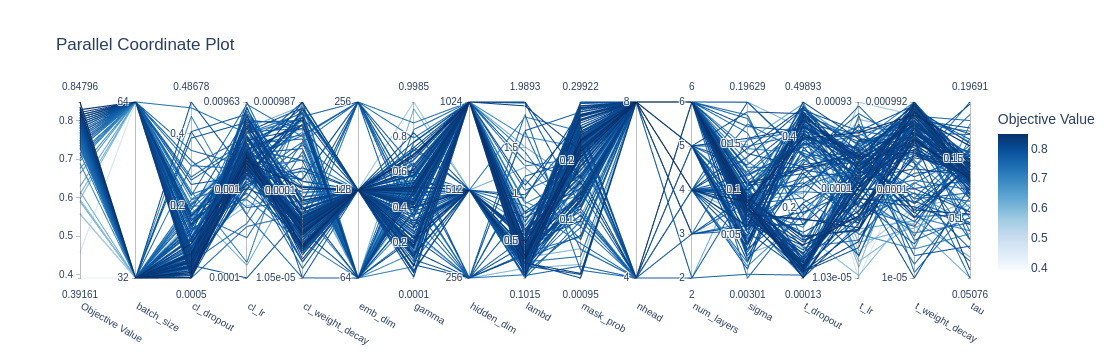}
\caption{Optuna parallel-coordinate plot over 150 trials for ZAYAN on the Urban land cover
dataset. Each polyline corresponds to a single trial and is colored by cross-validation
accuracy. High-performing trials concentrate around moderate contrastive dropout
(\texttt{cl\_dropout}), smaller Transformer dropout (\texttt{t\_dropout}), and
intermediate values of temperature~$\tau$ and Transformer weight decay.}
\label{fig:optuna_parallel}
\end{figure*}

\begin{figure*}[htbp]
\centering
\includegraphics[width=0.75\textwidth]{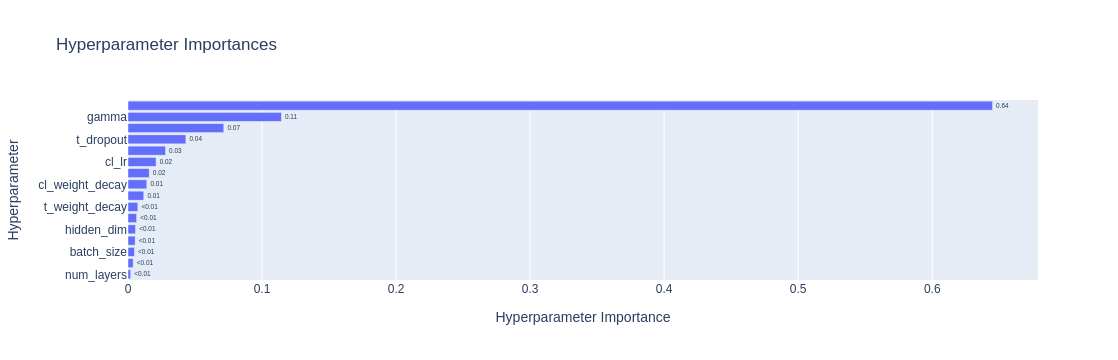}
\caption{Optuna hyperparameter-importance analysis (functional ANOVA). The most
influential knobs for ZAYAN are the contrastive regularization strength
$\gamma$, Transformer dropout (\texttt{t\_dropout}), contrastive learning rate
(\texttt{cl\_lr}), and contrastive weight decay (\texttt{cl\_weight\_decay}),
while batch size, hidden dimension, and depth have comparatively minor impact
within the explored ranges.}
\label{fig:optuna_importance}
\end{figure*}

\begin{figure*}[htbp]
\centering
\includegraphics[width=0.95\textwidth]{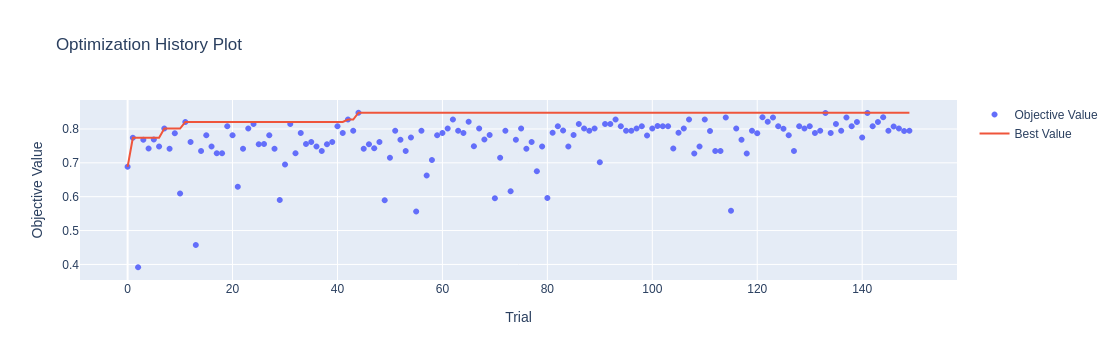}
\caption{Optimization history for the 150-trial Optuna search. The objective is
5-fold cross-validation accuracy on Urban land cover. ZAYAN reaches above
$0.80$ accuracy within the first $\approx 10$ trials, with subsequent trials
yielding incremental gains and several near-optimal configurations clustered
around the best value of $0.848$.}
\label{fig:optuna_history}
\end{figure*}

The global structure of the search is visualized in Fig.~\ref{fig:optuna_parallel}-\ref{fig:optuna_history}. The parallel-coordinate plot (Fig.~\ref{fig:optuna_parallel}) shows that high-performing trials cluster around moderate contrastive dropout (\texttt{cl\_dropout}) and relatively small Transformer dropout (\texttt{t\_dropout}), with intermediate values of temperature $\tau$ and Transformer weight decay also favored. The functional-ANOVA importance analysis (Fig.~\ref{fig:optuna_importance}) and the correlation summary in Table~\ref{tab:optuna_corr} agree that the most impactful knobs are \texttt{t\_weight\_decay}, $\tau$, the hidden dimension and depth of the Transformer, and the two dropout rates, whereas batch size and embedding dimension have comparatively weak effects within the explored ranges. The optimization-history plot (Fig.~\ref{fig:optuna_history}) further confirms that Optuna quickly reaches a high-accuracy regime (above $0.80$ within the first $\sim 10$ trials), after which the search primarily refines regularization and learning-rate settings.

\begin{figure*}[htbp]
\centering
\begin{minipage}{0.49\textwidth}
  \centering
  \includegraphics[width=\linewidth]{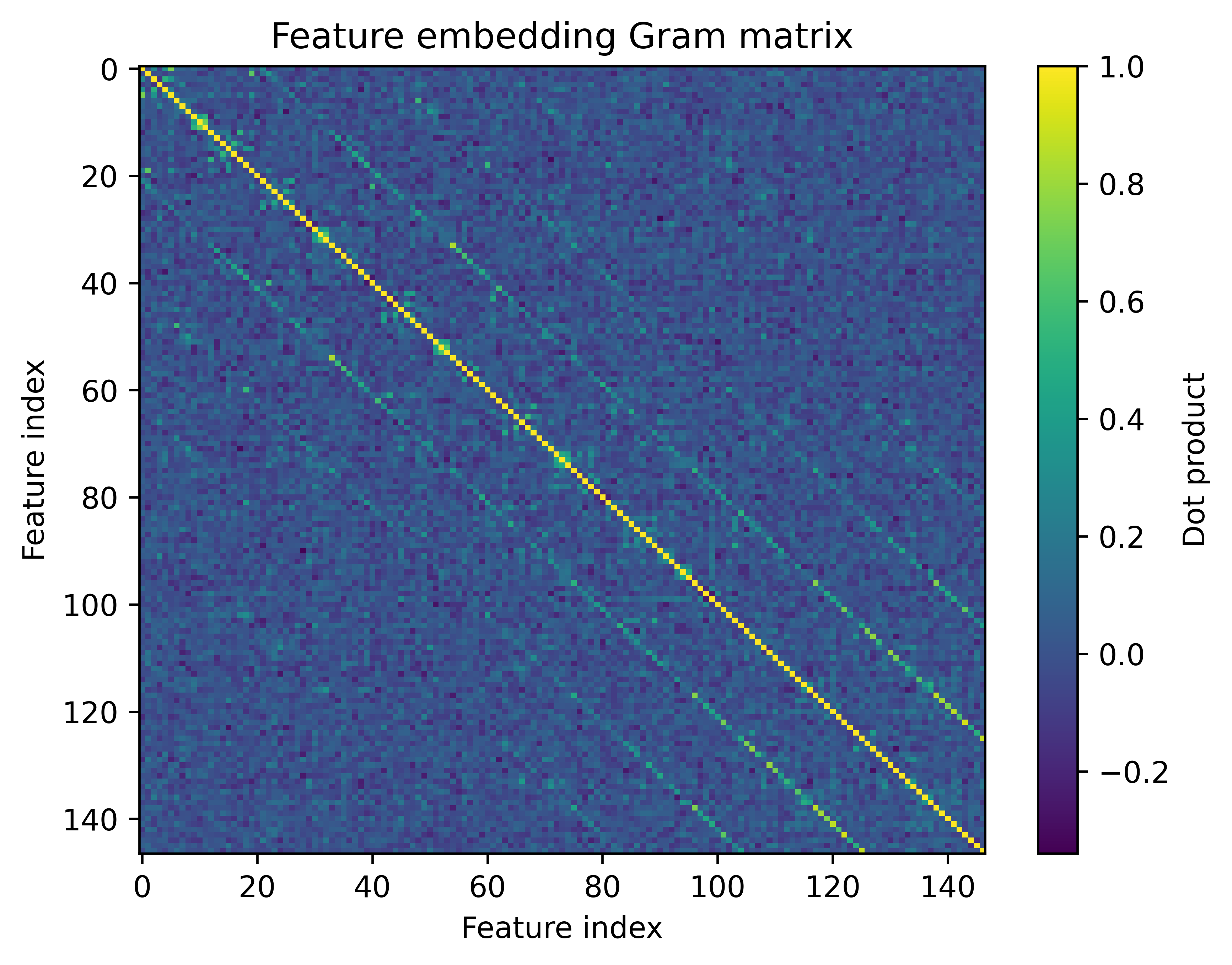}%
  \vspace{-0.5em}
  \caption*{(a) Feature-embedding Gram matrix}
\end{minipage}\hfill
\begin{minipage}{0.49\textwidth}
  \centering
  \includegraphics[width=\linewidth]{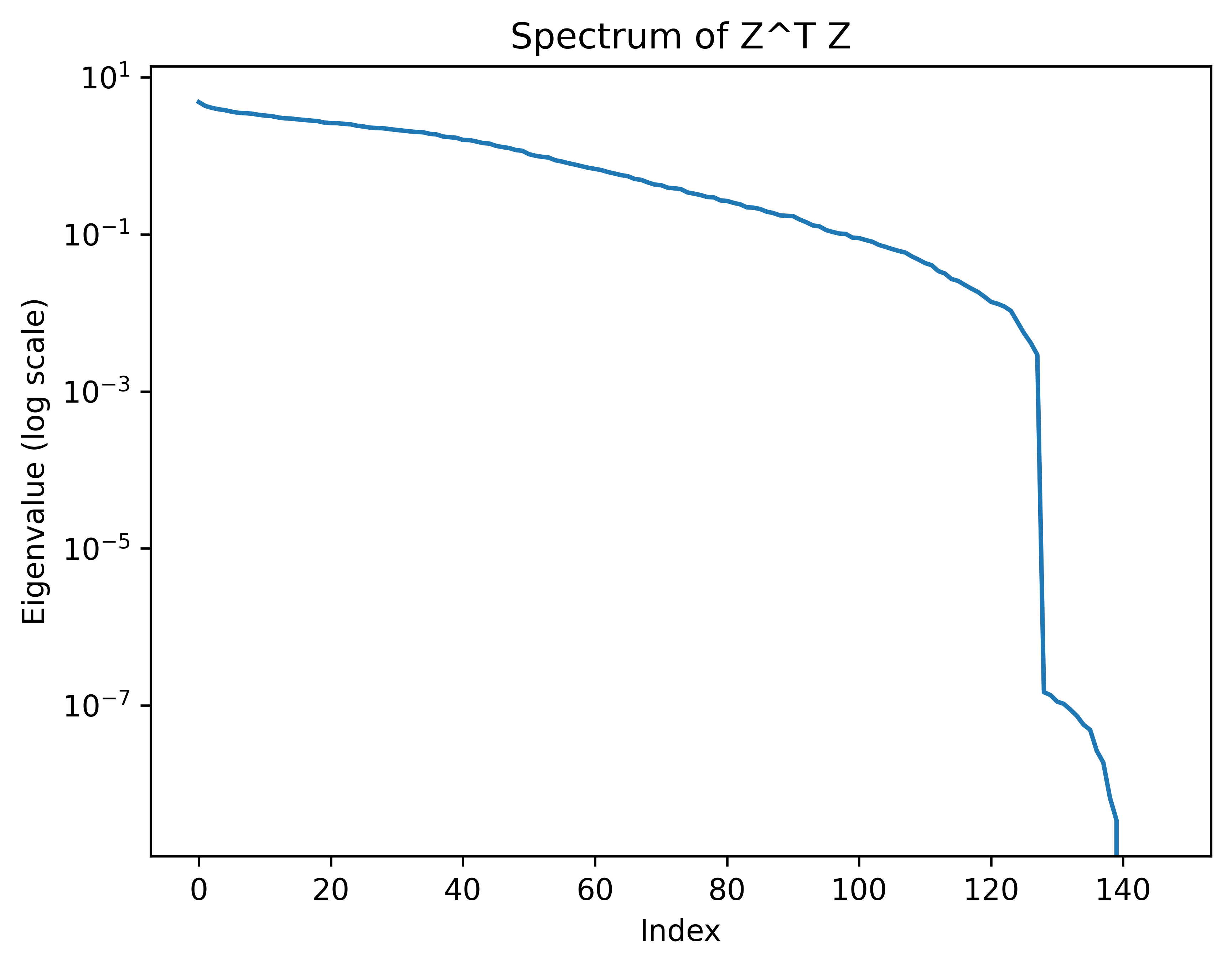}%
  \vspace{-0.5em}
  \caption*{(b) Eigenvalue spectrum of $Z^\top Z$ (log scale)}
\end{minipage}
\vspace{0.25em}
\caption{Redundancy diagnostics for the ZAYAN feature embeddings after Optuna tuning.
Panel (a) shows the Gram matrix of normalized feature embeddings, with an off-diagonal
mean of only $\approx 0.01$, indicating that the learned feature channels are nearly
orthogonal. Panel (b) plots the eigenvalue spectrum of $Z^\top Z$ on a log scale,
revealing a rapid decay and confirming that most variance is concentrated in a low-dimensional
subspace.}
\label{fig:optuna_redun_eigs}
\end{figure*}

Finally, the redundancy-oriented diagnostics in Fig.~\ref{fig:optuna_redun_eigs} indicate that the tuned ZAYAN model learns a well-conditioned representation. The feature-embedding Gram matrix has an off-diagonal mean of only $\approx 0.0098$, suggesting that different feature channels are close to orthogonal on average. The eigenvalue spectrum of $Z^\top Z$ decays rapidly over the 147 feature dimensions, matching the low effective dimensionality observed in our theory-inspired diagnostics and supporting the view that ZAYAN compresses the tabular signal into a compact, low-redundancy subspace.

\begin{table}[htbp]
\centering
\small
\caption{Summary statistics over 150 Optuna trials for ZAYAN on the Urban land cover dataset.
The objective is 5-fold cross-validation accuracy.}
\label{tab:optuna_global_summary}
\begin{tabular}{lcccc}
\toprule
Quantity & Min & Mean & Median & Max \\
\midrule
CV accuracy (objective) & 0.392 & 0.765 & 0.788 & 0.848 \\
Trial duration (s)      & 20.97 & 60.38 & 58.91 & 85.90 \\
\bottomrule
\end{tabular}
\end{table}

\begin{table}[htbp]
\centering
\small
\caption{Stability of the best Optuna trial (5-fold CV accuracy). The 95\% confidence
interval is computed from the empirical standard deviation.}
\label{tab:optuna_best_fold}
\begin{tabular}{cccccc}
\toprule
Fold 1 & Fold 2 & Fold 3 & Fold 4 & Fold 5 & Mean $\pm$ Std \\
\midrule
0.806 & 0.933 & 0.933 & 0.800 & 0.767 & $0.848 \pm 0.071$ \\
\multicolumn{6}{c}{95\% CI for mean: [0.788, 0.908]} \\
\bottomrule
\end{tabular}
\end{table}

\begin{table}[htbp]
\centering
\small
\caption{Pearson correlation between trial objective value and selected hyperparameters
over the 150 Optuna trials (Urban land cover). We report the strongest positive and
negative correlations.}
\label{tab:optuna_corr}
\begin{tabular}{lr}
\toprule
Hyperparameter & Corr.\ with CV accuracy \\
\midrule
\texttt{t\_weight\_decay} &  0.272 \\
\texttt{tau}              &  0.231 \\
\texttt{hidden\_dim}      &  0.226 \\
\texttt{num\_layers}      &  0.200 \\
\texttt{mask\_prob}       &  0.176 \\
\texttt{nhead}            &  0.154 \\
\texttt{emb\_dim}         &  0.141 \\
\texttt{cl\_lr}           &  0.099 \\
\midrule
\texttt{batch\_size}      & -0.059 \\
\texttt{gamma}            & -0.089 \\
\texttt{sigma}            & -0.092 \\
\texttt{lambd}            & -0.150 \\
\texttt{t\_dropout}       & -0.161 \\
\texttt{cl\_weight\_decay}& -0.194 \\
\texttt{t\_lr}            & -0.235 \\
\texttt{cl\_dropout}      & -0.311 \\
\bottomrule
\end{tabular}
\end{table}

\begin{table*}[htbp]
\centering
\tiny
\caption{List of baseline models and their implementation source URLs.}
\label{tab:model_sources}
\begin{tabular}{ll}
\toprule
Model & Source URL \\
\midrule
Naive Bayes      & \url{https://scikit-learn.org/stable/modules/naive_bayes.html} \\
Logistic Regression & \url{https://scikit-learn.org/stable/modules/generated/sklearn.linear_model.LogisticRegression.html} \\
KNN              & \url{https://scikit-learn.org/stable/modules/generated/sklearn.neighbors.KNeighborsClassifier.html} \\
SVM              & \url{https://scikit-learn.org/stable/modules/generated/sklearn.svm.SVC.html} \\
Decision Tree    & \url{https://scikit-learn.org/stable/modules/generated/sklearn.tree.DecisionTreeClassifier.html} \\
MLP              & \url{https://scikit-learn.org/stable/modules/generated/sklearn.neural_network.MLPClassifier.html} \\
Random Forest    & \url{https://scikit-learn.org/stable/modules/generated/sklearn.ensemble.RandomForestClassifier.html} \\
AdaBoost         & \url{https://scikit-learn.org/stable/modules/generated/sklearn.ensemble.AdaBoostClassifier.html} \\
GBM              & \url{https://scikit-learn.org/stable/modules/generated/sklearn.ensemble.GradientBoostingClassifier.html} \\
1D--CNN (PyTorch) & \url{https://pytorch.org/docs/stable/generated/torch.nn.Conv1d.html} \\
LGBM             & \url{https://github.com/microsoft/LightGBM} \\
XGBoost          & \url{https://github.com/dmlc/xgboost} \\
CatBoost         & \url{https://github.com/catboost/catboost} \\
TabNet           & \url{https://github.com/dreamquark-ai/tabnet} \\
TabTransformer   & \url{https://github.com/lucidrains/tab-transformer-pytorch} \\
FT-Transformer   & \url{https://github.com/lucidrains/tab-transformer-pytorch} \\
TabSeq           & \url{https://github.com/zadid6pretam/TabSeq} \\
TANGOS           & \url{https://github.com/OpenTabular/DeepTabular} \\
TabPFN           & \url{https://github.com/PriorLabs/TabPFN} \\
NODE             & \url{https://github.com/OpenTabular/DeepTabular} \\
SAINT            & \url{https://github.com/OpenTabular/DeepTabular} \\
DeepFM           & \url{https://github.com/shenweichen/DeepCTR-Torch} \\
DCN              & \url{https://github.com/shenweichen/DeepCTR-Torch} \\
AutoInt          & \url{https://github.com/OpenTabular/DeepTabular} \\
TabPFN v2        & \url{https://github.com/PriorLabs/tabpfn-extensions} \\
TabR             & \url{https://github.com/OpenTabular/DeepTabular} \\
ProtoGate        & \url{https://github.com/SilenceX12138/ProtoGate} \\
TabM             & \url{https://github.com/OpenTabular/DeepTab} \\
TabulaRNN        & \url{https://github.com/OpenTabular/DeepTab} \\
TANDEM           & \url{https://github.com/erelnaor3/tandem} \\
TabICL           & \url{https://github.com/soda-inria/tabicl/tree/main} \\
\bottomrule
\end{tabular}
\end{table*}
\begin{table}[t]
\centering
\scriptsize
\caption{Remote sensing and flood-related datasets used in our experiments, with download links.}
\label{tab:dataset_urls}
\begin{tabular}{lll}
\toprule
\textbf{Dataset} & \textbf{Source} & \textbf{URL} \\
\midrule
Urban Land Cover
  & UCI
  & \url{https://doi.org/10.24432/C53S48} \\
Forest Type Mapping
  & UCI
  & \url{https://doi.org/10.24432/C5QP56} \\
Crop Mapping Using Fused Optical-Radar
  & UCI
  & \url{https://doi.org/10.24432/C5G89D} \\
Wilt
  & UCI
  & \url{https://doi.org/10.24432/C5KS4M} \\
Flood Risk in India
  & Kaggle
  & \url{https://www.kaggle.com/datasets/s3programmer/flood-risk-in-india} \\
Pluvial Flood Dataset
  & Kaggle
  & \url{https://www.kaggle.com/datasets/oladapokayodeabiodun/pluvial-flood-dataset} \\
Satellite Image Classification
  & Kaggle
  & \url{https://www.kaggle.com/datasets/mahmoudreda55/satellite-image-classification} \\
Census of Individual Trees
  & Kaggle
  & \url{https://www.kaggle.com/datasets/noeyislearning/census-of-individual-trees} \\
\bottomrule
\end{tabular}
\end{table}